\documentclass[10pt,journal,compsoc]{IEEEtran}
\usepackage{amsmath,amsfonts}
\usepackage[nocompress]{cite}
\usepackage{array}
\usepackage{url}
\usepackage{graphicx}
\usepackage{pifont}
\usepackage{booktabs}
\usepackage{amssymb}
\usepackage[ruled,vlined,linesnumbered]{algorithm2e}
\usepackage{multirow}
\usepackage{makecell}
\newtheorem{theorem}{Theorem}
\newtheorem{definition}{Definition}

\usepackage{bm}
\usepackage{caption}
\usepackage{enumitem}
\usepackage[pagebackref=false,breaklinks=true,citecolor=blue,colorlinks,bookmarks=true]{hyperref}
\usepackage{physics}
\usepackage{tikz}
\usepackage{wrapfig}
\usepackage{bbding}
\usepackage{tcolorbox}
\usepackage{rotating}
\usepackage{diagbox}
\usepackage[edges]{forest}
\definecolor{hidden-draw}{RGB}{20,68,106}
\definecolor{hidden-pink}{RGB}{255,245,247}

\begin{document}

\title{Survey of Computerized Adaptive Testing: \\A Machine Learning Perspective}

\author{Yan Zhuang, Qi Liu,~\IEEEmembership{Member,~IEEE,} Haoyang Bi, Zhenya Huang,~\IEEEmembership{Member,~IEEE,} Weizhe Huang,\\ Jiatong Li, Junhao Yu, Zirui Liu, Zirui Hu, Yuting Hong, Zachary A. Pardos, Haiping Ma,\\ Mengxiao Zhu,~\IEEEmembership{Member,~IEEE,} Shijin Wang, Enhong Chen,~\IEEEmembership{Fellow,~IEEE}  
	
		\IEEEcompsocitemizethanks{
		\IEEEcompsocthanksitem 
		Yan Zhuang, Qi Liu, Haoyang Bi, Zhenya Huang, Weizhe Huang, Jiatong Li, Junhao Yu, Zirui Liu, Zirui Hu, Yuting Hong, Mengxiao Zhu, and Enhong Chen are with State Key Laboratory of Cognitive Intelligence, University of Science and Technology of China, China. Yan Zhuang is also with Nanjing University of Aeronautics and Astronautics, China. Zachary A. Pardos is with University of California, Berkeley, USA. Haiping Ma is with Anhui University, China. Shijin Wang is with iFLYTEK Co., Ltd, China. 
		\protect\\
		Corresponding E-mail: qiliuql@ustc.edu.cn
	}


}

%

\markboth{Journal of \LaTeX\ Class Files,~Vol.~14, No.~8, August~2021}%
{Shell \MakeLowercase{\textit{et al.}}: A Sample Article Using IEEEtran.cls for IEEE Journals}


\IEEEtitleabstractindextext{%
	\begin{abstract}
		Computerized Adaptive Testing (CAT) offers an efficient and personalized method for assessing examinee proficiency by dynamically adjusting test questions based on individual performance. Compared to traditional, non-personalized testing methods, CAT requires fewer questions and provides more accurate assessments. As a result, CAT has been widely adopted across various fields, including education, healthcare, sports, sociology, and the evaluation of AI models. While traditional methods rely on psychometrics and statistics, the increasing complexity of large-scale testing has spurred the integration of machine learning techniques. This paper aims to provide a machine learning-focused survey on CAT, presenting a fresh perspective on this adaptive testing paradigm. We delve into measurement models, question selection algorithm, bank construction, and test control within CAT, exploring how machine learning can optimize these components. Through an analysis of current methods, strengths, limitations, and challenges, we strive to develop robust, fair, and efficient CAT systems. By bridging psychometric-driven CAT research with machine learning, this survey advocates for a more inclusive and interdisciplinary approach to the future of adaptive testing.

\end{abstract}

}

\maketitle

\IEEEdisplaynontitleabstractindextext
\IEEEpeerreviewmaketitle

\section{Introduction}\label{introduction}


\IEEEPARstart{T}{he} assessment of intelligent agents, whether human or AI systems, is essential for ensuring that individuals are well-prepared to meet the demands of their respective roles \cite{mehrabi2021survey,rong2023towards}. For humans, assessment results can determine eligibility for opportunities such as admissions or employment. For AI models, these results can indicate whether a system is suitable for deployment and capable of making real-world decisions. Traditionally, assessments have often used a one-size-fits-all approach, where all examinees answer the same set of questions, and a final score is calculated. Examples include traditional paper-and-pencil tests for humans and various gold-standard benchmarks for AI models.

However, as the testing scale increases and the complexity and diversity of agents grow, traditional assessment methods face challenges in efficiency and reliability. Computerized Adaptive Testing (CAT), originating from psychometrics, offers a \textit{personalized} testing paradigm by {identifying and presenting the most informative and valuable questions to each examinee} \cite{chen2015computer,vie2017review}. This method has been widely adopted in high-stakes testing scenarios for humans, such as the SAT, GRE, and GMAT \cite{eignor1993case,luecht1998some}. Recently, CAT has also been increasingly used to assess AI's capabilities, such as textual entailment recognition, chatbots, machine translation, and general-purpose AI systems \cite{otani2016irt, lalor2016building, sedoc2020item,wang2023evaluating}. CAT approach has been proved to require fewer questions to achieve the same level of assessment accuracy for both humans and AI systems \cite{zhuang2025position,kipnis2024texttt}. Essentially, CAT aims to address a critical question about \textit{accuracy} and \textit{efficiency}: {How to accurately estimate an examinee's true proficiency while minimizing the number of questions provided?}

It is a dynamic and interactive process between an examinee  (human or AI model) and a testing system. The testing system includes four main components that take turns: At each test step, \textcolor{black}{the \textbf{Measurement Model}, as the user model}, first uses the examinee's previous responses to estimate their current proficiency, based on cognitive science or psychometrics \cite{ackerman2003using}. Then, the \textbf{Selection Algorithm} picks the next question from the \textbf{Question Bank} according to certain criteria \cite{lord2012applications, chang1996global, bi2020quality}. Most traditional criteria are statistical informativeness metrics, e.g., selecting the question whose difficulty matches the examinee's current proficiency estimate, meaning the examinee has roughly a 50\% chance of getting it right. The above process repeats until a predefined stopping rule is met. Throughout the assessment, \textbf{Test Control} governs various factors such as \textcolor{black}{exposure balance, fairness, and robustness of the testing}. At the conclusion of CAT, the final proficiency estimate—or diagnostic report—serves as the outcome of the assessment.

CAT represents a complex fusion of machine intelligence and assessment techniques. It needs to manage large question banks, adapt to varying examinee proficiencies, and real-time decision-making. Moreover, practical CAT also involves ensuring reliability, fairness, search efficiency, etc. These challenges make CAT a multifaceted decision-making problem. \textcolor{black}{
	With the rise of large-scale and diverse online testing platforms, these challenges have become even more significant. Machine learning (ML), particularly deep learning, offers promising solutions to enhance both the efficiency and accuracy of testing.} Previous CAT surveys \cite{chang2015psychometrics,vie2017review,cheng2008computerized,mujtaba2020artificial} have primarily focused on statistical and psychometric perspectives, concentrating mainly on human assessments. \textcolor{black}{Given CAT's interdisciplinary nature, this paper seeks to explore and review methodologies from a machine-learning perspective.} It is more accessible to a broader readership and provides insights into building strong testing systems for both humans and artificial intelligence.

In the realm of ML, CAT can be conceptualized as a \textit{parameter estimation problem} with a focus on data efficiency \cite{pmlr-v119-mirzasoleiman20a,zhuang2023bounded}: The objective is to determine the values of latent parameters within a model (i.e., the examinee's true proficiency) using the minimum amount of observed data (i.e., the fewest possible questions answered by the examinee). In recent years, there has been a growing interest in applying ML techniques to investigate the four components in CAT. For example, deep learning techniques diagnose examinee's proficiency \cite{wang2020neural} and automate question bank construction \cite{qiu2019question}; data-driven approaches optimize selection algorithms by learning from large-scale response data \cite{zhuang2022fully,li2020data,ghosh2021bobcat}. Despite these efforts, a comprehensive survey that captures the breadth of CAT solutions from a machine-learning perspective is still lacking. Furthermore, the ongoing evolution of machine learning presents new aspects for testing. The contributions of this paper are as follows:
\begin{itemize}
	
	\item 	To our knowledge, this represents the first attempt to comprehensively review CAT solutions through the lens of machine learning. By exploring the existing work in {Measurement Model}, {Selection Algorithms}, {Question Bank Construction}, and {Test Control}, the paper offers a unified framework and encompasses the entire life cycle of the CAT system.

	\item We summarize existing works and draw conclusions on the success and failure attempts of machine learning. Furthermore, we identify \textcolor{black}{key factors that are essential for building reliable and effective CAT systems for both human and AI model evaluation, including exposure control, fairness, robustness, and search efficiency. It offers a more comprehensive perspective.}

	\item We have open-sourced extensible and unified implementations of existing CAT models and relevant resources at \url{https://github.com/bigdata-ustc/EduCAT}. This library aims to assist researchers in swiftly developing a CAT system, encouraging collaboration, and ultimately leading to more sophisticated and effective CAT systems.

\end{itemize}
The paper is organized as follows. In Section \ref{EOC}, we introduce the background of CAT. Then in Section \ref{OV}, we provide the formulation of CAT's task. After that, Section \ref{CDM}, \ref{SA}, and \ref{QBC} respectively review the existing methods for the measurement model, selection algorithms, and question bank construction. Given the fact that the selection algorithm is the core component for achieving the adaptivity, \textcolor{black}{this survey mainly focuses on its  recent advancements in machine learning and deep learning.} In Section \ref{AOC}, we summarize the key factors in the application of CAT. Section \ref{EM} discusses how to evaluate the CAT.

\section{Evolution of CAT} \label{EOC}
The evolution of CAT is a fascinating journey through time, marked by significant milestones. Adaptive testing began with Alfred Binet's intelligence test in 1905 \cite{wainer2000computerized}. The 1950s saw the advent of computers, transforming adaptive testing into CAT. Key advancements in the 1970s and 1980s, particularly the integration of the psychometric model, enhanced assessment accuracy \cite{sands1997computerized, roskam1984new}. The 1990s internet boom made CAT widely accessible, leading to its use in major tests like the GRE, GMAT, and SAT. These tests, though evolved, still rely on adaptive principles. Various statistical methods optimize the testing experience, and CAT became a major focus in human measurement, covering education \cite{verschoor2010mathcat,luecht1998some,wainer1987item}, healthcare \cite{gibbons2016computerized,gibbons2013computerized,gibbons2017development}, sociology \cite{montgomery2013computerized}, and sports \cite{ando2018validity,yurtcu2021bibliometric}.

Recently, researchers have increasingly explored applying CAT to AI model evaluation. Existing benchmarks often contain redundant, low-quality, contaminated, or even erroneous questions, affecting the efficiency and reliability of AI assessments \cite{zhuang2025position}. By leveraging adaptive testing, researchers can analyze the characteristics of benchmark questions to customize assessments for each AI system and estimate the latent traits behind each model's responses, rather than merely calculating accuracy. Guided by CAT and psychometrics, various efficient methods have emerged in various aspects of AI evaluation, including performance estimation \cite{lalor2016building,polo2024tinybenchmarks}, question selection \cite{guinetautomated,rodriguez2021evaluation,zhuang2025position}, and understanding experimental results \cite{MARTINEZPLUMED201918,martinez2016making}. These methods aim to identify informative and valuable subsets from large-scale datasets to improve the reliability of AI system evaluations.

	Current CAT research spans a wide range of topics, including the development of question banks, question selection, proficiency estimation, and various issues related to test security and reliability. They are critical to ensuring that CAT remains a reliable, valid, and fair assessment. Machine Learning is revolutionizing CAT by enabling sophisticated analysis of large datasets, detailed behavior modeling, and flexible adaptation to diverse testing environments \cite{zhuang2022fully}. Despite the ML in CAT is still in its early stages, its potential is evident. \textcolor{black}{Machine learning offers new solutions to improve how we define, analyze, and apply CAT \cite{zheng2024mxml}.} This survey aims to provide an overview and understanding of traditional statistical-based and recent ML-based CAT.

\begin{figure*}[t]
	\centering
	\includegraphics[width=1\linewidth]{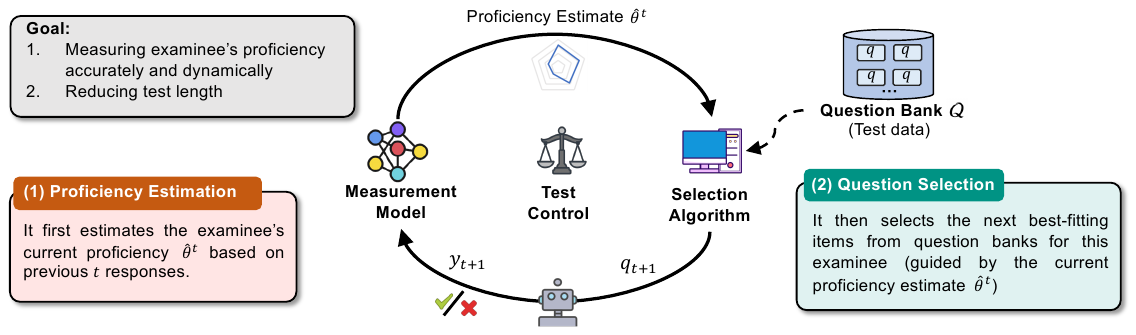}
		\vspace{-8pt}
	\caption{The workflow of CAT: At step $t$, the selection algorithm adaptively selects next question $q_{t+1}$ based on examinee's current proficiency ${\theta}^{t}$ estimated by measurement models.}
		\vspace{-8pt}
	\label{mini_frame}
\end{figure*}

\section{Overview} \label{OV}
An important assumption \cite{chang2015psychometrics} of CAT is that examinee's true proficiency level $\theta_0\in \mathbb{R}^d$ is constant throughout the test. \textcolor{black}{Here, \(d\) represents the proficiency's dimension; for example, \(\theta\) may correspond to a unidimensional overall ability level (\(d = 1\)) or a multidimensional vector representing mastery levels across \(d\) distinct knowledge concepts.} The primary goal of CAT is to \textit{accurately} and \textit{efficiently} estimate examinees' true proficiency levels by having them answer questions. Thus, CAT systems are designed to achieve two key objectives: \textbf{(1)} to use the responses to estimate an examinee's proficiency $\theta$ such that it closely approximates the true proficiency $\theta_0$ by the end of the test, and \textbf{(2)} to select the most valuable and fitting questions for each examinee, thereby reducing test length.

\subsection{Task Formalization}
To achieve the aforementioned objectives, CAT operates as an iterative and interactive process: As illustrated in \figurename\ \ref{mini_frame}, at test step $t\in[1, 2,...,T]$ in CAT, examinee's current proficiency estimate $\hat{\theta}^{t}$ is estimated using previous $t$ responses; then leverage $\hat{\theta}^{t}$ to retrieve the next question $q_{t+1}$ from question bank $\mathcal{Q}$ to ask examinee, and receive the next response label $y_{t+1}$. These interactions form a response sequence $\{(q_1,y_1),(q_2,y_2),...,(q_T,y_T)\}$, where $y_t=1$ if the response to $q_t$ is correct and 0 otherwise. To achieve the goals of CAT, each test step involves two critical processes:

\textbf{(1) Proficiency Estimation.} The Measurement Model, denoted by $f(\cdot)$, acts as a user model, predicting the probability of a correct response by an examinee with proficiency $\theta$, which is denoted as $f(q, \theta) = P(y=1|q,\theta)$. The implementation of measurement model often draws upon cognitive science \cite{de2009dina} or psychometrics\cite{ackerman2003using}. To accurately estimate examinee's proficiency at each step, various estimation methods can be used, e.g., Maximum Likelihood Estimation (MLE) or Bayesian Estimation. In applications, the binary cross-entropy loss is frequently utilized: at step $t$, given previous $t$ responses $\mathcal{D}_{1:t}=\{(q_1,y_1),(q_2,y_2),...,(q_t,y_t)\}$, the corresponding empirical loss is:
\begin{align}
			{L}(\theta)&=\sum_{(q,y)\in \mathcal{D}_{1:t}} \ell(y,f(q,\theta)) \\
			&=-\sum_{(q,y)\in \mathcal{D}_{1:t}}{y\log f(q,\theta)+(1-y)\log (1-f(q,\theta))},\nonumber
\end{align}
thus the current estimate of proficiency, $\hat{\theta}^t$, is obtained by minimizing the loss function $L(\theta)$: $\hat{\theta}^t=\mathop{\arg\min}_{\theta}{{L}(\theta)}$.

\textbf{(2) Question Selection. }The heart of CAT is an algorithm that picks the next question $q_{t+1}$ from the question bank $\mathcal{Q}$, using examinee's current proficiency estimate $\hat{\theta}^{t}$ as a guide:
\begin{equation} 
	q_{t+1} = \mathop{\arg\max}_{q\in \mathcal{Q}} \mathcal{V}_q(\hat{\theta}^{t}),
\end{equation}
where $\mathcal{V}_q(\hat{\theta}^{t})$ is the value of question $q$. For instance, $\mathcal{V}$ might be a measure of how much information the question will provide about the examinee's proficiency, or it could be the output of a policy $\pi$ specifically designed to determine question selection.

After receiving new response label $y_{t+1}$, measurement model updates and estimates proficiency $\hat{\theta}^{t+1}$. {The above process will be repeated for $T$ times, ensuring the final step estimate $\hat{\theta}^T$ close to the true $\theta_0$, i.e., 
	\begin{definition}[Definition of CAT] \label{def1}
		The goal of CAT is to find a question set $S = \{q_1, q_2, ..., q_T\}$ of size $T$, such that the final step estimate $\hat{\theta}^T$, derived from $S$ and their corresponding response labels $y$, closely approximates the examinee's true proficiency $\theta_0$:
		\begin{equation}\label{def_eq}
			\min_{|S|=T}	\Vert {\hat{\theta}^T}-\theta_0 \Vert .
		\end{equation}
	\end{definition}
	However, solving this optimization problem directly is impractical, as the true proficiency $\theta_0$ is not observable, and even the examinees may not know their exact proficiency level. Consequently, existing methods are all approximations of this target. For example, traditional statistical selection methods \cite{lord2012applications,chang1996global} utilize the asymptotic statistical properties of the MLE to reduce estimation uncertainty, e.g., selecting questions whose difficulty closely match the examinee's current estimated proficiency $\hat{\theta}^t$. More recent Subset Selection approaches \cite{zhuang2023bounded} try to identify a theoretical approximation of $\theta_0$ to serve as a new objective for optimization. For further details, refer to Section \ref{SA}. 
	
	Meanwhile, as a practical system, considerations extend beyond proficiency estimation objective (Definition \ref{def1}). Factors such as question exposure control, robustness, fairness, and search efficiency must be addressed as well. An exhaustive discussion of these factors is presented in Section \ref{AOC}.

	Evaluation Methods: To validate the accuracy of the estimated proficiency, two primary approaches are employed: 1) Performance prediction: using the examinee's estimated values within the measurement model to predict the correctness label $y$ of the responses on examinee's reserved response data, often measured by cross-entropy; 2) Proficiency estimation: using simulation to generate true proficiency values $\theta_0$, simulating the examinee's responses to each question. Then the Mean Squared Error (MSE) between the estimates and the simulated true values can be calculated. The details can be found in Section \ref{EM}.

	\tikzstyle{my-box}=[
	rectangle,
	draw=hidden-draw,
	rounded corners,
	text opacity=1,
	minimum height=1.5em,
	minimum width=5em,
	inner sep=2pt,
	align=center,
	fill opacity=.5,
	line width=0.8pt,
	]
	\tikzstyle{leaf}=[my-box, minimum height=1.5em,
	fill=hidden-pink!80, text=black, align=left,font=\normalsize,
	inner xsep=2pt,
	inner ysep=4pt,
	line width=0.8pt,
	]
	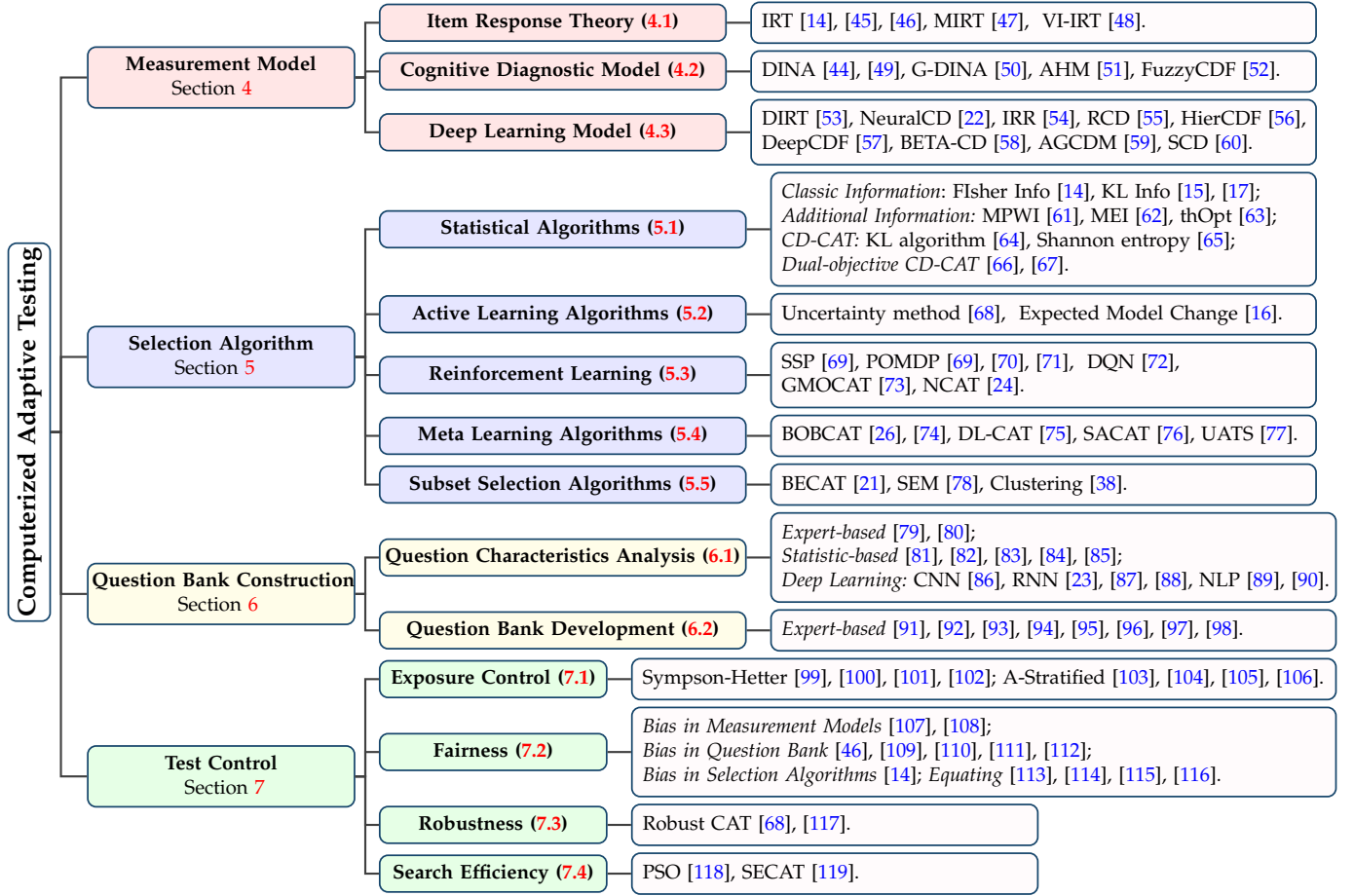
\begin{figure*}[t!]
		\centering
		\resizebox{\textwidth}{!}{
			\begin{forest}
				forked edges,
				for tree={
					grow=east,
					reversed=true,
					anchor=base west,
					parent anchor=east,
					child anchor=west,
					base=left,
					font=\large,
					rectangle,
					draw=hidden-draw,
					rounded corners,
					align=left,
					text centered,
					minimum width=4em,
					edge+={darkgray, line width=1pt},
					s sep=3pt,
					inner xsep=2pt,
					inner ysep=3pt,
					line width=0.8pt,
					ver/.style={rotate=90, child anchor=north, parent anchor=south, anchor=center},
				},
				where level=1{text width=14em,font=\normalsize,}{},
				where level=2{text width=15em,font=\normalsize,}{},
				where level=3{text width=17em,font=\normalsize,}{},
				[
				\textbf{Computerized Adaptive Testing}, ver
				[
				\textbf{Measurement Model} \\ {\qquad  } Section \ref{CDM}, fill=red!10, text width=13em
				[ 
				\textbf{Item Response Theory (\ref{LTM})}, fill=red!10, text width=17em
				[
				{\ }IRT \cite{embretson2013item, hambleton1991fundamentals, lord2012applications}{, }MIRT \cite{reckase200618}{,
				} VI-IRT  \cite{wu2020virt}{.}, leaf, text width=28em
				]
				]
				[ 
				\textbf{Cognitive Diagnostic Model (\ref{DCM})}, fill=red!10, text width=17em
				[                            
				{\ }DINA \cite{de2009dina,von2014dina}{, }G-DINA \cite{de2011gdina}{, }AHM \cite{leighton2004ahm}{, }FuzzyCDF \cite{liu2018fuzzycd}{.}, leaf, text width=28em
				]
				]
				[
				\textbf{Deep Learning Model (\ref{DLM})}, fill=red!10, text width=17em
				[
				{\ }DIRT \cite{cheng2019dirt}{,} NeuralCD \cite{wang2020neural}{, }IRR \cite{tong2021irr}{, }RCD \cite{gao2021rcd}{, }HierCDF \cite{li2022hiercdf}{, }\\{\ }DeepCDF \cite{gao2022deepcdf}{, }BETA-CD \cite{bi2023beta}{, }AGCDM \cite{pei2022self}{, }SCD \cite{wang2023scd}{.},leaf, text width=28em
				]
				]
				]
				[
				\textbf{Selection Algorithm} \\ {\qquad \ }Section \ref{SA}, fill=blue!10, text width=13em
				[
				\textbf{Statistical Algorithms (\ref{StaA})}, fill=blue!10, text width=18em
				[
				{\ }\textit{Classic  Information}: FIsher Info \cite{lord2012applications}{, }KL Info \cite{chang1996global,chang2015psychometrics}{; }\\ {\ }\textit{Additional  Information:} MPWI \cite{barrada2009metodologia}{, }MEI \cite{van1998bayesian}{, }thOpt \cite{barrada2006maximum}{; }\\ {\ }\textit{CD-CAT:} KL algorithm \cite{cheng2009cognitive}{, }Shannon entropy \cite{tatsuoka2002data}{; } \\ {\ }\textit{Dual-objective CD-CAT} \cite{kang2017dual,zheng2018information}{.},leaf, text width=27em
				]
				]
				[
				\textbf{Active Learning Algorithms (\ref{ALA})}, fill=blue!10, text width=18em
				[
				{\ }Uncertainty method \cite{veldkamp2019robust}{, } Expected Model Change \cite{bi2020quality}{.}, leaf, text width=27em
				]
				]
				[
				\textbf{Reinforcement Learning (\ref{rl_method})}, fill=blue!10, text width=18em
				[
				{\ }SSP \cite{gilavert2022computerized}{, }POMDP \cite{nurakhmetov2019reinforcement,gilavert2022computerized,chen2018recommendation}{, }
				DQN \cite{li2023deep}{, }\\{\ }GMOCAT \cite{wang2023gmocat}{, }NCAT \cite{zhuang2022fully}{.}, leaf, text width=27em
				]
				]
				[
				\textbf{Meta Learning Algorithms (\ref{meta})}, fill=blue!10, text width=18em
				[
				{\ }BOBCAT \cite{ghosh2021bobcat,feng2023balancing}{, }DL-CAT \cite{ma2023novel}{, }SACAT \cite{yu2023sacat}{, }UATS\cite{yu2024a}{.}, leaf, text width=27em
				]
				]
				[
				\textbf{Subset Selection Algorithms (\ref{SSA})}, fill=blue!10, text width=18em
				[
				{\ }BECAT \cite{zhuang2023bounded}{, }SEM \cite{mujtaba2021multi}{, }Clustering \cite{polo2024tinybenchmarks}{.}, leaf, text width=27em
				]
				]
				]
				[
				\textbf{Question Bank Construction} \\ {\qquad \qquad} Section \ref{QBC}, fill=yellow!10, text width=13em
				[
				\textbf{Question Characteristics Analysis (\ref{qca})}, fill=yellow!10, text width=18em
				[
				{\ }\textit{Expert-based} \cite{lopez2008helping,kozierkiewicz2014item}{; }\\{\ }\textit{Statistic-based} \cite{chen2004personalized,magno2009demonstrating,chang2009applying,sun2014alternating, xiong2022data}{; }\\{\ }\textit{Deep Learning:} CNN \cite{huang2017question}{, }RNN \cite{qiu2019question,huang2019hierarchical,lei2021consistency}{, }NLP \cite{yin2019quesnet, ning2023towards}{.}
				, leaf, text width=28em
				]
				]
				[
				\textbf{Question Bank Development (\ref{QBD})}, fill=yellow!10, text width=18em
				[
				{\ }\textit{Expert-based} \cite{reckase2010designing,he2014item,van2000integer,way2005developing,van2006assembling,ariel2004constructing,gulliksen2013theory,swanson1993model}{.}, leaf, text width=28em
				]
				]
				]
				[
				\textbf{Test Control} \\ {     \ \ }Section \ref{AOC}, fill=green!10, text width=13em
				[
				\textbf{Exposure Control (\ref{CC})}, fill=green!10, text width=11em
				[
				{\ }Sympson-Hetter \cite{sympson1985controlling,van2004constraining,van2007conditional,barrada2009multiple}{; }A-Stratified \cite{chang1999stratified,chang2001stratified,yi2001stratified,barrada2014optimal}{.}, leaf, text width=35em
				]
				]
				[
				\textbf{Fairness (\ref{Fairness})}, fill=green!10, text width=11em
				[
				{\ }\textit{Bias in Measurement Models} \cite{Thompson2022Is,kizilcec2022algorithmic}{; }\\{\ }\textit{Bias in Question Bank} \cite{hambleton1991fundamentals,chu2013detecting,camilli1994methods,mellenbergh1989item,Roberts2017Stan}{; }\\{\ }\textit{Bias in Selection Algorithms} \cite{lord2012applications}{; }\textit{Equating} \cite{brigman1976multiple,van2000test,jansubpopulation,sawaki2001comparability}{.},leaf, text width=35em
				]
				]
				[
				\textbf{Robustness (\ref{Robustness})}, fill=green!10, text width=11em
				[
				{\ }Robust CAT\cite{veldkamp2019robust,zhuang2022robust}{.},leaf, text width=20em
				]
				]
				[
				\textbf{Search Efficiency (\ref{SE})}, fill=green!10, text width=11em
				[
				{\ }PSO \cite{huang2009adaptive}{, }SECAT \cite{hong2023search}{.},leaf, text width=20em
				]
				]
				]
				]
			\end{forest}
		}
		\caption{Summary of representative Computerized Adaptive Testing methods in machine learning perspective.}
		\vspace{-10pt}
		\label{summary}
	\end{figure*}

	\subsection{Categorization}

	As shown in \figurename\  \ref{summary}, we categorize existing CAT research into four major components involved in the testing process described above: 1) Measurement Model, 2) Selection Algorithm, 3) Question Bank Construction, and 4) Test Control. Each part is further divided based on the different techniques employed. The following four sections (Sections~\ref{CDM}–\ref{AOC}) provide detailed introductions and literature reviews of these components, \textcolor{black}{with a particular focus on their theoretical foundations and methodological developments from a machine learning perspective.}

	\section{Measurement Model} \label{CDM}

			\textcolor{black}{The existing methods for Measurement Model can be categorized into three main types: Item Response Theory (IRT), Cognitive Diagnostic Model (CDM), and Deep Learning Model.
In the first 50 years of CAT development, IRT was the dominant modeling framework and widely adopted in operational systems. It was not until 2009, with the introduction of CD-CAT \cite{cheng2009cognitive}, that CDM began to be used as the underlying measurement model in adaptive testing. More recently, with the increasing scale and complexity of adaptive assessments and the rise of deep learning, a variety of new measurement models have emerged that go beyond traditional IRT and CDM.}
	
	 Such classification is grounded on the nature of \textit{proficiency representation} ($\theta$) in CAT—ranging from an overall numerical ability value (i.e., IRT), to discrete cognitive states across different knowledge concepts (i.e., CDM), to a unified modeling approach via deep learning techniques (i.e., Deep Learning Model). The choice of model should depend on the specific goals of the assessment, the nature of the data, and the resources available. Regardless of the chosen Measurement Model for estimating proficiency, the objective remains consistent: to minimize the error between the estimate and the true value at each step, expressed as $\Vert {\hat{\theta}^t}-\theta_0 \Vert \to 0$.

	\subsection{Item Response Theory}\label{LTM}
	\textcolor{black}{In IRT \cite{embretson2013item}, an examinee's proficiency is typically represented as a continuous scalar variable, referred to as overall ability. As a foundational framework in measurement models, IRT represents the examinee's general level of proficiency using a latent trait parameter $\theta$.} One of the most widely used models in IRT is the Three-Parameter Logistic Model (3PL-IRT). It utilizes a logistic-like interaction function to model the probability of examinee's correct response to question $j$, i.e., 
	\begin{equation}
		f(q_j, \theta)=c_j + \frac{1-c_j}{1+e^{-\alpha_{j}(\theta-\beta_j)}}.
	\end{equation}
	The 3PL-IRT model introduces three parameters ($\beta_j, \alpha_j, c_j$) for each test question $j$: The difficulty parameter $\beta_j$ corresponds to the level of proficiency at which an examinee has a 50\% chance of answering the question correctly; The discrimination parameter $\alpha_j$ describes how well the question differentiates between examinees with different ability; The guessing parameter $c_j$ represents the probability that an examinee with a very low proficiency will answer the question correctly. {In CAT systems, these parameters are pre-calibrated and remain fixed during the testing process}, with annotation and calibration methods detailed in Section~\ref{QBC}.
	IRT takes into account the number of questions answered correctly and the difficulty of the question. Almost all major adaptive tests for humans, such as SAT and GRE, are developed by using IRT, because \textit{the methodology can significantly improve measurement reliability and interpretability} \cite{an2014item}. Recently, for AI system evaluation, Polo et al. \cite{polo2024tinybenchmarks} successfully selected 100 informative curated questions from MMLU \cite{hendrycks2021measuring}, a popular multiple-choice QA benchmark consisting of 14K questions, and accurately estimated the performance of LLMs.

 Multidimensional IRT (MIRT) \cite{reckase200618}, on the other hand, extends IRT to multiple dimensions, allowing for the modeling of multiple latent traits simultaneously. Despite the great interpretability of (M)IRT models, their performance is constrained by the simplicity of the interaction function, and they lack fine-grained modeling about examinee's cognitive states on individual knowledge concepts.

	\subsection{Cognitive Diagnostic Model}\label{DCM}
Cognitive Diagnostic Model (CDM) is another representative class of measurement models, focusing on discrete knowledge concepts. Specifically, in CDMs, examinee proficiency is \textit{knowledge concept-wise} and usually \textit{dichotomous}, which indicates whether an examinee has mastered a knowledge concept or not. \textcolor{black}{For example, in a mathematics assessment, knowledge concepts may include addition, fractions, or solving linear equations. We continue to use $\theta$ to denote examinee proficiency for consistency.} For example, the DINA method \cite{de2009dina, von2014dina} models examinee proficiency $\theta=\{ \theta_{(1)},\theta_{(2)},...,\theta_{(K)} \}$ as their dichotomous knowledge mastery levels on all $K$ concepts. Given the Q-matrix $Q\in \mathbb{R}^{|\mathcal{Q}|\times K}$which is a binary matrix that indicates which knowledge concepts are associated with a question in bank $\mathcal{Q}$. DINA method focuses only on the knowledge concepts related to the target question $j$, where $Q_{jk}=1$. Thus, the examinee's binary response variable (with proficiency $\theta$) to question $j$ is $\prod_{k,Q_{jk}=1}\theta_{(k)}$, and models questions as ``slip'' and ``guess'' parameters:
	\begin{equation}
		f(q_j, \theta)=(1-s_j)^{\prod_{k,Q_{jk}=1}\theta_{(k)}}g_j^{1-\prod_{k,Q_{jk}=1}\theta_{(k)}},
	\end{equation}
	where $s_j$ is the slip parameter, indicating the likelihood of an incorrect response despite mastery, and $g_j$ is the guess parameter, reflecting the chance of a correct guess in the absence of mastery. Its extension G-DINA\cite{de2011gdina} provides a granular view of examinee proficiency, while FuzzyCDF\cite{liu2018fuzzycd} leverages fuzzy set theory for nuanced diagnostics from both objective and subjective data. Another approach, the Attribute Hierarchy Method \cite{leighton2004ahm}, applies rule space theory to structure knowledge dependencies and align examinee proficiencies with the nearest ideal cognitive patterns to obtain diagnostic results.

	Compared to IRT, CDM offers a more granular and comprehensive assessment of examinee proficiencies. They are particularly adept at {providing detailed feedback on an individual's strengths and weaknesses across multiple knowledge concepts}, which is important for CAT and targeted further interventions. These models underscore a critical shift towards a more nuanced understanding of learning and proficiency, recognizing the multifaceted nature of knowledge acquisition and adaptive testing.

	\subsection{Deep Learning Model}\label{DLM}
	In recent years, the rapid growth of deep learning techniques stimulates the development of deep learning-driven Measurement Models. Compared to traditional models, deep learning methods are more suitable for measurements in large-scale data scenarios (e.g., online learning platforms) due to their efficiency and ability to learn the complex interaction pattern between examinees and questions.

	\textcolor{black}{In these models, an examinee's proficiency $\theta$ is typically represented by a high-dimensional latent vector (embedding). Similarly, each question is encoded as an question embedding \(e_j = \text{Embed}(q_j)\). These embeddings are passed through a multi-layer neural network to predict the probability of a correct response:
		\begin{equation}
			f(q_j, \theta) = \phi_n\left( \cdots \phi_1\left( W [\theta ; e_j] + b \right) \cdots \right),
		\end{equation}
		where \(W\) and \(b\) are the weight matrix and bias vector, respectively, and \(\phi_k(\cdot)\) denotes the activation function at the \(k\)-th layer (e.g., ReLU, Tanh, or Sigmoid).}

Based on this framework, several deep learning-based measurement models have demonstrated strong performance. For example, DIRT \cite{cheng2019dirt} uses a neural network to capture semantic information from question texts to empower accuracy. NeuralCD \cite{wang2020neural} utilizes a non-negative full connection neural network to capture the complex interaction, with the ability to generalize to other measurement models. Considering the complex heterogeneous relationships between examinees, questions, and knowledge concepts, massive efforts have also been made to leverage them to enhance measurements \cite{gao2021rcd,li2022hiercdf,gao2022deepcdf}.

		{\textit{Discussion:}} In the CAT process, only a limited number of examinee responses can be obtained for proficiency estimation. To some extent, \textit{CAT can be viewed as a proficiency measurement under a cold start scenario}. The performance of the measurement model is a critical factor in ensuring the accuracy of proficiency estimations within CAT. Meanwhile, it is important to note that the choice of the measurement model can significantly influence the selection of corresponding question selection algorithm.

	\section{Selection Algorithm} \label{SA}
	The selection algorithm is CAT's core of implementing adaptivity and is the focal point of this survey. It utilizes the proficiency estimate obtained from the Measurement Model (introduced in the above section) to choose the next most suitable question, ensuring an accurate estimation of proficiency while using the fewest possible questions. Question selection algorithms can be categorized into traditional methods based on statistical information, as well as more recent machine learning methods, e.g., data-driven approaches (i.e., Reinforcement Learning and Meta Learning), and Subset Selection are becoming increasingly prevalent. 
	
	\subsection{Statistical Algorithms}\label{StaA}
Generally, a practical approach to designing a selection algorithm involves developing quantitative methods to assign a numerical value to each question in the bank $\mathcal{Q}$. Classical statistical selection algorithms define the value of a question as the \textit{informativeness} it provides about the examinee's potential ability estimation. The next question index $j_{t+1}$ can be selected from bank $\mathcal{Q}$ based on current estimate $\hat{\theta}^{t}$: 
	\begin{equation} \label{infor}
		j_{t+1} = \arg\max_{q_j\in \mathcal{Q}} \mathcal{I}_j(\hat{\theta}^{t}),
	\end{equation}
	where $\mathcal{I}_j(\cdot)$ is the informativeness of question $q_j$ (e.g., Fisher information). As illustrated in Definition \ref{def1}, CAT assumes that each examinee has a true proficiency value ($\theta_0$) and it is considered as a \textit{parameter estimation process}. The informativeness of an question can thus be interpreted as the expected contribution of the response on this question to the parameter estimation. This concept will be reflected in various selection algorithms discussed later.

	\textbf{Fisher Information.} In the parameter estimation problems, Fisher Information \cite{rissanen1996fisher} is a concept from information theory and statistics that measures the amount of information that an observable random variable carries about the unknown parameter. In CAT, Fisher Information is often used to quantify the amount of information that a question provides about an examinee's proficiency \cite{cheng2008computerized}. Specifically, we consider a random variable $\mathcal{D}_j=(q_j,y_j)$ for which the pdf or pmf is $f(q_j, \theta)$, where $\theta$ is the unknown parameter. The fisher info contained in the variable $\mathcal{D}_j$ is defined as: $\mathcal{I}_j(\theta) =\mathbb{E}_{y_j} [(\nabla_{\theta} L(\mathcal{D}_j|\theta))^2]  = \frac{(\nabla_{\theta} f(q_j,\theta))^2}{f(q_j,\theta)(1-f(q_j,\theta))}$, where $L(\mathcal{D}|\theta)={y\log f(q,\theta)+(1-y)\log (1-f(q,\theta))}$ is the likelihood function of $\mathcal{D}$ with respect to the parameter $\theta$. 
	
	Thus, when using 3PL-IRT to model the $f(q,\theta)$ and given current estimate $\hat{\theta}^t$, the Fisher Information of question $j$ can be calculated as:
	\begin{equation}\label{fisher}
		\mathcal{I}_j(\hat{\theta}^t) = \frac{(1-c_j)\alpha_j^2 e^{-\alpha_j(\hat{\theta}^t-\beta_j)}}{(1+e^{-\alpha_j(\hat{\theta}^t-\beta_j)})^2[1-c_j+c_j (1+e^{-\alpha_j(\hat{\theta}^t-\beta_j)})]}. \nonumber
	\end{equation}
	One crucial property of Fisher information is that its reciprocal (matrix inverse), is the variance (covariance matrix) of the asymptotic distribution of the proficiency estimate:

	\begin{theorem}[The asymptotic distribution of MLE proficiency estimate \cite{ross2014first}] \label{theorm}
		At each step $t$, based on the observation of examinee's previous $t$ responses, the current proficiency estimate $\hat{\theta}^t$ (estimated by MLE) satisfies the asymptotic normal distribution: $\hat{\theta}^t \sim \mathcal{N}\left(\theta_0, \frac{1}{t\mathcal{I}(\theta_0)}\right).$
	\end{theorem}
	Obviously, as the number of questions $t$ or the Fisher Information $\mathcal{I}(\theta_0)$ increases, the variance of the estimate decreases. \textcolor{black}{Since $\hat{\theta}^t$ is asymptotically unbiased (i.e., $\mathbb{E}[\hat{\theta}^t]=\theta_0$), a lower variance implies a more concentrated distribution around $\theta_0$, thereby reducing estimation uncertainty and improving the estimation efficiency.}

Fisher information has been popular in the development of personalized testing over the decades and extensively applied in various standardized human assessments. Similarly, for AI model evaluations, particularly for LLMs, the simple Fisher method allows for accurate performance estimation using only a small sample of test data. For example, Kipnis et al. \cite{kipnis2024texttt} reduce six commonly used benchmarks to less than 3\% of their original size while accurately estimating the performance of over 5,000 LLMs.

	When the measurement model is MIRT, Fisher information naturally extends from a scalar to a matrix \cite{hooker2009paradoxical}. Specifically, the information matrix provided by question $j$ at proficiency $\theta$ (now a vector) is defined as: $\mathcal{I}_j(\theta) = \mathbb{E}_{y_j}\left[\nabla \log L(\mathcal{D}_j|\theta) \nabla \log L(\mathcal{D}_j|\theta)^\top\right]$. This matrix's inverse approximates the covariance of the MLE proficiency estimate. Based on this, various selection algorithms have been proposed, including D-Optimality (maximizing the determinant), A-Optimality (minimizing the trace of the inverse), and E-Optimality (maximizing the smallest eigenvalue) \cite{reckase200618}.

	\textbf{Kullback-Leibler Information.}
	Fisher information is widely used in CAT for its theoretical foundation and mathematical simplicity. However, its effectiveness diminishes when the proficiency estimate deviates from the true value $\theta_0$ \cite{chang2015psychometrics}. This issue becomes particularly evident in the early stages of a test when the estimate is still unstable due to limited responses. To address this issue, Chang et al. \cite{chang1996global} proposed a global information measure based on Kullback–Leibler (KL) divergence. For the given question $q_j$ (with response $\mathcal{D}_j$), the KL divergence between a candidate proficiency level $\theta$ and the true proficiency $\theta_0$ is defined as:
		\begin{align}\label{KL_info}
		&KL_j(\theta \| {\theta}_{0})
		=\mathbb{E}_{y_j}\log\frac{L(\mathcal{D}_j|\theta_0)}{L(\mathcal{D}_j|\theta)} \nonumber \\ 
		=& f(q_j,\theta_0)\log\frac{f(q_j,\theta_0)}{f(q_j,\theta)}+(1-f(q_j,\theta_0))\log\frac{1-f(q_j,\theta_0)}{1-f(q_j,\theta)}.\nonumber
	\end{align}

 The corresponding question selection algorithm integrates KL over a neighborhood of the current estimate $\hat{\theta}^t$:
	\begin{gather}
	\mathcal{I}_j(\hat{\theta}^{t}) = \int_{\hat{\theta}^{t}-\delta}^{\hat{\theta}^{t}+\delta} KL_j(\theta||\hat{\theta}^{t})d\theta.
\end{gather}
where $\delta = 3/\sqrt{t}$. The integration range is wide at the beginning of the test and and gradually narrows as $t$ increases. In MIRT, this extends naturally to a multivariate integral. Essentially, the KL information identifies questions that will \textit{provide the greatest differentiation between the examinee's possible proficiency levels}. Unlike Fisher information, which depends on a single point estimate, KL information measures the discrepancy between two proficiency levels, $\theta$ and $\theta_0$, and remains effective even when they differ significantly. This is why KL information is \textit{global} while Fisher information is \textit{local} \cite{chang2015psychometrics}. Specific examples comparing the two can be found in the appendix.

	\textbf{Advanced Statistical Algorithms.} Numerous works based on Fisher and KL information have been proposed. These methods try to introduce more information in selection to improve the efficiency of proficiency estimation. The Maximum Likelihood Weighted Information \cite{veerkamp1997some} weights the Fisher information by the likelihood function of the examinee's current response results. Its rationale is similar to KL information and aims to improve the local limitations of Fisher information:
	selecting the one that maximizes the integral of the likelihood function times the Fisher $\mathcal{I}_j(\theta)$ over the proficiency level: $\mathcal{I}_j(\hat{\theta}^{t}) =\int_{\hat{\theta}^{t}-\delta}^{\hat{\theta}^{t}+\delta} L(\mathcal{D}_{1:t-1}|\theta) \mathcal{I}_j(\theta)d\theta$, where $L(\mathcal{D}_{1:t-1}|\theta) =\sum_{j=1}^{t-1} L(\mathcal{D}_{j}|\theta) $ is the likelihood function of previous $t$ response. Furthermore, the Maximum Posterior Weighted Information \cite{van1998bayesian,barrada2009metodologia} further weights the Fisher and KL information with an additional posterior probability distribution $P(\theta|\mathcal{D}_{1:t-1})$: $\mathcal{I}_j(\hat{\theta}^{t}) =\int_{\hat{\theta}^{t}-\delta}^{\hat{\theta}^{t}+\delta} P(\theta|\mathcal{D}_{1:t-1})L(\mathcal{D}_{1:t-1}|\theta) \mathcal{I}_j(\theta)d\theta$, where $\mathcal{I}(\theta)$ can be the Fisher Information or KL divergence. Maximum Expected Information \cite{van1998bayesian} accounts for all possible outcomes $y_t$ and their impact on the updated proficiency estimate when weighting Fisher information. Lastly, the theta-Optimization (thOpt) process \cite{barrada2006maximum} selects questions by aligning the maximized information with the current proficiency estimate. \textcolor{black}{Some non-parametric machine learning methods (e.g., decision trees) can also be explored, showing that a small number of questions can match or exceed traditional methods in accuracy, especially under high-dimensional and imbalanced data conditions \cite{zheng2020using}.}

The selection algorithms discussed earlier are based on IRT and do not directly apply to other measurement models. \textcolor{black}{As noted in Section \ref{CDM}, measurement models like CDM represent proficiency $\theta$ as discrete states across different knowledge concepts.} Adaptive selection algorithms under such models is known as Cognitive Diagnosis CAT (CD-CAT) \cite{cheng2009cognitive}. While the selection principles remain similar (maximizing information from selected items), the information measures differ. In CD-CAT, techniques based on KL divergence \cite{henson2005test} and Shannon entropy \cite{tatsuoka2002data} are commonly used. Although continuous traits (as modeled in IRT) and discrete states (as modeled in DINA) describe different aspects of proficiency, they are complementary. This has led to the development of dual-objective CD-CAT methods \cite{kang2017dual,zheng2018information,dai2016exploration} that aim to assess both simultaneously .

		{\textit{Discussion:}}  Selection algorithms in CAT have primarily relied on the above statistical heuristic approaches, which require domain experts to consider every possible testing scenario and manually design corresponding selection algorithms. These methods are \textit{model-specific}, requiring distinct selection algorithms for different measurement models. For example, the above Fisher information \cite{lord2012applications} is specifically crafted for (M)IRT. 
		Consequently, previous statistical methods lack flexibility, and the selection algorithm must be re-designed if the underlying measurement model changes.  
		
		\begin{figure}[t]
		\centering
		\includegraphics[width=0.8\linewidth]{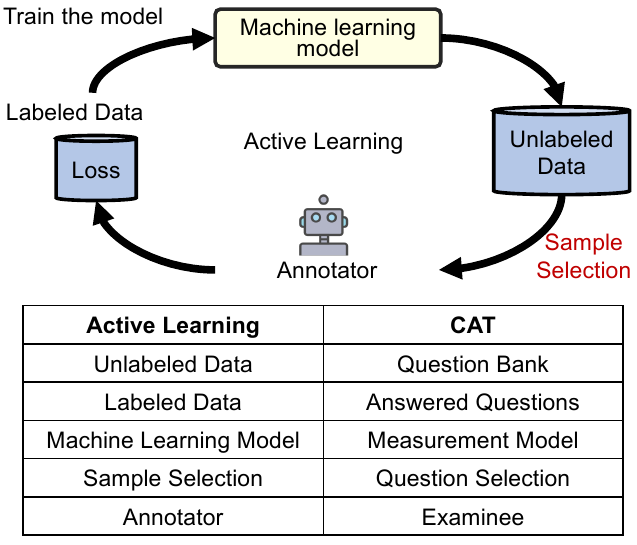}
				\vspace{-8pt}
		\caption{The Active Learning Framework and the relationship/correspondence between each component of Active Learning and those of CAT.}
				\vspace{-8pt}
		\label{active}
	\end{figure}
	
	\subsection{Active Learning Algorithms}\label{ALA}
	
	To design selection algorithms that are effective across different measurement models i.e., \textit{model-agnostic}, researchers explore a general machine learning technique for data selection: \textit{Active Learning} \cite{article07ac}. Active Learning is to actively choose some valuable data, thus can train better models with less data. This technique has improved data efficiency in numerous learning tasks \cite{6747346}.

	As shown in \figurename\ \ref{active}, active learning operates in cycles: where a selection algorithm iteratively chooses unlabeled samples based on the model's current performance and queries a human annotator for labels. This process augments limited labeled data to improve model performance. The core challenge lies in designing effective sample selection algorithms, typically based on two criteria: \textit{informativeness}, selecting samples that reduce model uncertainty \cite{yoo2019learning}, and \textit{representativeness}, selecting samples that reflect the overall data distribution \cite{ghorbani2022data,10372131}, or a combination of both \cite{wang2020dual}. Active learning shares a similar structure with CAT. Here, the measurement model plays the role of the learning model, the question selection corresponds to sample selection, and examinee responses serve as annotations. The goal is to estimate proficiency using as few questions as possible. This model-agnostic perspective avoids reliance on specific measurement model assumptions. Bi et al. \cite{bi2020quality} propose MAAT, a model-agnostic adaptive testing framework that evaluates the change (analogous to a gradient) in proficiency estimates after each response:
		\begin{equation}
		j_{t+1} = \arg\max_{q_j\in\mathcal{Q}}\mathbb{E}_{y_j} 
		\left\Vert \nabla_{\theta} L(\mathcal{D}_{1:t}\cup \{(q_j,y_j)\}|\theta) \right \Vert.
	\end{equation}
	\textcolor{black}{Since the true responses to candidate questions are not available during selections, it computes the expected gradient norm with respect to the response label $y$ for each candidate question. This expectation quantifies the potential impact of each question on the proficiency estimation.} The intuition behind this framework is that \textit{it prefers questions that are likely to most influence the proficiency estimation (i.e., have the greatest impact on its parameters)}. \textcolor{black}{Notably, this approach places no restriction on the specific type of measurement model, as long as it supports gradient-based optimization.}

		{\textit{Discussion:}} 
		With the rapid development of intelligent testing platforms (ranging from human's online testing systems to AI model's evaluation leaderboards), large-scale examinee response data has been accumulated. However, such data cannot be effectively leveraged by the above rules-based approaches (i.e., statistical algorithms and Active Learning algorithms) \cite{ghosh2021bobcat,li2020data}. In contrast, recent data-driven approaches based on Reinforcement Learning (Section~\ref{rl_method}) and Meta-Learning (Section~\ref{meta}) have gained increasing attention. \textcolor{black}{They automatically learn/optimize effective selection algorithms from large-scale response data without relying on manually defined heuristics or rules, and have demonstrated superior performance.}

	\subsection{Reinforcement Learning Algorithms}\label{rl_method}

	Reinforcement learning (RL), a subfield of machine learning, is a powerful approach that enables an agent to learn how to make optimal decisions automatically \cite{sutton2018reinforcement}. It has been successfully applied in various domains, including robotics, autonomous vehicles, education, and healthcare \cite{arulkumaran2017brief,kober2013reinforcement}. In RL, an agent interacts with an environment and receives feedback in the form of rewards or penalties based on its actions. The goal is to learn a policy $\pi$, which can maximize the long-term cumulative reward. As shown in \figurename\;\ref{rl_cat}, the policy can be learned by exploring the environment and learning from its consequences of actions. In essence, researchers in CAT utilize RL methodologies to address a question: \textit{Can the selection algorithm (policy) be automatically learned and optimized from data or examinee interactions, thus circumventing the necessity for expert intervention?}

	\textbf{Markov Decision Process Formulation.} The interaction between agent and environment can be viewed as a Markov Decision Process (MDP) \cite{feinberg2012handbook}. Specifically, at each step, the agent observes current environment's state ($s$), and interacts with the environment by selecting its actions ($a$). Simultaneously, the agent receives a reward ($r$) from these interactions, influencing or changing the current state of the environment. The objective is to select a best sequence of actions, resulting in the highest cumulative reward ($
	\sum_t r_t$). Therefore, most RL problems are formally described as estimating the optimality of the agent's behavior in a given state (value-based methods \cite{sutton2018reinforcement}) or the optimality of the action policy itself (policy-based methods \cite{agarwal2021theory}) or the hybrid approaches \cite{mnih2016asynchronous}. The overall RL framework for CAT is illustrated in \figurename~\ref{rl_cat}. We formulate the CAT problem as an MDP, where the key RL components in the testing system are defined below:
	\begin{itemize}
		\item \textit{State}: A state $s_t\in \mathcal{S}$ represents the current condition or situation at each test step $t$. It captures relevant information about examinee and the CAT system. Generally, the state includes the examinee's previous response sequence (or a latent vector to represent the current proficiency estimate \cite{li2020deep}) and the candidate questions in the question bank \cite{zhuang2022fully,nurakhmetov2019reinforcement}: $s_t=(\{q_1,y_1,...,q_{t},y_{t}\}, \mathcal{Q})$\footnote{\textcolor{black}{This aligns with the labeled data (answered questions), and the unlabeled data (question bank) in active learning (Section 5.2)}}.

		\item \textit{Action}: An action $a$ refers to the choices that the CAT system can take in current state $s_t$, i.e., the selection of the next question from bank $q_{t+1}\in \mathcal{Q}$.

		\item \textit{Transition}: The transition function is the probability of seeing state $s_{t+1}$ after taking action $q_t$ at current state $s_t$: $P(s_{t+1} | s_t, q_{t+1})$. At each step, the uncertainty comes from the examinee's response correctness label $y_{t+1}$ to question $q_{t+1}$.

		\item \textit{Reward}: A reward $r$ is a scalar feedback that the CAT receives after selecting a question for the examinee. To achieve CAT's goal in Definition \ref{def1}, {the reward function can be defined as the accuracy of proficiency estimation at each step}\footnote{As the true value is often unobtainable, it is commonly derived through simulation experiments.}\cite{ma2023novel,nurakhmetov2019reinforcement,wang2023gmocat}, i.e., $\Vert\hat{\theta}^t-\theta_0\Vert$, \textcolor{black}{or the performance prediction loss of $\hat{\theta}^t$ on the held-out response data $\mathcal{D}$ \cite{zhuang2022fully,shin2022building}, i.e., $L(\mathcal{D}|\hat{\theta}^t)$.} This reward signal is pivotal in guiding the policy $\pi$ to select the best-fitting question that can reduce the estimation error.

	\end{itemize}

	\begin{figure}[t]
		\centering
		\includegraphics[width=0.84\linewidth]{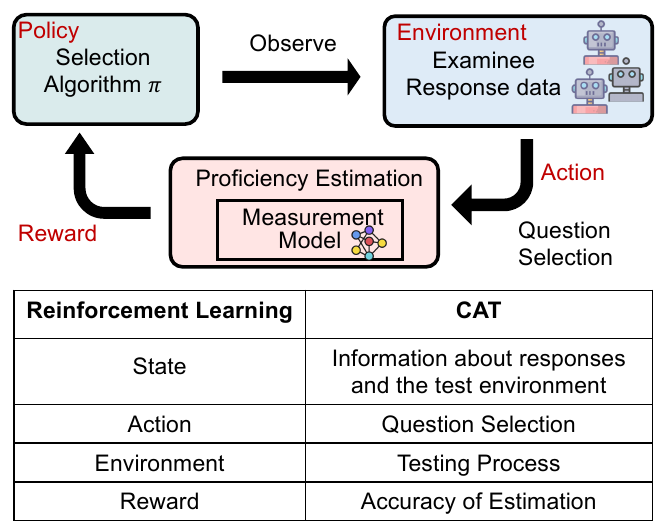}
				\vspace{-6pt}
		\caption{The overall Reinforcement Learning framework of CAT. The objective is to optimize the selection algorithm $\pi$ (i.e., policy) by exploring the large-scale examinee response data (i.e., environment).}
				\vspace{-6pt}
		\label{rl_cat}
	\end{figure}

	Recently, with the advancements in deep learning, an increasing number of studies are leveraging Deep Reinforcement Learning to tackle the MDP problems in CAT. Li et al. \cite{li2023deep} utilize the Deep Q-Network to represent the action-value function $Q_w(s,q)$, representing the value of choosing question $q$ in state $s$, and $w$ denotes its parameter of the network's fully connected layer. The most suitable question is selected according to the policy: 
	\begin{equation}
		\pi^*(q|s)=\arg\max_{q\in \mathcal{Q}} {Q}_w(s,q).
	\end{equation} 
To further capture the complex interactions between examinees and questions in practical testing scenarios, a Transformer-based Q-Network named NCAT \cite{zhuang2022fully} has been proposed. \textcolor{black}{NCAT incorporates multiple functional modules, including a Double-Channel Performance Learning module that independently captures diverse aspects of examinee performance, and a Contradiction Learning module that identifies and extracts inconsistencies in examinee behavior, such as guessing and slipping.}

	\textbf{Stochastic Shortest Path Formulation.} Furthermore, CAT can be defined as a Stochastic Shortest Path (SSP) problem \cite{bertsekas1991analysis}, which is a special case of MDP. In an SSP, the objective is to find the shortest path (i.e., the minimum test step) from a given initial state $s_0$ to goal states. In CAT, the goal state typically represents the completion of the test or the attainment of a predetermined level of proficiency estimation precision. Gilavert et al. \cite{gilavert2022computerized} use Linear Programming to find the optimal testing policy $\pi^*$, treating CAT like a flow network where each state must have balanced inflow and outflow (except for the start and end points). It denotes variables $x_{s,a}$ as the expected accumulated occurrence frequency for every pair (state $s \in \mathcal{S}$, qu $q \in \mathcal{Q}$), and equalizes $in(s)$ and $out(s)$ flow model for every state $s$. The flow into a state $s$ is the sum of the expected frequencies of all actions in all other states $s'$ that lead to $s$: ${in}(s) = \sum_{s', q} x_{s',q}P(s|s', q)$. The flow out of a state $s$ is the sum of the expected frequencies of all actions in state $s$: ${out}(s) = \sum_{q} x_{s,q}$. The objective function is to maximize the total expected reward $r$, which is the sum of the expected frequencies times the immediate rewards $r(s,q)$ for all state-action pairs: $\min_{x_{s,q}} \sum_{s\in \mathcal{S}, q\in \mathcal{Q} }{x_{s,q}r(s,q)}$. Thus the optimal question selection policy $\pi^*$ can be obtained by:
	\begin{equation}
		\pi^*(q|s) = \frac{x_{s,q}}{\sum_{q'\in \mathcal{Q}}x_{s,q'}}.
	\end{equation}

	\textbf{Partial-Observable MDP Formulation.} Partial-Observable MDP (POMDP) extends the standard MDP framework to settings where the environment is only partially observable \cite{6616533}. Traditional CAT models often assume that an examinee's proficiency can be fully inferred from previous responses, allowing it to be treated as an MDP with the proficiency estimate as the state. However, in practice, proficiency cannot be perfectly inferred due to some inherent uncertainty \cite{zhuang2022robust,drousiotis2021capturing}.

	To this end, many works \cite{nurakhmetov2019reinforcement,gilavert2022computerized,chen2018recommendation} model CAT as a POMDP. Compared with MDP, the POMDP model has two additional elements. $O$: A set of observations; $Z$: Observation probabilities. $Z(o|s',q)$ is the probability of making observation $o$ after selecting question $q$ and transitioning to state $s'$. While the underlying state (proficiency) remains, it is not fully observable. Instead, these methods maintains a belief state \(b(s)\), a probability distribution over possible proficiencies, which is updated via Bayes' rule: When the agent select action (question) $q$ in belief state $b$ and makes observation $o$, it updates its belief state to $b'(s')$: $b'(s') = \eta{Z(o|s',q) \sum_{s \in \mathcal{S}} P(s'|s,q) b(s)}$. where $\eta$ is a normalizing constant. POMDPs can be solved by many algorithms, such as Grid-based algorithms \cite{hoerger2021line}, Monte Carlo tree search \cite{schwartz2022online}. However, due to its partial observability of the environment, POMDPs are more challenging to solve than MDPs.

	\subsection{Meta Learning Algorithms}\label{meta}
	Another data-driven machine learning approach that can address this complex CAT problem is meta-learning \cite{finn2017model}: It involves training a model on various tasks to acquire cross-task knowledge or learn how to learn efficiently. Specifically, the {base-learner} is trained across a variety of related tasks, allowing it to gather \textit{cross-task insights} and \textit{general knowledge} about how to learn efficiently. Then, the {meta-learner} leverages this knowledge to swiftly adapt to new, unseen tasks \cite{9238451}. In CAT, each examinee's testing process can be seen as a \textit{task} because it involves selecting appropriate test questions based on the proficiency level. The selection algorithm can be regarded as a form of \textit{general knowledge} because it represents the accumulated knowledge and experience gained from a diverse set of examinees (\figurename\ \ref{bobcat}). This knowledge can include the best policy for question selection, information about the characteristics of different test questions, the examinee proficiency prior, etc. By learning from these diverse examinees (tasks) in the large-scale response dataset, it can acquire a good question selection that can adapt to individual examinees.

	\begin{figure}[t]
		\centering
		\includegraphics[width=0.9\linewidth]{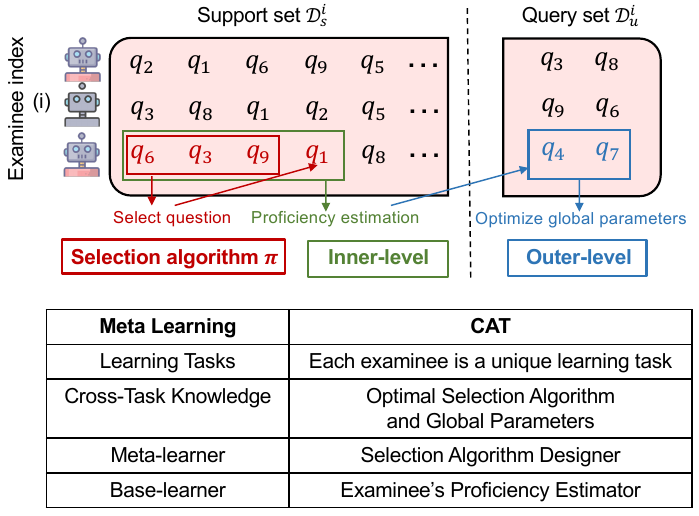}
		\vspace{-2pt}
		\caption{The overall Meta Learning framework of CAT, and this figure is adapted from \cite{ghosh2021bobcat}. The objective is to optimize the selection algorithm $\pi$ by exploring the large-scale examinee response data.}
		\vspace{-7pt}
		\label{bobcat}
	\end{figure}

	\textbf{Bi-Level Optimization.} Bi-Level optimization is a classical meta-learning approach commonly applied in CAT. It decomposes the learning process into two nested levels: an inner level that adapts to individual examinees and an outer level that learn general knowledge. Ghosh et al. \cite{ghosh2021bobcat} propose a bi-level optimization framework for CAT (BOBCAT) to directly learn the data-driven selection algorithm $\pi$. Specifically: let $N$ denote the number of examinees in the response dataset for training $\pi$. The responses of each examinee $i$ are randomly divided into a support set $\mathcal{D}_s^i$ and a query set $\mathcal{D}_u^i$, where $\pi$ sequentially select a total of $t$ questions $\{q_{1} ,...,q_{t}\}$ from $\mathcal{D}_s^i$, observe their responses, and predict their response on the held-out query set $\mathcal{D}_{u}^i$. The global knowledge (i.e., selection algorithm $\pi$ and global parameters $\gamma$) is redefined as the objective of bi-level optimization:
	\begin{align}
		&\min_{\pi, \gamma}\frac{1}{N}\sum_{i=1}^{N}{\sum_{(q,y)\in\mathcal{D}_{u}^i}{\ell(y, f(q, \hat{\theta}_i))}}, \label{outer}\\
		&\mathrm{ s.t. } \;\; \hat{\theta}_i=\mathop{\arg\min}_{\theta_i}{\sum_{(q,y)\in \mathcal{D}_s^{i}}{\ell\left(y,f\left(q, \theta_i\right)\right)}},\label{inner1}\\
		&  \mathrm{where} \quad q_{t+1} \sim \pi\left(q| q_{1},y_{i(1)},...,q_{t},y_{i(t)}\right) \in \mathcal{D}_s^{i}.
	\end{align}
	\figurename\ \ref{bobcat} shows the overall meta learning framework. In the \emph{inner-level} (Eq.(\ref{inner1})), the question in the support set $\mathcal{D}_s^{i}$ for examinee $i$ is sequentially selected by $\pi$, according to the previous responses; then binary cross-entropy loss $\ell(\cdot)$ on $\mathcal{D}_s^{i}$ is minimized for estimating the proficiency $\hat{\theta}_i$ for the outer-level. In the \emph{outer-level} (Eq.(\ref{outer})), the loss of the estimate $\hat{\theta}_i$ on the query set $\mathcal{D}_{u}^i$ is minimized to learn the selection algorithm $\pi$ and the global parameters $\gamma$ (e.g., question characteristics). The algorithm $\pi$ is also model-agnostic. It could be adapted to the given measurement model ($f$) automatically by optimizing this problem for efficient selection.
	
	\textcolor{black}{Through large-scale sampling and training, this framework learns to estimate and quantify the value of each question for different examinees and under varying contexts. Even for questions whose IDs do not appear in the training set, their value can be inferred from their characteristics via \(\gamma\). }Once the question selection algorithm is trained, its parameters do not update during the CAT process and adaptively select the next question based on previous response behaviors.

	Based on BOBCAT, there have been increasing efforts to improve upon it. Ma et al. \cite{ma2023novel} propose a flexible optimization framework Decoupled Learning CAT (DL-CAT). The original BOBCAT obtains the parameters of two modules (i.e., examinee proficiency estimation and question selection algorithm) through coupled inner and outer optimizations, i.e., the result of the outer optimization model is used to measure the quality of the inner. DL-CAT  devises a ground-truth construction strategy, and a pairwise loss function, allowing these two models to be trained independently; Feng et al. \cite{feng2023balancing} introduces a constrained version of BOBCAT to address the question exposure and test overlap issues. Yu et al. \cite{yu2023sacat} recently introduce the collaborative information of examinees in optimizing this bi-level problem, achieving fast convergence of proficiency estimation.

	\textbf{Meta-Learning vs Reinforcement Learning.} In CAT, meta-learning methods can be seen as a higher-level learning process that learns how to adapt a general strategy for question selection to specific examinees based on their responses. Actually, it can be reframed as an RL problem. Zhuang et al. \cite{zhuang2022fully} propose NCAT to transform the meta-learning problem in CAT into an RL problem. Because the test may stop at any step according to different stopping rules, NCAT simplifies the original objective (Eq(\ref{outer})) and sums all the steps to minimize the loss:
	\begin{eqnarray}
		&&\min_{\pi} \frac{1}{N}\sum_{i=1}^{N}\sum_{t=1}^T{\sum_{(q,y)\in\mathcal{D}_{u}^i}{\ell(y,f(q,\hat{\theta}_i^t))}} \nonumber \\
		&\triangleq&\mathop{\max}\limits_{\pi} \mathbb{E}_{i\sim\pi}\left[\sum_{t=1}^T{-{\sum_{(q,y)\in\mathcal{D}_{u}^i}{\ell(y,f(q,\hat{\theta}_i^t))}}}\right] \nonumber\\
		&=&\max_{\pi}\mathbb{E}_{i\sim \pi}\left[\sum_{t=1}^T{-L(\mathcal{D}_{u}^i|{\hat{\theta}_i^t})} \right], \label{target_rl}
	\end{eqnarray}
	where $\hat{\theta}_i^t=\mathop{\arg\min}_{\theta_i}{\sum_{(q,y)\in \mathcal{D}_s^{i(t)}}{\ell\left(y,f\left(q, \theta_i\right)\right)}}$ and $\mathcal{D}_s^{i(t)}=\{q_{1},y_{i(1)},...,q_{t},y_{i(t)}\}$. Thus, the bi-level optimization is transformed into maximizing the expected
	cumulative reward (i.e., $-L(\mathcal{D}_{u}^i|{\hat{\theta}_i^t})$) in RL settings, where the reward is the negative loss of the estimated proficiency of examinee $i$ on the query set at step $t$. Recently, GMOCAT \cite{wang2023gmocat} has been proposed as a Multi-Objective RL framework. \textcolor{black}{GMOCAT uses Graph Neural Networks to capture the complex relationships between questions and skills. It adopts an Actor-Critic architecture and incorporates three objectives into the reward function: (1) improving prediction accuracy, (2) enhancing concept (skill) diversity, and (3) reducing question exposure.}

		{\textit{Discussion:}} The aforementioned data-driven machine learning approaches, i.e., Reinforcement Learning and Meta Learning, are capable of uncovering latent patterns and correlations from data, and directly optimizing question selection policies. By fitting to large-scale data, they can approximate the ultimate goal of CAT. \textcolor{black}{However, potential issues such as data bias, model overfitting, and high training overhead should not be overlooked.}

		\begin{figure}[t]
		
		\centering
		
		\includegraphics[width=0.85\linewidth]{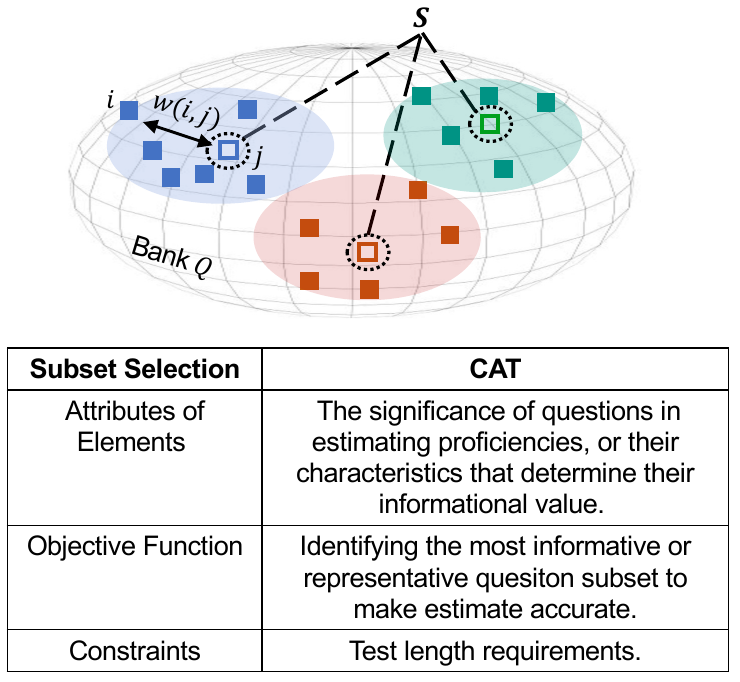}
				\vspace{-5pt}
		\caption{Illustration of the subset optimization problem, adapted from \cite{zhuang2023bounded}: Selecting subset $S$ to cover the bank $Q$. Rectangles represent different questions, with $w(i,j)$ measuring the similarity of question pair.}
		\label{becat}
				\vspace{-18pt}
	\end{figure}

	\subsection{Subset Selection Algorithms}\label{SSA}
	
	The ultimate objective of CAT is to measure examinees' abilities both efficiently and accurately. Specifically, as illustrated in Definition \ref{def1}, the goal is to find a subset $S$ of $T$ questions from question bank $\mathcal{Q}$, so that the final proficiency estimate $\hat{\theta}^T$ can approach the true proficiency $\theta_0$:
	\begin{equation}
		\mathop{\mathrm{min}}\limits_{|S|=T}	\Vert {\hat{\theta}^T}-\theta_0 \Vert,
	\end{equation}
	where $\hat{\theta}^T=\arg\min_{\theta}{ \sum_{(q,y)\in S}{\ell(y,f(q,\theta))}}$ is the final proficiency estimate when the test ends with the corresponding $T$ responses. In contrast to previous sequential selection methods, it \textit{essentially doesn't require perfect selection at each step, but rather emphasizes the accuracy of the final estimate.}
	
	From a global perspective, CAT essentially is a Subset Selection problem \cite{miller2002subset}, a fundamental challenge in machine learning and optimization. It revolves around choosing a subset of elements $S$ from a larger set $\mathcal{Q}$ that optimizes a particular objective function $F(S)$ while adhering to specific constraints.  However, we cannot directly solve the above optimization problem due to the following main challenge: \textit{The true proficiency of the examinee, denoted by $\theta_0$, is unknown.} It is not available in the dataset, which prevents us from directly optimizing or designing the question selection algorithm. To address this issue, some researchers have developed heuristic methods. For example, Mujtaba et al. \cite{mujtaba2021multi} use the standard error of measurement as the objective \( F(S) \), which provides a measure of confidence in an estimate from a test. At each step, it uses multi-objective evolutionary algorithms to obtain the set of Pareto-optimal solutions \cite{deb2011multi} by maximizing precision and minimizing the number of questions. Recently, for AI model evaluation, clustering techniques (e.g., K-means) have been used to select representative subsets \( S \) from benchmarks \cite{polo2024tinybenchmarks}.

To develop a more general and scalable CAT framework, Zhuang et al. \cite{zhuang2023bounded} propose BECAT, which reformulates the question selection problem in a data summary manner. Since the true proficiency \(\theta_0\) is unobservable, they approximate it using \(\theta^*\): the proficiency estimated from an examinee's full responses to the entire question bank \(\mathcal{Q}\), i.e., \(\theta^* \approx \theta_0\). This approximation enables the selection algorithm to target \(\theta^*\) instead of the unknown \(\theta_0\): \textit{Select a subset of questions \(S \subseteq \mathcal{Q}\) such that the estimated proficiency based on \(S\) closely approximates \(\theta^*\)} (i.e., the estimate that would be obtained if optimizing on the full responses to $\mathcal{Q}$).
	\begin{align}\label{opt_alg}
	&	\min_{|S|=T}	\Vert {\hat{\theta}^T}-\theta_0 \Vert \Rightarrow \min_{|S|=T}	\Vert {\hat{\theta}^T}-\theta^* \Vert  \nonumber  \\
		\Rightarrow &\min_{|S|=T} \max_{\theta\in \Theta} \Big\Vert \sum_{(q,y)\in S}{\gamma \nabla \ell(y,f(q,\theta))} - \sum_{(q,y)\in \mathcal{Q}}{\nabla \ell(y,f(q,\theta))} \Big\Vert \nonumber  \\
		\Rightarrow
			&\mathop{\mathrm{min}}\limits_{|S|=T} \mathop{\mathrm{max}}\limits_{\theta\in \Theta}  \sum_{i\in \mathcal{Q}} \mathrm{min}_{j\in S}\Vert \nabla \ell_i(\theta)-\nabla \ell_j(\theta) \Vert  \nonumber\\
			\Rightarrow &
		\mathop{\mathrm{max}}\limits_{|S|=T} \sum_{i\in \mathcal{Q}}\mathrm{max}_{j\in S} \; w(i,j),
	\end{align}
	where $w(i,j)\triangleq d-\mathrm{max}_{\theta\in \Theta} {\Vert \nabla \ell_i(\theta) - \nabla \ell_{j}(\theta)\Vert}$ is the gradient similarity between question pair $(q_i,q_j)$ for this examinee, thus the objective function  $F(S)= \sum_{i\in \mathcal{Q}}\mathrm{max}_{j\in S} \; w(i,j)$. The core of BECAT's subset selection algorithm is to find a subset $S$ of size $T$ that maximizes the coverage of $ \mathcal{Q}$, quantified by the similarity measure $w(i,j)$. This approach (\figurename\;\ref{becat}) essentially seeks the most representative questions, aligning with prior selection algorithms but under a new, more rigorous theoretical framework.

	Given the NP-Hard nature of this optimization, BECAT employs a submodular function approximation. A simple greedy algorithm can finally solve this subset selection problem, with BECAT ensuring that the estimate error remains upper-bounded at each step. The subset selection problem in CAT is a fresh direction with significant potential. This method offers a universal framework for question selection, applicable across various complex measurement models that can utilize gradient-based estimations, including neural network models.

	\begin{table*}[t]
		\renewcommand{\arraystretch}{1.2}
		\centering
		\caption{Comparison of Different Question Selection Algorithms in CAT}
		\label{tab:selection-algorithms-cat}
		\resizebox{\textwidth}{!}{%
			\begin{tabular}{|m{2.1cm}|m{1.3cm}|m{2.1cm}|m{1.9cm}|m{4cm}|m{4cm}|}
				\hline
				\textbf{Category} & \textbf{Generality} & \textbf{Interpretability} &  \textbf{Need Training} & \textbf{Advantages} & \textbf{Disadvantages} \\ \hline \hline
				Statistical\ \;\quad \quad Algorithms &\centering \XSolidBrush  & \centering \Checkmark & \centering \XSolidBrush & Simple implementation and efficient operation & Dependent on IRTs and requires expert knowledge for design \\ \hline
				Active Learning & \centering \Checkmark & \centering\Checkmark  & \centering \XSolidBrush & Model-agnostic and flexible & Neglect the nuanced information within measurement model parameters \\ \hline
				Reinforcement Learning & \centering\Checkmark & \centering\XSolidBrush & \centering \Checkmark & Automatic generation of selection algorithm; Sequential Decision Making & Incurs additional training costs and potential bias from data-driven selection \\ \hline
				Meta Learning & \centering\Checkmark & \centering\XSolidBrush  & \centering \Checkmark& Automatic generation of selection algorithm; Fast Adaptation & Incurs additional training costs and potential bias from data-driven selection \\ \hline
				Subset Selection &\centering \Checkmark & \centering\Checkmark  & \centering \XSolidBrush &  Strong theoretical guarantees for estimation accuracy & Faces challenges in the initial stages of CAT \\ \hline
			\end{tabular}%
		}
	\end{table*}

		{\textit{Discussion:}} It is noteworthy that, despite the superior performance demonstrated by the latest machine learning and deep learning approaches \cite{yu2024a}, \textcolor{black}{they \textit{have not yet replaced traditional statistical approaches in practice}. Particularly in testing scenarios that prioritize interpretability or efficiency, statistical methods remain predominant}. In the Appendix, we compare these five categories of selection algorithms in CAT systems, highlighting the generality and interpretability of each category, along with their main advantages and limitations. This overview assists researchers in identifying the most suitable algorithm for their CAT applications, balancing efficiency and complexity.

	\section{Question Bank Construction}\label{QBC}
	To develop a high-quality CAT, the foundational step is to construct a high-quality question bank. The bank construction can be decomposed into two main stages: Question Characteristics Analysis and Question Bank Development: (1) Question Characteristics Analysis first detailedly examines the properties and attributes of potential questions. Then, (2) Question Bank Development assembles the final question bank $\mathcal{Q}$ from the analyzed questions.

	\subsection{Question Characteristics Analysis}\label{qca}
	
	The first stage, question characteristics analysis, involves a detailed examination of the properties and attributes of potential questions, e.g., difficulty, discrimination, and the knowledge concepts required to answer the question. For example, when selecting questions based on Fisher Information, one must leverage pre-calibrated parameters like difficulty ($\beta_j$), discrimination ($\alpha_j$), and guessing factor ($c_j$), alongside the current proficiency estimate, to compute the Information value $\mathcal{I}_j(\theta)$ for each question $j$. The methods of characteristics analysis can be categorized into three main approaches: expert-based, statistic-based, and deep learning-based methods.

\textit{Expert-based Characteristics Annotation.} In expert-based annotation, domain experts assess question parameters, as seen in online CAT systems like SIETTE \cite{conejo2004siette} and GenTAI \cite{lopez2006adaptive}. Effective expert estimation often involves structured questionnaires \cite{lopez2008helping}, followed by discussions to resolve divergent opinions. Results are aggregated using averages for continuous attributes or voting for discrete ones \cite{kozierkiewicz2014item}. Expert judgments can be subjective, leading to potential inaccuracies, especially with limited or inconsistent expert input. With the advancement of generative AI, LLMs can also be used to annotate question characteristics \cite{liu2024leveraging}.

	\textit{Statistic-based Characteristics Annotation.} The statistic-based method for annotating question characteristics requires gathering responses from a large group of examinees. It is resource-intensive nature and involves pre-testing with examinees \cite{de2010primer}. In Classic Test Theory, question difficulty is calculated as the proportion of correct responses within examinees \cite{magno2009demonstrating,devellis2006classical}, while discrimination is derived from performance disparities between higher and lower ability examinees \cite{chang2009applying}. The Q-matrix is another crucial characteristic of questions. It is a binary matrix that indicates which knowledge concepts are associated with a question. Numerous researchers have attempted to employ some parameter estimation approaches  (e.g., maximum likelihood estimation and Bayesian estimation), to learn these characteristic parameters from response data \cite{liu2013theory,sun2014alternating,xiong2022data}.

	\textit{Deep Learning-based Characteristics Annotation.} With the rise of Natural Language Processing (NLP), there has been an increasing trend in recent years to directly use the textual information of questions to analyze various attributes. For difficulty prediction, attention-based CNN models and domain adaptation strategies have been used to evaluate reading questions and medical question complexity \cite{huang2017question, qiu2019question, huang2021stan}. For knowledge concept (Q-matrix) prediction, which typically exhibits a hierarchical structure, a Hierarchical attention-based Recurrent Neural Network has been proposed \cite{huang2019hierarchical,huang2022hmcnet}. Lei et al. \cite{lei2021consistency} further take into account the multi-modal features of questions, such as images and formulas. Pre-trained NLP models have also proven effective for automated question analysis \cite{yin2019quesnet, ning2023towards}.

	\subsection{Question Bank Development}\label{QBD}

	The second stage, question bank development, involves the actual assembly of the question bank from the analyzed questions of the first stage. This process should aim to create a balanced and varied bank that can cater to different levels of proficiency and different areas of knowledge \cite{revuelta1998comparison,reckase2010designing,segall2005computerized}. According to different scenarios, the approaches to developing a question bank can be categorized into the following three aspects.

	\textit{Question Bank Blueprint Design.} The goal of the blueprint design is to create an optimal framework for a bank, outlining the distribution of questions based on various attributes. Reckase et al. \cite{reckase2010designing} analyze the characteristics of an optimal question bank in a CAT system using a 1PL-IRT model with a maximum Fisher information selection algorithm. They propose the bin-and-union method to allow a maximum deviation \(r\) between optimal difficulty and estimated proficiency, extending these methods for large-scale CAT systems and continuous new question pretesting \cite{he2014item, van2000integer}.

	\textit{Question Bank Assembly.} While the blueprint design focuses on creating an optimal framework, the assembly process involves generating question banks from an existing master bank according to specific requirements. Way et al. \cite{way2005developing} discussed the development and maintenance of a master bank, including constraints to ensure the assembled question bank meets desired specifications. A mixed-integer programming \cite{van2006assembling} was proposed to create a bank that satisfies content specifications and maximizes information at selected proficiency values. 
	

	\textit{Question Bank Rotating.} Rotating the question bank involves dividing a master bank into smaller banks with overlapping elements, ensuring balanced exposure rates \cite{stocking1998optimal, ariel2004constructing, swanson1993model}. Ariel et al. \cite{ariel2004constructing} proposed dividing a master bank into smaller banks using Gulliksen's matched random subtests method \cite{gulliksen2013theory} to prevent over- or underexposure. The Weighted Deviation Model \cite{swanson1993model} manages the degree of overlap, maintaining representativeness and preventing question overexposure.

{\textit{Discussion:}}
Think of the entire bank development process as creating and managing a library. The blueprint design is like the architectural plan for the library, defining where each section (e.g., fiction, non-fiction) will be located; The assembly process resembles acquiring books from suppliers based on specific demands; Rotating the question bank is similar to periodically rotating the books on display. Even though the library has a vast collection, only a subset is displayed prominently at any given time. This rotation ensures that different books get exposure, and library visitors encounter a variety of books over time. {\color{black}{Although this section has so far focused on classical methods, the bank construction pipeline can also incorporate LLMs as auxiliary components. In other words, LLMs can be integrated into the bank development stage to improve scalability and reduce manual cost, while the psychometric principles of CAT remain unchanged.}}
{\color{black}{In this analogy, LLMs (or agents) can be viewed as ``librarians'' that help scale and accelerate curation: they can draft candidate items on demand under explicit constraints, and generate useful metadata (topic tags, expected solution outlines, and common error patterns) that supports indexing and retrieval.}}
{The construction of a high-quality question bank introduced in this section is not just a prerequisite for CAT, but also a continuous process.} It requires regular updates and refinements to ensure the relevance and effectiveness of the adaptive testing system.

	\section{Test Control of CAT} \label{AOC}
	When implementing a testing system, in addition to considering the three components mentioned above, several key factors need to be taken into consideration, such as exposure control, fairness, robustness, and search efficiency.

	\subsection{\textcolor{black}{Exposure Control}}\label{CC}
	 \textcolor{black}{Exposure control aim to balance the frequency of each question's use from the question bank}. \textcolor{black}{Proper exposure control can help mitigate the risk of overexposure of questions}, minimize question waste, and maximize test coverage. Two popular strategies for exposure control are the Sympson-Hetter method \cite{sympson1985controlling} and the A-Stratified method \cite{chang1999stratified}: (1) Sympson-Hetter Method manages question exposure rates using conditional probabilities. It doesn't assign a selected question to the examinee immediately; instead, it passes through a probability filter. The actual chance a question is given to an examinee depends on both its selection likelihood \textcolor{black}{and a exposure control parameter}, keeping question exposure within acceptable limits. However, this method may not effectively increase the usage rate of low-exposure questions. Enhancements to this method have been developed to address these limitations \cite{van2004constraining,van2007conditional,barrada2009multiple}; (2) A-Stratified Method and its subsequent researches \cite{chang2001stratified,barrada2006maximum,barrada2014optimal} are designed to counteract selection biases of algorithms favoring certain questions (e.g., Fisher Information prefers highly differentiated questions). On the other hand, numerous studies \cite{bi2020quality,wang2023gmocat,ocal15effect} have attempted to incorporate the coverage of knowledge concepts as a criterion in question selection, aiming to make the assessment more comprehensive. 
	
	This factor is crucial for both humans and AI. When students are familiar with exam questions beforehand, the test results lose credibility. Similarly, for AI model evaluations, it has been observed that benchmarks released before the creation date of LLM's training data generally perform better than those released afterward \cite{li2024task}. Increasingly, the AI evaluation is being questioned regarding data contamination \cite{oren2023proving}. Therefore, controlling question exposure rates is a necessary measure to improve the reliability of assessments.

\subsection{Fairness}\label{Fairness}
Fairness is a topic of profound societal significance in both education and machine learning research fields, sparking numerous discussions and leading to the development of many fairness-aware learning algorithms \cite{Cleary1968TESTBP,Chai2022FairnessWA,Li2022AchievingFA}. As a technology with potential applications in high-stakes testing, fairness in CAT is a paramount concern. In CAT, the bias that leads to fairness issues can be introduced through three components:

\begin{itemize}
	
	\item \textit{Bias in Measurement Models.} Biases in measurement models may stem from the skewed training data, which could reflect the underrepresentation of certain groups or pre-existing educational disparities \cite{liu2022learning,Thompson2022Is,kizilcec2022algorithmic}. Such biases can lead to an inaccurate and biased estimation of an examinee's proficiency $\theta$, resulting in unfair outcomes.
{\color{black}{A practical mitigation is to evaluate calibration and model fit across subpopulations (e.g., invariance checks) and apply multi-group calibration when needed. Fairness-aware calibration objectives can also be used as a light regularizer to reduce spurious group effects.}}
	
	\item \textit{Bias in Question Bank.}  The question bank may contain biases if questions are not equally applicable or relatable to all examinees, potentially disadvantaging certain groups \cite{camilli1994methods,hambleton1991fundamentals}. For example, some questions in NAPLAN have been deemed unfair for rural examinees, as these questions don't relate to their real-life experiences \cite{Roberts2017Stan}. Various methods have been proposed to detect this type of bias \cite{chu2013detecting,camilli1994methods,mellenbergh1989item}. 
	{\color{black}{For example, mitigation often follows an audit--repair loop: DIF analyses flag potentially biased items, which are then revised, replaced, or retired. This is usually paired with expert review to separate unintended context bias from construct-relevant differences.}}
	
	\item \textit{Bias in Selection Algorithms.}  Selection algorithm can introduce bias since every algorithm has its own ``selection preferences''. For example, the Maximum Fisher Information tends to select questions with high discrimination \cite{lord2012applications}. If such questions unexpectedly correlate with specialized knowledge known only to a specific group, bias may ensue.
	
\end{itemize}

Concerns about fairness in CAT also stem from the fact that examinees answer different questions \cite{green1984technical}. Equating, a technique used to ensure score equivalence across different tests, is commonly employed to address such concerns \cite{brigman1976multiple}. Many further studies about equating scores have been conducted \cite{van2000test,jansubpopulation,sawaki2001comparability}.  In real-world tests such as the GRE, equating has been used to standardize scores and percentiles, taking into account the difficulty of the questions answered. This process ensures that scores can be compared fairly across different examinees worldwide.
{\color{black}{In practice, equating is complemented by routine drift checks and periodic DIF re-audits, especially when new items are added or rotated. This helps preserve comparability as the bank evolves.}}

	\subsection{Robustness}\label{Robustness}
	Noise in CAT can impact the precision of the estimated proficiency of an examinee, leading to potential errors in score interpretation. In CAT, noise usually refers to the random variability or measurement error that can affect the accuracy of estimation. It can arise from various sources such as test administration conditions, examinee behavior, or question characteristics. \textcolor{black}{For example, an examinee may be distracted by environmental noise during the test, leading to an incorrect response that does not reflect their true ability. Alternatively, a poorly worded or ambiguous question may confuse examinees, introducing unintended variability in responses.}

	To mitigate the effects of noise in CAT, a robustness factor is introduced to help stabilize the estimation of proficiency by incorporating additional information, thereby counteracting the impact of noise and improving the reliability \cite{baker2004item}. In machine learning, various robustness techniques are employed to enhance the performance of models in the presence of noise, such as regularization methods \cite{li2021improved}, data augmentation \cite{wang2021regularizing}, adversarial methods \cite{dong2020benchmarking}, ensemble methods \cite{kariyappa2019improving}. In the CAT testing process, significant sources of noise such as guess and slip factors made by examinees, introduce uncertainty. For example, an examinee's proficiency level may not be uniquely determined by their responses, as they may solve a particular question correctly using different knowledge concepts or even by guessing.  The presence of noise and uncertainty poses a significant challenge to the robustness of CAT systems. Veldkamp et al. \cite{veldkamp2019robust} consider the uncertainty in question parameters during the selection process. More recently, ensemble learning has been explored to combine multiple potential estimates at each step, thereby enhancing proficiency estimation \cite{zhuang2022robust}.

	\subsection{Search Efficiency} \label{SE}
	In large-scale educational testing, efficient question selection is a critical challenge. Traditional selection algorithms often evaluate all candidate questions in a brute-force manner, resulting in \textit{a linear time complexity of \( O(|\mathcal{Q}|)\)}, where \( \mathcal{Q} \) is the question bank. This becomes a computational bottleneck in intelligent testing systems. To mitigate this, some organizations like GMAT \cite{rudner2009implementing} rely on manual filtering rules crafted by experts, which is labor-intensive and lacks scalability. Recent research has explored two main directions to improve efficiency:
	\begin{itemize}
	\item Heuristic Search via PSO: Particle Swarm Optimization (PSO) has been applied in IRT-based adaptive testing \cite{huang2009adaptive,8304801}. PSO enables \textit{parallel exploration} of the search space, where each particle represents a candidate question. This parallelism accelerates convergence toward optimal selections, reducing computational burden.

\item Tree-Based Indexing: Inspired by recommendation systems and information retrieval, efficient search structures such as balanced trees have been adopted \cite{zhu2018learning,9674822}. Hong et al. \cite{hong2023search} propose a Search-Efficient CAT framework that employs examinee-aware space partitioning to construct a tree-based index. This method significantly narrows the search space and avoids redundant computations across testing rounds, reducing the search complexity from $O(|\mathcal{Q}|)$ to $O(\log |\mathcal{Q}|)$.

	\end{itemize}

			{\textit{Discussion:}}
	While accuracy and efficiency stand as primary objectives, these factors hold significant importance for practical settings, especially in high-stakes testing scenarios (e.g., competitive or selective examinations). However, consideration of these factors may inevitably reduce accuracy. For example, when considering the additional fairness to ensure equity among different groups, it might be necessary to deviate from the optimal trajectory of a well-trained selection algorithm. Thus, CAT poses a multidimensional decision-making challenge, necessitating the consideration of various factors at the same time using diverse machine learning techniques. In the Appendix, we show the underlying causes and advantages of different factors in CAT test control.

	\section{Evaluation}\label{EM}
	

	Various metrics have been developed to assess the performance of CAT methods, such as correlation coefficients, bias, and measurement error \cite{crocker1986introduction,van2010elements}. This section introduces two of the most extensively utilized evaluation methods: simulation of proficiency estimation and examinee score prediction.

	
	{\textbf{Simulation of Proficiency Estimation.}}The simulation of ability estimation is a foundational evaluation technique in CAT \cite{vie2017review}. Since true proficiency (\(\theta_0\)) is unobservable, we simulate it by sampling a set of values \(\{\theta_0^1, \theta_0^2, ..., \theta_0^N\}\) to represent a virtual group of examinees. This approach enables us to further emulate the interactions between examinees (with these proficiencies) and any question from the question bank, utilizing measurement models. Consequently, the estimated final proficiency values $\hat{\theta}^T$ can be directly compared with the true values $\theta_0$. For example, by computing the Mean Square Error (MSE), i.e., \(\mathbb{E}\Vert \hat{\theta}^T - \theta_0 \Vert\), to evaluate the accuracy of the CAT system \cite{bi2020quality, cheng2009cognitive}.

	{\textbf{Examinee Score Prediction.}} 	In machine learning–based CAT systems, proficiency estimates are often validated by predicting whether examinees will answer unseen questions correctly. Typically, examinees are split into training, validation, and test sets (e.g., 70\%-20\%-10\%), ensuring no overlap. The training set is used to calibrate item parameters (Section~\ref{qca}) and train selection algorithms (Sections~\ref{rl_method}, \ref{meta}). During validation or testing, the responses of each examinee are further divided into a candidate set $\mathcal{Q}_i$ for selecting questions and a held-out meta set $\mathcal{M}_i$ for evaluation. The candidate set $\mathcal{Q}_i$ (with corresponding response label $y$) is used to simulate the CAT procedure: Selecting questions from $\mathcal{Q}_i$, updating proficiency estimates after each step, and then accessing estimate's precision by predicting responses on $\mathcal{M}_i$. \textit{The assumption is that better score predictions reflect more accurate proficiency estimates.} Thus, the binary classification metrics can be used for evaluations, e.g., Prediction Accuracy (ACC) and Area Under ROC Curve (AUC) \cite{bradley1997use}.

	\textbf{Datasets. }To evaluate the effectiveness \textcolor{black}{and generalizability of a CAT system, it is crucial to use diverse datasets that not only challenge the algorithm but also reflect real-world testing scenarios. Such datasets typically contain the question bank, examinee response data, and relevant contextual data. Each of these components is essential for validating the CAT system itself in realistic settings.} Three types of data can be used for this evaluation:

	\textbf{(1) Human Educational Data:} 
	This category includes data collected from educational environments in practice, such as schools, universities, and online learning platforms. It provides insights into how examinees interact with educational content and assessments in a natural setting. The data may encompass examinee information, performance responses, learning behaviors, question characteristics, etc. We 
	have open-sourced a comprehensive education-related dataset library: \url{https://github.com/bigdata-ustc/EduData}. It includes a range of publicly available datasets along with previously private datasets, e.g., ASSISTments \cite{feng2009addressing}, Junyi \cite{chang2015modeling}, EdNet \cite{choi2020ednet}, and Eedi2020 \cite{wang2020diagnostic}. Additionally, we have provided a detailed data analysis to support further CAT research and application in educational settings, which can be found at the EduData GitHub link..

	\textbf{(2) AI Model Response Data:} The CAT paradigm is playing a crucial role in the evaluation of AI models. In particular, proficiency estimates are used to assess performance and rank models, especially for contemporary LLM evaluations. Various large-scale benchmarks and their corresponding response data can be utilized to build and test CAT systems, such as Google's BIG-bench \cite{srivastava2023beyond}, HuggingFace’s Open LLM Leaderboard \cite{open-llm-leaderboard-v1}, HELM \cite{liang2022holistic}, and AlpacaEval \cite{li2023alpacaeval}. These benchmarks encompass a wide range of tasks, with topics spanning linguistics, mathematics, medicine, common-sense reasoning, biology, physics, social bias, programming, and beyond.

	\textbf{(3) Simulated Datasets. }These are artificially created datasets that mimic the characteristics of real examinee responses as illustrated above. They can be tailored to include specific patterns, noise levels, and distributions, allowing for controlled testing of the CAT system under various scenarios. Monte Carlo simulations can also be used to generate datasets with known properties and ground truth \cite{van2010elements}. These datasets are useful for validating the CAT system's capability to estimate proficiencies accurately and to adapt to the simulated changes during the testing process.

	\section{Opportunities for Future Research} \label{OFFR}
	The integration of machine learning into CAT is poised to revolutionize the field. This section explores the future potential of machine learning to expand the applicability, interpretability, and multi-dimensionality of CAT systems.

	{\textit{Multi-Dimensionality of the Assessment Process.}} Future research should harness machine learning to enhance the multi-dimensionality of the assessment. This involves not only the traditional response patterns but also the nuanced analysis of process data such as response times and mouse movements, which can provide insights into an examinee's problem-solving strategies and levels of engagement. Moreover, integrating learning data, such as the examinee's prior interactions with materials, can offer a longitudinal perspective on their learning trajectory and readiness for new concepts. Additionally, the analysis of content, encompassing textual, visual, and auditory materials \cite{xiao2022fine,shi2014multifaceted}, allows for a richer understanding of how examinees interact with multifaceted information. Such machine learning-driven approaches promise to refine CAT systems comprehensively, enabling them to deliver assessments that are not just accurate reflections of an examinee's proficiency but also predictive of their potential for future learning.

		{\textit{Towards Explainable Machine Learning in CAT.}} Traditional CAT systems, particularly those based on information and statistic approaches, are lauded for their interpretability, from the parameters of measurement models to the logic behind the question selection algorithm. This transparency provides valuable insights to all stakeholders, including examinees, parents, and educators, and supports developers in debugging and refining the CAT system. {\color{black}{In practice, interpretability is often important for deploying CAT in high-stakes settings: test providers may need to explain why certain items were chosen, demonstrate fair treatment across groups, and support audits or appeals. Even if the selection policy is complex, this can be addressed by making the main constraints transparent and recording a simple, human-readable reason for each selection.}}
		However, recent machine learning approaches, especially those employing deep learning, have an overwhelming advantage in capabilities on knowledge discovery, at the cost of reduced interpretability. Bridging the gap between these paradigms to create CAT systems that are both accurate and self-explanatory is a significant challenge that future research must address.  This is particularly crucial for high-stakes standardized testing, where the outcomes carry significant consequences.

		{\textit{Empowering CAT with Generative AI.}} Generative artificial intelligence (e.g., LLMs) is trained on massive, cross-domain datasets, endowing it with versatility and a profound repository of world knowledge \cite{dong2023towards,chang2023survey}. These models have already shown preliminary progress in user modeling, such as recommendation systems, and in the generation of personalized strategy \cite{xu2024openp5,zhu2024reliable}. {\color{black}{This connection is conceptually aligned with CAT: both aim to infer latent user traits (e.g., proficiency) from observed behavior and then adapt subsequent interactions accordingly. LLMs/agents can enrich the observation space beyond binary correctness by leveraging intermediate steps, explanations, hesitation patterns, and error types, which can support finer-grained proficiency estimation when properly calibrated.}}
		In the future, there is potential for these large models to significantly enhance CAT systems in various aspects, such as question selection, proficiency assessment, and even the automatic generation of novel, tailored questions on the fly \cite{bhandari23evaluating,robstad2024dattit} -- questions that are not pre-existing in the bank. {\color{black}{A practical integration is to use LLMs as assistive modules: they can draft candidate items conditioned on a targeted construct/skill label, format constraints, and an intended difficulty region, and produce useful metadata (topic tags, expected solution outlines, and common misconceptions) that helps index and retrieve items efficiently.}} We can envision a future where testing paradigms evolve towards greater intelligence and automation. A well-trained testing agent could engage with examinees in natural language interactions, utilizing various cues and process details to conduct a comprehensive assessment of abilities. This approach would move beyond the monotonous task of having examinees respond to questions from a predefined bank or benchmark one by one. Such advancements could lead to more effective and personalized testing experiences.

\textit{Improving Machine Intelligence Evaluation.} Traditional AI model evaluation relies extensively on large, gold-standard benchmarks. The maxim ``more is better'' has driven the use of larger benchmarks to provide comprehensive assessments. However, the sheer size of these benchmarks incurs significant time and computational costs, making fast and economical evaluations challenging. For example, evaluating the performance of a single LLM on the full HELM benchmark can consume over 4,000 GPU hours (or cost over \$10,000 for APIs) \cite{liang2022holistic}. Moreover, these benchmarks are often plagued by low-quality questions, errors, and contamination issues \cite{rodriguez2021evaluation}. As discussed, an increasing number of researchers are attempting to leverage CAT and psychometrics to identify and address these issues, reducing evaluation overhead and gradually transforming it into a new evaluation paradigm. \textcolor{black}{This shift is especially valuable as AI systems approach human-level performance, where CAT can offer finer-grained analysis of cognitive-like behaviors.}
{\color{black}{Although advanced LLMs differ fundamentally from humans in architecture, their learned behaviors often exhibit similar characteristics, since they are trained on large-scale human-produced data and display cognitive-like signatures\cite{zhuang2025position}. CAT does not assume models are ``human''; it only requires observable responses that can be consistently scored and related to item statistics.}} Ultimately, this emerging paradigm may lead to smarter, faster, and more cost-effective evaluations: deepening our understanding of both human and machine intelligence.

	\section{Conclusion}\label{con}

	\textcolor{black}{Computerized Adaptive Testing (CAT) has evolved over more than five decades, achieving remarkable progress in the intelligent evaluation of both humans and AI models through the support of statistical learning. In the past five years, the growing integration of deep learning into CAT has led to the emergence of innovative approaches that were previously unimaginable. These include algorithms for question selection learned directly from large-scale data, retrieval-based methods that improve selection efficiency by up to 200$\times$, and theoretical investigations into the upper bounds of estimation error. Although many of these methods are still in the early stages and not widely used in practice yet, they clearly point to a promising future for smarter testing systems powered by today's wave of AI.}

		This comprehensive survey has highlighted the intricate and expansive nature of CAT, emphasizing the potential and prospects of integrating machine learning to enhance CAT systems. The paper primarily focused on the dual concerns of accuracy and efficiency within machine/human assessment. The insights presented are accessible and relevant not only to specialists in education and psychometrics but also to a broad spectrum of researchers. We encourage interested readers to explore the transformative impact of machine learning in this field and to use this survey as a reference for future research.

\bibliographystyle{ieeetr}
\renewcommand{\baselinestretch}{0.97} 
\bibliography{sample-base}

@article{ocal15effect,
	title={Effect of Content Balancing on Measurement Precision in Computer Adaptive Testing Applications},
	author={{\"O}cal, {\.I}lkay {\"U}{\c{c}}g{\"u}l and Do{\u{g}}an, Nuri},
	journal={Journal of Measurement and Evaluation in Education and Psychology},
	volume={15},
	year={2024},
	number={4},
	pages={395--407},
	publisher={Association for Measurement and Evaluation in Education and Psychology}
}

@mastersthesis{robstad2024dattit,
	title        = {DaTT-IT: Exploring The Effect of Combining Generative AI-Generated Feedback with Computerized Adaptive Testing},
	author       = {Robstad, Anders and Sadun, Robin Lund},
	year         = {2024},
	type         = {Master's thesis},
	school       = {Norwegian University of Science and Technology (NTNU), Faculty of Information Technology and Electrical Engineering, Department of Computer Science},
	language     = {English},
	supervisor   = {Giannakos, Michail and Vesin, Boban},
}

@incollection{rudner2009implementing,
	title={Implementing the graduate management admission test computerized adaptive test},
	author={Rudner, Lawrence M},
	booktitle={Elements of adaptive testing},
	pages={151--165},
	year={2009},
	publisher={Springer}
}

@inproceedings{zhu2018learning,
	title={Learning tree-based deep model for recommender systems},
	author={Zhu, Han and Li, Xiang and Zhang, Pengye and Li, Guozheng and He, Jie and Li, Han and Gai, Kun},
	booktitle={Proceedings of the 24th ACM SIGKDD International Conference on Knowledge Discovery \& Data Mining},
	pages={1079--1088},
	year={2018}
}

@article{huang2009adaptive,
	title={An adaptive testing system for supporting versatile educational assessment},
	author={Huang, Yueh-Min and Lin, Yen-Ting and Cheng, Shu-Chen},
	journal={Computers \& Education},
	volume={52},
	number={1},
	pages={53--67},
	year={2009},
	publisher={Elsevier}
}

@inproceedings{hong2023search,
	title={Search-Efficient Computerized Adaptive Testing},
	author={Hong, Yuting and Tong, Shiwei and Huang, Wei and Zhuang, Yan and Liu, Qi and Chen, Enhong and Li, Xin and He, Yuanjing},
	booktitle={Proceedings of the 32nd ACM International Conference on Information and Knowledge Management},
	pages={773--782},
	year={2023}
}

@book{sands1997computerized,
	title={Computerized adaptive testing: From inquiry to operation.},
	author={Sands, William A and Waters, Brian K and McBride, James R},
	year={1997},
	publisher={American Psychological Association}
}

@incollection{roskam1984new,
	title={A new derivation of the Rasch model},
	author={Roskam, Edw E and Jansen, Paul GW},
	booktitle={Advances in Psychology},
	volume={20},
	pages={293--307},
	year={1984},
	publisher={Elsevier}
}

@article{verschoor2010mathcat,
	title={MATHCAT: A flexible testing system in mathematics education for adults},
	author={Verschoor, Angela J and Straetmans, Gerard JJM},
	journal={Elements of adaptive testing},
	pages={137--149},
	year={2010},
	publisher={Springer}
}

@article{luecht1998some,
	title={Some practical examples of computer-adaptive sequential testing},
	author={Luecht, Richard M and Nungester, Ronald J},
	journal={Journal of Educational Measurement},
	volume={35},
	number={3},
	pages={229--249},
	year={1998},
	publisher={Wiley Online Library}
}

@article{wainer1987item,
	title={Item clusters and computerized adaptive testing: A case for testlets},
	author={Wainer, Howard and Kiely, Gerard L},
	journal={Journal of Educational measurement},
	volume={24},
	number={3},
	pages={185--201},
	year={1987},
	publisher={Wiley Online Library}
}

@inproceedings{bi2020quality,
	title={Quality meets diversity: A model-agnostic framework for computerized adaptive testing},
	author={Bi, Haoyang and Ma, Haiping and Huang, Zhenya and Yin, Yu and Liu, Qi and Chen, Enhong and Su, Yu and Wang, Shijin},
	booktitle={2020 IEEE International Conference on Data Mining (ICDM)},
	pages={42--51},
	year={2020},
	organization={IEEE}
}

@article{cheng2009cognitive,
	title={When cognitive diagnosis meets computerized adaptive testing: CD-CAT},
	author={Cheng, Ying},
	journal={Psychometrika},
	volume={74},
	pages={619--632},
	year={2009},
	publisher={Springer}
}

@article{vie2017review,
	title={A review of recent advances in adaptive assessment},
	author={Vie, Jill-J{\^e}nn and Popineau, Fabrice and Bruillard, {\'E}ric and Bourda, Yolaine},
	journal={Learning analytics: fundaments, applications, and trends},
	pages={113--142},
	year={2017},
	publisher={Springer}
}

@book{lord2012applications,
	title={Applications of item response theory to practical testing problems},
	author={Lord, Frederic M},
	year={2012},
	publisher={Routledge}
}

@book{embretson2013item,
	title={Item response theory},
	author={Embretson, Susan E and Reise, Steven P},
	year={2013},
	publisher={Psychology Press}
}

@article{chang2015psychometrics,
	title={Psychometrics behind computerized adaptive testing},
	author={Chang, Hua-Hua},
	journal={Psychometrika},
	volume={80},
	number={1},
	pages={1--20},
	year={2015},
	publisher={Springer}
}

@article{chang1996global,
	title={A global information approach to computerized adaptive testing},
	author={Chang, Hua-Hua and Ying, Zhiliang},
	journal={Applied Psychological Measurement},
	volume={20},
	number={3},
	pages={213--229},
	year={1996},
	publisher={Sage Publications Sage CA: Thousand Oaks, CA}
}

@article{hooker2009paradoxical,
	title={Paradoxical results in multidimensional item response theory},
	author={Hooker, Giles and Finkelman, Matthew and Schwartzman, Armin},
	journal={Psychometrika},
	volume={74},
	number={3},
	pages={419--442},
	year={2009},
	publisher={Springer}
}

@inproceedings{oren2023proving,
	title={Proving Test Set Contamination for Black-Box Language Models},
	author={Oren, Yonatan and Meister, Nicole and Chatterji, Niladri S and Ladhak, Faisal and Hashimoto, Tatsunori},
	booktitle={The Twelfth International Conference on Learning Representations},
	year={2023}
}

@inproceedings{li2024task,
	title={Task contamination: Language models may not be few-shot anymore},
	author={Li, Changmao and Flanigan, Jeffrey},
	booktitle={Proceedings of the AAAI Conference on Artificial Intelligence},
	volume={38},
	number={16},
	pages={18471--18480},
	year={2024}
}

@article{ackerman2003using,
	title={Using multidimensional item response theory to evaluate educational and psychological tests},
	author={Ackerman, Terry A and Gierl, Mark J and Walker, Cindy M},
	journal={Educational Measurement: Issues and Practice},
	volume={22},
	number={3},
	pages={37--51},
	year={2003},
	publisher={Wiley Online Library}
}

@ARTICLE{wang2020neural,
	author={Wang, Fei and Liu, Qi and Chen, Enhong and Huang, Zhenya and Yin, Yu and Wang, Shijin and Su, Yu},
	journal={IEEE Transactions on Knowledge and Data Engineering}, 
	title={NeuralCD: A General Framework for Cognitive Diagnosis}, 
	year={2023},
	volume={35},
	number={8},
	pages={8312-8327},
	doi={10.1109/TKDE.2022.3201037}
}

@ARTICLE{9674822,
	author={Bao, Shilong and Xu, Qianqian and Yang, Zhiyong and Cao, Xiaochun and Huang, Qingming},
	journal={IEEE Transactions on Pattern Analysis and Machine Intelligence}, 
	title={Rethinking Collaborative Metric Learning: Toward an Efficient Alternative Without Negative Sampling}, 
	year={2023},
	volume={45},
	number={1},
	pages={1017-1035},
	doi={10.1109/TPAMI.2022.3141095}
}

@ARTICLE{8304801,
	author={Lee, Chang-Shing and Wang, Mei-Hui and Wang, Chi-Shiang and Teytaud, Olivier and Liu, Jialin and Lin, Su-Wei and Hung, Pi-Hsia},
	journal={IEEE Transactions on Fuzzy Systems}, 
	title={PSO-Based Fuzzy Markup Language for Student Learning Performance Evaluation and Educational Application}, 
	year={2018},
	volume={26},
	number={5},
	pages={2618-2633},
	doi={10.1109/TFUZZ.2018.2810814}
}

@article{arulkumaran2017brief,
	title={A brief survey of deep reinforcement learning},
	author={Arulkumaran, Kai and Deisenroth, Marc Peter and Brundage, Miles and Bharath, Anil Anthony},
	journal={arXiv preprint arXiv:1708.05866},
	year={2017}
}

@article{kober2013reinforcement,
	title={Reinforcement learning in robotics: A survey},
	author={Kober, Jens and Bagnell, J Andrew and Peters, Jan},
	journal={The International Journal of Robotics Research},
	volume={32},
	number={11},
	pages={1238--1274},
	year={2013},
	publisher={SAGE Publications Sage UK: London, England}
}

@inproceedings{finn2017model,
	title={Model-agnostic meta-learning for fast adaptation of deep networks},
	author={Finn, Chelsea and Abbeel, Pieter and Levine, Sergey},
	booktitle={International Conference on Machine Learning},
	pages={1126--1135},
	year={2017},
	organization={PMLR}
}

@book{wainer2000computerized,
	title={Computerized adaptive testing: A primer},
	author={Wainer, Howard and Dorans, Neil J and Flaugher, Ronald and Green, Bert F and Mislevy, Robert J},
	year={2000},
	publisher={Routledge}
}

@article{li2020deep,
	title={Deep reinforcement learning for adaptive learning systems},
	author={Li, Xiao and Xu, Hanchen and Zhang, Jinming and Chang, Hua-hua},
	journal={arXiv preprint arXiv:2004.08410},
	year={2020}
}

@article{wang2020diagnostic,  
	title={Diagnostic questions: The neurips 2020 education challenge},
	author={Wang, Zichao and Lamb, Angus and Saveliev, Evgeny and Cameron, Pashmina and Zaykov, Yordan and Hern{\'a}ndez-Lobato, Jos{\'e} Miguel and Turner, Richard E and Baraniuk, Richard G and Barton, Craig and Jones, Simon Peyton and Woodhead, Simon and Zhang, Cheng},
	journal={arXiv preprint arXiv:2007.12061},  
	year={2020}
}

@book{cheng2008computerized,
	title={Computerized adaptive testing—new developments and applications},
	author={Cheng, Ying},
	year={2008},
	publisher={University of Illinois at Urbana-Champaign}
}

@article{van1998bayesian,
	title={Bayesian item selection criteria for adaptive testing},
	author={van der Linden, Wim J},
	journal={Psychometrika},
	volume={63},
	number={2},
	pages={201--216},
	year={1998},
	publisher={Springer}
}

@article{veerkamp1997some,
	title={Some new item selection criteria for adaptive testing},
	author={Veerkamp, Wim JJ and Berger, Martijn PF},
	journal={Journal of Educational and Behavioral Statistics},
	volume={22},
	number={2},
	pages={203--226},
	year={1997},
	publisher={Sage Publications Sage CA: Los Angeles, CA}
}

@article{von2014dina,
	title={The DINA model as a constrained general diagnostic model: Two variants of a model equivalency},
	author={Von Davier, Matthias},
	journal={British Journal of Mathematical and Statistical Psychology},
	volume={67},
	number={1},
	pages={49--71},
	year={2014},
	publisher={Wiley Online Library}
}

@article{de2009dina,
	title={DINA model and parameter estimation: A didactic},
	author={De La Torre, Jimmy},
	journal={Journal of educational and behavioral statistics},
	volume={34},
	number={1},
	pages={115--130},
	year={2009},
	publisher={Sage Publications Sage CA: Los Angeles, CA}
}

@article{bradley1997use,
	title={The use of the area under the ROC curve in the evaluation of machine learning algorithms},
	author={Bradley, Andrew P},
	journal={Pattern recognition},
	volume={30},
	number={7},
	pages={1145--1159},
	year={1997},
	publisher={Elsevier}
}

@inproceedings{chen2015computer,
	title={Computer Adaptive Testing Using the Same-Decision Probability.},
	author={Chen, Suming Jeremiah and Choi, Arthur and Darwiche, Adnan},
	booktitle={BMA@ UAI},
	pages={34--43},
	year={2015}
}

@book{ross2014first,
	title={A first course in probability},
	author={Ross, Sheldon M},
	year={2014},
	publisher={Pearson}
}

@inproceedings{gao2021rcd,
	title={RCD: Relation Map Driven Cognitive Diagnosis for Intelligent Education Systems},
	author={Gao, Weibo and Liu, Qi and Huang, Zhenya and Yin, Yu and Bi, Haoyang and Wang, Mu-Chun and Ma, Jianhui and Wang, Shijin and Su, Yu},
	booktitle={Proceedings of the 44th International ACM SIGIR Conference on Research and Development in Information Retrieval},
	pages={501--510},
	year={2021}
}

@inproceedings{ghosh2021bobcat,
	title     = {BOBCAT: Bilevel Optimization-Based Computerized Adaptive Testing},
	author    = {Ghosh, Aritra and Lan, Andrew},
	booktitle = {Proceedings of the Thirtieth International Joint Conference on
	Artificial Intelligence, {IJCAI-21}},
	publisher = {International Joint Conferences on Artificial Intelligence Organization},
	pages     = {2410--2417},
	year      = {2021},
	month     = {8}
}

@inproceedings{cheng2019dirt,
	title={DIRT: Deep learning enhanced item response theory for cognitive diagnosis},
	author={Cheng, Song and Liu, Qi and Chen, Enhong and Huang, Zai and Huang, Zhenya and Chen, Yiying and Ma, Haiping and Hu, Guoping},
	booktitle={Proceedings of the 28th ACM International Conference on Information and Knowledge Management},
	pages={2397--2400},
	year={2019}
}

@article{zhuang2022fully,
	title={Fully Adaptive Framework: Neural Computerized Adaptive Testing for Online Education}, volume={36},
	number={4},
	journal={Proceedings of the AAAI Conference on Artificial Intelligence}, author={Zhuang, Yan and Liu, Qi and Huang, Zhenya and Li, Zhi and Shen, Shuanghong and Ma, Haiping},
	year={2022},
	month={Jun.},
	pages={4734-4742} 
}

@InProceedings{pmlr-v119-mirzasoleiman20a,
	title = 	 {Coresets for Data-efficient Training of Machine Learning Models},
	author =       {Mirzasoleiman, Baharan and Bilmes, Jeff and Leskovec, Jure},
	booktitle = 	 {Proceedings of the 37th International Conference on Machine Learning},
	pages = 	 {6950--6960},
	year = 	 {2020},
	editor = 	 {III, Hal Daumé and Singh, Aarti},
	volume = 	 {119},
	series = 	 {Proceedings of Machine Learning Research},
	month = 	 {13--18 Jul},
	publisher =    {PMLR}
}

@article{article07ac,
	author = {Krishnakumar, Anita},
	year = {2007},
	month = {07},
	pages = {},
	title = {Active Learning Literature Survey}
}

@article{reckase200618,
	title={18 Multidimensional Item Response Theory},
	author={Reckase, Mark D},
	journal={Handbook of statistics},
	volume={26},
	pages={607--642},
	year={2006},
	publisher={Elsevier}
}

@ARTICLE{6747346,
	author={Huang, Sheng-Jun and Jin, Rong and Zhou, Zhi-Hua},
	journal={IEEE Transactions on Pattern Analysis and Machine Intelligence}, 
	title={Active Learning by Querying Informative and Representative Examples}, 
	year={2014},
	volume={36},
	number={10},
	pages={1936-1949},
	doi={10.1109/TPAMI.2014.2307881}
}

@ARTICLE{9238451,
	author={Xu, Zhixiong and Chen, Xiliang and Cao, Lei},
	journal={IEEE Transactions on Cybernetics}, 
	title={Fast Task Adaptation Based on the Combination of Model-Based and Gradient-Based Meta Learning}, 
	year={2022},
	volume={52},
	number={6},
	pages={5209-5218},
	doi={10.1109/TCYB.2020.3028378}
}

@ARTICLE{6616533,
	author={Doshi-Velez, Finale and Pfau, David and Wood, Frank and Roy, Nicholas},
	journal={IEEE Transactions on Pattern Analysis and Machine Intelligence}, 
	title={Bayesian Nonparametric Methods for Partially-Observable Reinforcement Learning}, 
	year={2015},
	volume={37},
	number={2},
	pages={394-407},
	doi={10.1109/TPAMI.2013.191}
}

@ARTICLE{10372131,
	author={Li, Jingyao and Chen, Pengguang and Yu, Shaozuo and Liu, Shu and Jia, Jiaya},
	journal={IEEE Transactions on Pattern Analysis and Machine Intelligence}, 
	title={BAL: Balancing Diversity and Novelty for Active Learning}, 
	year={2024},
	volume={46},
	number={5},
	pages={3653-3664},
	doi={10.1109/TPAMI.2023.3345844}
}

@article{rong2023towards,
	title={Towards human-centered explainable ai: A survey of user studies for model explanations},
	author={Rong, Yao and Leemann, Tobias and Nguyen, Thai-Trang and Fiedler, Lisa and Qian, Peizhu and Unhelkar, Vaibhav and Seidel, Tina and Kasneci, Gjergji and Kasneci, Enkelejda},
	journal={IEEE transactions on pattern analysis and machine intelligence},
	year={2023},
	publisher={IEEE}
}

@inproceedings{rodriguez2021evaluation,
	title={Evaluation examples are not equally informative: How should that change NLP leaderboards?},
	author={Rodriguez, Pedro and Barrow, Joe and Hoyle, Alexander Miserlis and Lalor, John P and Jia, Robin and Boyd-Graber, Jordan},
	booktitle={Proceedings of the 59th Annual Meeting of the Association for Computational Linguistics and the 11th International Joint Conference on Natural Language Processing (Volume 1: Long Papers)},
	pages={4486--4503},
	year={2021}
}

@phdthesis{li2020data,
	title={Data-driven adaptive learning systems},
	author={Li, Xiao},
	year={2020}
}

@inproceedings{gilavert2022computerized,
	title={Computerized Adaptive Testing: A Unified Approach Under Markov Decision Process},
	author={Gilavert, Patricia and Freire, Valdinei},
	booktitle={International Conference on Computational Science and Its Applications},
	pages={591--602},
	year={2022},
	organization={Springer}
}

@article{shin2022building,
	title={Building an intelligent recommendation system for personalized test scheduling in computerized assessments: A reinforcement learning approach},
	author={Shin, Jinnie and Bulut, Okan},
	journal={Behavior Research Methods},
	volume={54},
	number={1},
	pages={216--232},
	year={2022},
	publisher={Springer}
}

@article{ma2023novel,
	title={A novel computerized adaptive testing framework with decoupled learning selector},
	author={Ma, Haiping and Zeng, Yi and Yang, Shangshang and Qin, Chuan and Zhang, Xingyi and Zhang, Limiao},
	journal={Complex \& Intelligent Systems},
	pages={1--12},
	year={2023},
	publisher={Springer}
}

@article{nurakhmetov2019reinforcement,
	title={Reinforcement learning applied to adaptive classification testing},
	author={Nurakhmetov, Darkhan},
	journal={Theoretical and Practical Advances in Computer-based Educational Measurement},
	pages={325--336},
	year={2019},
	publisher={Springer International Publishing}
}

@article{segall2005computerized,
	title={Computerized adaptive testing},
	author={Segall, Daniel O},
	journal={Encyclopedia of social measurement},
	volume={1},
	pages={429--438},
	year={2005},
	publisher={Elsevier Amsterdam}
}

@article{revuelta1998comparison,
	title={A comparison of item exposure control methods in computerized adaptive testing},
	author={Revuelta, Javier and Ponsoda, Vicente},
	journal={Journal of Educational Measurement},
	volume={35},
	number={4},
	pages={311--327},
	year={1998},
	publisher={Wiley Online Library}
}

@article{reckase2010designing,
	title={Designing item pools to optimize the functioning of a computerized adaptive test},
	author={Reckase, Mark D},
	journal={Psychological Test and Assessment Modeling},
	volume={52},
	number={2},
	pages={127},
	year={2010},
	publisher={PABST Science Publishers}
}

@article{he2014item,
	title={Item pool design for an operational variable-length computerized adaptive test},
	author={He, Wei and Reckase, Mark D},
	journal={Educational and Psychological Measurement},
	volume={74},
	number={3},
	pages={473--494},
	year={2014},
	publisher={SAGE Publications Sage CA: Los Angeles, CA}
}

@article{van2000integer,
	title={An Integer-Programming Approach to Item Pool Design. Law School Admission Council Computerized Testing Report. LSAC Research Report Series.},
	author={Van Der Linden, Wim J and Veldkamp, Bernard P and Reese, Lynda M},
	year={2000},
	publisher={ERIC}
}

@article{van2006assembling,
	title={Assembling a computerized adaptive testing item pool as a set of linear tests},
	author={van der Linden, Wim J and Ariel, Adelaide and Veldkamp, Bernard P},
	journal={Journal of Educational and Behavioral Statistics},
	volume={31},
	number={1},
	pages={81--99},
	year={2006},
	publisher={Sage Publications Sage CA: Los Angeles, CA}
}

@incollection{way2005developing,
	title={Developing, maintaining, and renewing the item inventory to support CBT},
	author={Way, Walter D and Steffen, Manfred and Anderson, Gordon Stephen},
	booktitle={Computer-Based Testing},
	pages={143--164},
	year={2005},
	publisher={Routledge}
}

@article{ariel2004constructing,
	title={Constructing rotating item pools for constrained adaptive testing},
	author={Ariel, Adelaide and Veldkamp, Bernard P and van der Linden, Wim J},
	journal={Journal of Educational Measurement},
	volume={41},
	number={4},
	pages={345--359},
	year={2004},
	publisher={Wiley Online Library}
}

@book{gulliksen2013theory,
	title={Theory of mental tests},
	author={Gulliksen, Harold},
	year={2013},
	publisher={Routledge}
}

@article{swanson1993model,
	title={A model and heuristic for solving very large item selection problems},
	author={Swanson, Len and Stocking, Martha L},
	journal={Applied Psychological Measurement},
	volume={17},
	number={2},
	pages={151--166},
	year={1993},
	publisher={Sage Publications Sage CA: Thousand Oaks, CA}
}

@article{stocking1998optimal,
	title={Optimal design of item banks for computerized adaptive tests},
	author={Stocking, Martha L and Swanson, Len},
	journal={Applied Psychological Measurement},
	volume={22},
	number={3},
	pages={271--279},
	year={1998},
	publisher={SAGE PUBLICATIONS, INC. 2455 Teller Road, Thousand Oaks, CA 91320}
}

@inproceedings{lopez2008helping,
	title={Helping tools for item bank calibration and development of computerized adaptive tests},
	author={L{\'o}pez-Cuadrado, Javier and Armendariz, AJ and P{\'e}rez, Tom{\'a}s A and Arruabarrena, Rosa},
	booktitle={International Technology, Education, and Development Conference (INTED2008). Valencia, Espa{\~n}a: International Association of Technology, Education, and Development},
	year={2008}
}

@article{conejo2004siette,
	title={SIETTE: A web-based tool for adaptive testing},
	author={Conejo, Ricardo and Guzm{\'a}n, Eduardo and Mill{\'a}n, Eva and Trella, M{\'o}nica and P{\'e}rez-De-La-Cruz, Jos{\'e} Luis and R{\'\i}os, Antonia},
	journal={International Journal of Artificial Intelligence in Education},
	volume={14},
	number={1},
	pages={29--61},
	year={2004},
	publisher={Ios Press}
}

@article{lopez2006adaptive,
	title={Adaptive evaluation in an e-learning system architecture},
	author={L{\'o}pez-Cuadrado, J and Armendariz, A and P{\'e}rez, TA},
	journal={Current Developments in Technology-Assisted Education},
	pages={1507--1511},
	year={2006},
	publisher={Citeseer}
}

@inproceedings{kozierkiewicz2014item,
	title={An item bank calibration method for a computer adaptive test},
	author={Kozierkiewicz-Hetma{\'n}ska, Adrianna and Poniatowski, Rafa{\l}},
	booktitle={Asian Conference on Intelligent Information and Database Systems},
	pages={375--383},
	year={2014},
	organization={Springer}
}

@article{de2010primer,
	title={A primer on classical test theory and item response theory for assessments in medical education},
	author={De Champlain, Andr{\'e} F},
	journal={Medical education},
	volume={44},
	number={1},
	pages={109--117},
	year={2010},
	publisher={Wiley Online Library}
}

@article{chang2009applying,
	title={Applying IRT to Estimate Learning Ability and K-means Clustering in Web based Learning.},
	author={Chang, Wen-Chih and Yang, Hsuan-Che},
	journal={J. Softw.},
	volume={4},
	number={2},
	pages={167--174},
	year={2009},
	publisher={Citeseer}
}

@inproceedings{chen2004personalized,
	title={A personalized courseware recommendation system based on fuzzy item response theory},
	author={Chen, Chih-Ming and Duh, Ling-Jiun and Liu, Chao-Yu},
	booktitle={IEEE International Conference on e-Technology, e-Commerce and e-Service, 2004. EEE'04. 2004},
	pages={305--308},
	year={2004},
	organization={IEEE}
}

@article{devellis2006classical,
	title={Classical test theory},
	author={DeVellis, Robert F},
	journal={Medical care},
	pages={S50--S59},
	year={2006},
	publisher={JSTOR}
}

@inproceedings{huang2017question,
	title={Question Difficulty Prediction for READING Problems in Standard Tests},
	author={Huang, Zhenya and Liu, Qi and Chen, Enhong and Zhao, Hongke and Gao, Mingyong and Wei, Si and Su, Yu and Hu, Guoping},
	booktitle={Proceedings of the AAAI Conference on Artificial Intelligence},
	volume={31},
	number={1},
	year={2017}
}

@inproceedings{qiu2019question,
	title={Question difficulty prediction for multiple choice problems in medical exams},
	author={Qiu, Zhaopeng and Wu, Xian and Fan, Wei},
	booktitle={Proceedings of the 28th ACM International Conference on Information and Knowledge Management},
	pages={139--148},
	year={2019}
}

@inproceedings{huang2021stan,
	title={Stan: adversarial network for cross-domain question difficulty prediction},
	author={Huang, Ye and Huang, Wei and Tong, Shiwei and Huang, Zhenya and Liu, Qi and Chen, Enhong and Ma, Jianhui and Wan, Liang and Wang, Shijin},
	booktitle={2021 IEEE International Conference on Data Mining (ICDM)},
	pages={220--229},
	year={2021},
	organization={IEEE}
}

@inproceedings{huang2019hierarchical,
	title={Hierarchical multi-label text classification: An attention-based recurrent network approach},
	author={Huang, Wei and Chen, Enhong and Liu, Qi and Chen, Yuying and Huang, Zai and Liu, Yang and Zhao, Zhou and Zhang, Dan and Wang, Shijin},
	booktitle={Proceedings of the 28th ACM international conference on information and knowledge management},
	pages={1051--1060},
	year={2019}
}

@article{huang2022hmcnet,
	title={HmcNet: A General Approach for Hierarchical Multi-Label Classification},
	author={Huang, Wei and Chen, Enhong and Liu, Qi and Xiong, Hui and Huang, Zhenya and Tong, Shiwei and Zhang, Dan},
	journal={IEEE Transactions on Knowledge and Data Engineering},
	year={2022},
	publisher={IEEE}
}

@inproceedings{lei2021consistency,
	title={Consistency-aware Multi-modal Network for Hierarchical Multi-label Classification in Online Education System},
	author={Lei, Siqi and Huang, Wei and Tong, Shiwei and Liu, Qi and Huang, Zhenya and Chen, Enhong and Su, Yu},
	booktitle={2021 IEEE International Conference on Big Knowledge (ICBK)},
	pages={1--8},
	year={2021},
	organization={IEEE}
}

@inproceedings{yin2019quesnet,
	title={Quesnet: A unified representation for heterogeneous test questions},
	author={Yin, Yu and Liu, Qi and Huang, Zhenya and Chen, Enhong and Tong, Wei and Wang, Shijin and Su, Yu},
	booktitle={Proceedings of the 25th acm sigkdd international conference on knowledge discovery \& data mining},
	pages={1328--1336},
	year={2019}
}

@article{ning2023towards,
	title={Towards a Holistic Understanding of Mathematical Questions with Contrastive Pre-training},
	author={Ning, Yuting and Huang, Zhenya and Lin, Xin and Chen, Enhong and Tong, Shiwei and Gong, Zheng and Wang, Shijin},
	journal={arXiv preprint arXiv:2301.07558},
	year={2023}
}

@inproceedings{wang2023gmocat,
	title={GMOCAT: A Graph-Enhanced Multi-Objective Method for Computerized Adaptive Testing},
	author={Wang, Hangyu and Long, Ting and Yin, Liang and Zhang, Weinan and Xia, Wei and Hong, Qichen and Xia, Dingyin and Tang, Ruiming and Yu, Yong},
	booktitle={Proceedings of the 29th ACM SIGKDD Conference on Knowledge Discovery and Data Mining},
	pages={2279--2289},
	year={2023}
}

@book{feinberg2012handbook,
	title={Handbook of Markov decision processes: methods and applications},
	author={Feinberg, Eugene A and Shwartz, Adam},
	volume={40},
	year={2012},
	publisher={Springer Science \& Business Media}
}

@book{sutton2018reinforcement,
	title={Reinforcement learning: An introduction},
	author={Sutton, Richard S and Barto, Andrew G},
	year={2018},
	publisher={MIT press}
}

@article{agarwal2021theory,
	title={On the theory of policy gradient methods: Optimality, approximation, and distribution shift},
	author={Agarwal, Alekh and Kakade, Sham M and Lee, Jason D and Mahajan, Gaurav},
	journal={The Journal of Machine Learning Research},
	volume={22},
	number={1},
	pages={4431--4506},
	year={2021},
	publisher={JMLRORG}
}

@inproceedings{mnih2016asynchronous,
	title={Asynchronous methods for deep reinforcement learning},
	author={Mnih, Volodymyr and Badia, Adria Puigdomenech and Mirza, Mehdi and Graves, Alex and Lillicrap, Timothy and Harley, Tim and Silver, David and Kavukcuoglu, Koray},
	booktitle={International conference on machine learning},
	pages={1928--1937},
	year={2016},
	organization={PMLR}
}

@article{Cleary1968TESTBP,
	title={Test Bias: Prediction of Grades of Negro and White Students in Integrated Colleges},
	author={T. Anne Cleary},
	journal={Journal of Educational Measurement},
	year={1968},
	volume={5},
	pages={115-124},
}

@inproceedings{Chai2022FairnessWA,
	title     = {Fairness with Adaptive Weights},
	author    = {Junyi Chai and Xiaoqian Wang},
	booktitle = 	 {Proceedings of the 39th International Conference on Machine Learning},
	year      = {2022},
	volume       = {162},
	pages        = {2853--2866},
}

@inproceedings{Li2022AchievingFA,
	title     = {Achieving Fairness at No Utility Cost via Data Reweighing},
	author    = {Peizhao Li and Hongfu Liu},
	booktitle = 	 {Proceedings of the 39th International Conference on Machine Learning},
	year      = {2022},
	volume       = {162},
	pages        = {12917--12930},
}

@misc{Thompson2022Is,
	title = {Is the NAPLAN results delay about politics or precision?},
	author = {Greg Thompson},
	howpublished = {\url{https://blog.aare.edu.au/is-the-naplan-results-delay-about-politics-or-precision/}},
	note = {Accessed: 2022-8-29}
}

@misc{Roberts2017Stan,
	title = {Standardised tests are culturally biased against rural students},
	author = {Philip Roberts},
	howpublished = {https://theconversation.com/standardised-tests-are-culturally-biased-against-rural-students-86305},
	note = {Accessed: 2017-11-21}
}

@misc{li2023alpacaeval,
	title={Alpacaeval: An automatic evaluator of instruction-following models},
	author={Li, Xuechen and Zhang, Tianyi and Dubois, Yann and Taori, Rohan and Gulrajani, Ishaan and Guestrin, Carlos and Liang, Percy and Hashimoto, Tatsunori B},
	year={2023}
}

@article{liang2022holistic,
	title={Holistic evaluation of language models},
	author={Liang, Percy and Bommasani, Rishi and Lee, Tony and Tsipras, Dimitris and Soylu, Dilara and Yasunaga, Michihiro and Zhang, Yian and Narayanan, Deepak and Wu, Yuhuai and Kumar, Ananya and others},
	journal={arXiv preprint arXiv:2211.09110},
	year={2022}
}

@article{srivastava2023beyond,
	title={Beyond the Imitation Game: Quantifying and extrapolating the capabilities of language models},
	author={Srivastava, Aarohi and Rastogi, Abhinav and Rao, Abhishek and Shoeb, Abu Awal and Abid, Abubakar and Fisch, Adam and Brown, Adam R and Santoro, Adam and Gupta, Aditya and Garriga-Alonso, Adri and others},
	journal={Transactions on machine learning research},
	year={2023}
}

@misc{open-llm-leaderboard-v1,
	author = {Edward Beeching and Clémentine Fourrier and Nathan Habib and Sheon Han and Nathan Lambert and Nazneen Rajani and Omar Sanseviero and Lewis Tunstall and Thomas Wolf},
	title = {Open LLM Leaderboard (2023-2024)},
	year = {2023},
	publisher = {Hugging Face},
	howpublished = "\url{https://huggingface.co/spaces/open-llm-leaderboard-old/open_llm_leaderboard}"
}

@inproceedings{xu2024openp5,
	title={OpenP5: An Open-Source Platform for Developing, Training, and Evaluating LLM-based Recommender Systems},
	author={Xu, Shuyuan and Hua, Wenyue and Zhang, Yongfeng},
	booktitle={Proceedings of the 47th International ACM SIGIR Conference on Research and Development in Information Retrieval},
	pages={386--394},
	year={2024}
}

@inproceedings{zhu2024reliable,
	title={How Reliable is Your Simulator? Analysis on the Limitations of Current LLM-based User Simulators for Conversational Recommendation},
	author={Zhu, Lixi and Huang, Xiaowen and Sang, Jitao},
	booktitle={Companion Proceedings of the ACM on Web Conference 2024},
	pages={1726--1732},
	year={2024}
}

@incollection{kizilcec2022algorithmic,
	title={Algorithmic fairness in education},
	author={Kizilcec, Ren{\'e} F and Lee, Hansol},
	booktitle={The ethics of artificial intelligence in education},
	pages={174--202},
	year={2022},
	publisher={Routledge}
}

@book{camilli1994methods,
	title={Methods for identifying biased test items},
	author={Camilli, Gregory and Shepard, Lorrie A},
	volume={4},
	year={1994},
	publisher={Sage}
}

@book{hambleton1991fundamentals,
	title={Fundamentals of item response theory},
	author={Hambleton, Ronald K and Swaminathan, Hariharan and Rogers, H Jane},
	volume={2},
	year={1991},
	publisher={Sage}
}

@article{chu2013detecting,
	title={Detecting biased items using CATSIB to increase fairness in computer adaptive tests},
	author={Chu, Man-Wai and Lai, Hollis},
	journal={Alberta Journal of Educational Research},
	volume={59},
	number={4},
	pages={630--643},
	year={2013}
}

@article{mellenbergh1989item,
	title={Item bias and item response theory},
	author={Mellenbergh, Gideon J},
	journal={International journal of educational research},
	volume={13},
	number={2},
	pages={127--143},
	year={1989},
	publisher={Elsevier}
}

@article{li2023deep,
	title={Deep reinforcement learning for adaptive learning systems},
	author={Li, Xiao and Xu, Hanchen and Zhang, Jinming and Chang, Hua-hua},
	journal={Journal of Educational and Behavioral Statistics},
	volume={48},
	number={2},
	pages={220--243},
	year={2023},
	publisher={SAGE Publications Sage CA: Los Angeles, CA}
}

@article{bertsekas1991analysis,
	title={An analysis of stochastic shortest path problems},
	author={Bertsekas, Dimitri P and Tsitsiklis, John N},
	journal={Mathematics of Operations Research},
	volume={16},
	number={3},
	pages={580--595},
	year={1991},
	publisher={INFORMS}
}

@inproceedings{zhuang2022robust,
	title={A Robust Computerized Adaptive Testing Approach in Educational Question Retrieval},
	author={Zhuang, Yan and Liu, Qi and Huang, Zhenya and Li, Zhi and Jin, Binbin and Bi, Haoyang and Chen, Enhong and Wang, Shijin},
	booktitle={Proceedings of the 45th International ACM SIGIR Conference on Research and Development in Information Retrieval},
	pages={416--426},
	year={2022}
}

@inproceedings{hoerger2021line,
	title={An on-line POMDP solver for continuous observation spaces},
	author={Hoerger, Marcus and Kurniawati, Hanna},
	booktitle={2021 IEEE International Conference on Robotics and Automation (ICRA)},
	pages={7643--7649},
	year={2021},
	organization={IEEE}
}

@inproceedings{schwartz2022online,
	title={Online Planning for Interactive-POMDPs Using Nested Monte Carlo Tree Search},
	author={Schwartz, Jonathon and Zhou, Ruijia and Kurniawati, Hanna},
	booktitle={2022 IEEE/RSJ International Conference on Intelligent Robots and Systems (IROS)},
	pages={8770--8777},
	year={2022},
	organization={IEEE}
}

@article{chen2018recommendation,
	title={Recommendation system for adaptive learning},
	author={Chen, Yunxiao and Li, Xiaoou and Liu, Jingchen and Ying, Zhiliang},
	journal={Applied psychological measurement},
	volume={42},
	number={1},
	pages={24--41},
	year={2018},
	publisher={Sage Publications Sage CA: Los Angeles, CA}
}

@article{feng2023balancing,
	title={Balancing Test Accuracy and Security in Computerized Adaptive Testing},
	author={Feng, Wanyong and Ghosh, Aritra and Sireci, Stephen and Lan, Andrew S},
	journal={arXiv preprint arXiv:2305.18312},
	year={2023}
}

@inproceedings{mujtaba2020artificial,
	title={Artificial intelligence in computerized adaptive testing},
	author={Mujtaba, Dena F and Mahapatra, Nihar R},
	booktitle={2020 International Conference on Computational Science and Computational Intelligence (CSCI)},
	pages={649--654},
	year={2020},
	organization={IEEE}
}

@inproceedings{wu2020virt,
	author       = {Mike Wu and
	Richard Lee Davis and
	Benjamin W. Domingue and
	Chris Piech and
	Noah D. Goodman},
	editor       = {Anna N. Rafferty and
	Jacob Whitehill and
	Crist{\'{o}}bal Romero and
	Violetta Cavalli{-}Sforza},
	title        = {Variational Item Response Theory: Fast, Accurate, and Expressive},
	booktitle    = {Proceedings of the 13th International Conference on Educational Data
	Mining, {EDM} 2020, Fully virtual conference, July 10-13, 2020},
	publisher    = {International Educational Data Mining Society},
	year         = {2020}
}

@article{zheng2024mxml,
	title={MxML (Exploring the relationship between measurement and machine learning): Current state of the field},
	author={Zheng, Yi and Nydick, Steven and Huang, Sijia and Zhang, Susu},
	journal={Educational Measurement: Issues and Practice},
	volume={43},
	number={1},
	pages={19--38},
	year={2024},
	publisher={Wiley Online Library}
}

@article{zheng2020using,
	title={Using machine learning methods to develop a short tree-based adaptive classification test: Case study with a high-dimensional item pool and imbalanced data},
	author={Zheng, Yi and Cheon, Hyunjung and Katz, Charles M},
	journal={Applied psychological measurement},
	volume={44},
	number={7-8},
	pages={499--514},
	year={2020},
	publisher={Sage Publications Sage CA: Los Angeles, CA}
}

@article{de2011gdina,
	title = {The generalized {DINA} model framework},
	volume = {76},
	issn = {1860-0980},
	doi = {10.1007/s11336-011-9207-7},
	number = {2},
	journal = {Psychometrika},
	author = {de la Torre, Jimmy},
	year = {2011},
	note = {Place: Germany
	Publisher: Springer},
	pages = {179--199},
}

@article{liu2018fuzzycd,
	title={Fuzzy cognitive diagnosis for modelling examinee performance},
	author={Liu, Qi and Wu, Runze and Chen, Enhong and Xu, Guandong and Su, Yu and Chen, Zhigang and Hu, Guoping},
	journal={ACM Transactions on Intelligent Systems and Technology (TIST)},
	volume={9},
	number={4},
	pages={1--26},
	year={2018},
	publisher={ACM New York, NY, USA}
}

@article{leighton2004ahm,
	title={The attribute hierarchy method for cognitive assessment: A variation on Tatsuoka's rule-space approach},
	author={Leighton, Jacqueline P and Gierl, Mark J and Hunka, Stephen M},
	journal={Journal of educational measurement},
	volume={41},
	number={3},
	pages={205--237},
	year={2004},
	publisher={Wiley Online Library}
}

@inproceedings{tong2021irr,
	title={Item Response Ranking for Cognitive Diagnosis.},
	author={Tong, Shiwei and Liu, Qi and Yu, Runlong and Huang, Wei and Huang, Zhenya and Pardos, Zachary A and Jiang, Weijie},
	booktitle={IJCAI},
	pages={1750--1756},
	year={2021}
}

@inproceedings{li2022hiercdf,
	title={HierCDF: A Bayesian Network-based Hierarchical Cognitive Diagnosis Framework},
	author={Li, Jiatong and Wang, Fei and Liu, Qi and Zhu, Mengxiao and Huang, Wei and Huang, Zhenya and Chen, Enhong and Su, Yu and Wang, Shijin},
	booktitle={Proceedings of the 28th ACM SIGKDD Conference on Knowledge Discovery and Data Mining},
	pages={904--913},
	year={2022}
}

@article{gao2022deepcdf,
	title = {Deep cognitive diagnosis model for predicting students’ performance},
	volume = {126},
	issn = {0167-739X},
	urldate = {2023-11-13},
	journal = {Future Generation Computer Systems},
	author = {Gao, Lina and Zhao, Zhongying and Li, Chao and Zhao, Jianli and Zeng, Qingtian},
	month = jan,
	year = {2022},
	pages = {252--262}
}

@inproceedings{pei2022self,
	title={Self-Attention Gated Cognitive Diagnosis for Faster Adaptive Educational Assessments},
	author={Pei, Xiaohuan and Yang, Shuo and Huang, Jiajun and Xu, Chang},
	booktitle={2022 IEEE International Conference on Data Mining (ICDM)},
	pages={408--417},
	year={2022},
	organization={IEEE}
}

@inproceedings{wang2023scd,
	title={Self-supervised graph learning for long-tailed cognitive diagnosis},
	author={Wang, Shanshan and Zeng, Zhen and Yang, Xun and Zhang, Xingyi},
	booktitle={Proceedings of the AAAI Conference on Artificial Intelligence},
	volume={37},
	number={1},
	pages={110--118},
	year={2023}
}

@inproceedings{bi2023beta,
	title={BETA-CD: A Bayesian meta-learned cognitive diagnosis framework for personalized learning},
	author={Bi, Haoyang and Chen, Enhong and He, Weidong and Wu, Han and Zhao, Weihao and Wang, Shijin and Wu, Jinze},
	booktitle={Proceedings of the AAAI Conference on Artificial Intelligence},
	volume={37},
	number={4},
	pages={5018--5026},
	year={2023}
}

@article{rissanen1996fisher,
	title={Fisher information and stochastic complexity},
	author={Rissanen, Jorma J},
	journal={IEEE transactions on information theory},
	volume={42},
	number={1},
	pages={40--47},
	year={1996},
	publisher={IEEE}
}

@article{barrada2009metodologia,
	title={METODOLOG{\'I}A: Comparison of methods for controlling maximum exposure rates in computerized adaptive testing},
	author={Barrada, Juan Ram{\'o}n and Abad, Francisco Jos{\'e} and Veldkamp, Bernard P},
	journal={Psicothema},
	pages={313--320},
	year={2009}
}

@article{chang1999stratified,
	title={A-stratified multistage computerized adaptive testing},
	author={Chang, Hua-Hua and Ying, Zhiliang},
	journal={Applied Psychological Measurement},
	volume={23},
	number={3},
	pages={211--222},
	year={1999},
	publisher={Sage Publications Sage CA: Thousand Oaks, CA}
}

@article{henson2005test,
	title={Test construction for cognitive diagnosis},
	author={Henson, Robert and Douglas, Jeff},
	journal={Applied Psychological Measurement},
	volume={29},
	number={4},
	pages={262--277},
	year={2005},
	publisher={Sage Publications Sage CA: Thousand Oaks, CA}
}

@article{tatsuoka2002data,
	title={Data analytic methods for latent partially ordered classification models},
	author={Tatsuoka, Curtis},
	journal={Journal of the Royal Statistical Society Series C: Applied Statistics},
	volume={51},
	number={3},
	pages={337--350},
	year={2002},
	publisher={Oxford University Press}
}

@article{zheng2018information,
	title={The information product methods: A unified approach to dual-purpose computerized adaptive testing},
	author={Zheng, Chanjin and He, Guanrui and Gao, Chunlei},
	journal={Applied Psychological Measurement},
	volume={42},
	number={4},
	pages={321--324},
	year={2018},
	publisher={SAGE Publications Sage CA: Los Angeles, CA}
}

@article{kang2017dual,
	title={Dual-Objective Item Selection Criteria in Cognitive Diagnostic Computerized Adaptive Testing},
	author={Kang, Hyeon-Ah and Zhang, Susu and Chang, Hua-Hua},
	journal={Journal of Educational Measurement},
	volume={54},
	number={2},
	pages={165--183},
	year={2017},
	publisher={Wiley Online Library}
}

@article{dai2016exploration,
	title={Exploration of item selection in dual-purpose cognitive diagnostic computerized adaptive testing: Based on the RRUM},
	author={Dai, Buyun and Zhang, Minqiang and Li, Guangming},
	journal={Applied Psychological Measurement},
	volume={40},
	number={8},
	pages={625--640},
	year={2016},
	publisher={SAGE Publications Sage CA: Los Angeles, CA}
}

@inproceedings{yoo2019learning,
	title={Learning loss for active learning},
	author={Yoo, Donggeun and Kweon, In So},
	booktitle={Proceedings of the IEEE/CVF conference on computer vision and pattern recognition},
	pages={93--102},
	year={2019}
}

@inproceedings{ghorbani2022data,
	title={Data shapley valuation for efficient batch active learning},
	author={Ghorbani, Amirata and Zou, James and Esteva, Andre},
	booktitle={2022 56th Asilomar Conference on Signals, Systems, and Computers},
	pages={1456--1462},
	year={2022},
	organization={IEEE}
}

@inproceedings{wang2020dual,
	title={Dual adversarial network for deep active learning},
	author={Wang, Shuo and Li, Yuexiang and Ma, Kai and Ma, Ruhui and Guan, Haibing and Zheng, Yefeng},
	booktitle={Computer Vision--ECCV 2020: 16th European Conference, Glasgow, UK, August 23--28, 2020, Proceedings, Part XXIV 16},
	pages={680--696},
	year={2020},
	organization={Springer}
}

@incollection{deb2011multi,
	title={Multi-objective optimisation using evolutionary algorithms: an introduction},
	author={Deb, Kalyanmoy},
	booktitle={Multi-objective evolutionary optimisation for product design and manufacturing},
	pages={3--34},
	year={2011},
	publisher={Springer}
}

@inproceedings{mujtaba2021multi,
	title={Multi-objective optimization of item selection in computerized adaptive testing},
	author={Mujtaba, Dena F and Mahapatra, Nihar R},
	booktitle={Proceedings of the Genetic and Evolutionary Computation Conference},
	pages={1018--1026},
	year={2021}
}

@inproceedings{zhuang2023bounded,
	title={A Bounded Ability Estimation for Computerized Adaptive Testing},
	author={Zhuang, Yan and Liu, Qi and Zhao, GuanHao and Huang, Zhenya and Huang, Weizhe and Pardos, Zachary and Chen, Enhong and Wu, Jinze and Li, Xin},
	booktitle={Thirty-seventh Conference on Neural Information Processing Systems},
	year={2023}
}

@article{liu2013theory,
	title={Theory of the self-learning Q-matrix},
	author={Liu, Jingchen and Xu, Gongjun and Ying, Zhiliang},
	journal={Bernoulli: official journal of the Bernoulli Society for Mathematical Statistics and Probability},
	volume={19},
	number={5A},
	pages={1790},
	year={2013},
	publisher={NIH Public Access}
}

@inproceedings{sun2014alternating,
	title={Alternating recursive method for Q-matrix learning},
	author={Sun, Yuan and Ye, Shiwei and Inoue, Shunya and Sun, Yi},
	booktitle={Educational Data Mining 2014},
	year={2014}
}

@article{xiong2022data,
	title={Data-driven Q-matrix learning based on Boolean matrix factorization in cognitive diagnostic assessment},
	author={Xiong, Jianhua and Luo, Zhaosheng and Luo, Guanzhong and Yu, Xiaofeng},
	journal={British Journal of Mathematical and Statistical Psychology},
	volume={75},
	number={3},
	pages={638--667},
	year={2022},
	publisher={Wiley Online Library}
}

@inproceedings{sympson1985controlling,
	title={Controlling item-exposure rates in computerized adaptive testing},
	author={Sympson, JB and Hetter, RD},
	booktitle={Proceedings of the 27th annual meeting of the Military Testing Association},
	pages={973--977},
	year={1985}
}

@article{van2004constraining,
	title={Constraining item exposure in computerized adaptive testing with shadow tests},
	author={van der Linden, Wim J and Veldkamp, Bernard P},
	journal={Journal of Educational and Behavioral Statistics},
	volume={29},
	number={3},
	pages={273--291},
	year={2004},
	publisher={Sage Publications Sage CA: Los Angeles, CA}
}

@article{van2007conditional,
	title={Conditional item-exposure control in adaptive testing using item-ineligibility probabilities},
	author={van der Linden, Wim J and Veldkamp, Bernard P},
	journal={Journal of Educational and Behavioral Statistics},
	volume={32},
	number={4},
	pages={398--418},
	year={2007},
	publisher={Sage Publications Sage CA: Thousand Oaks, CA}
}

@article{barrada2009multiple,
	title={Multiple maximum exposure rates in computerized adaptive testing},
	author={Barrada, Juan Ramon and Veldkamp, Bernard P and Olea, Julio},
	journal={Applied Psychological Measurement},
	volume={33},
	number={1},
	pages={58--73},
	year={2009},
	publisher={Sage Publications Sage CA: Los Angeles, CA}
}

@inproceedings{yi2001stratified,
	title={a-Stratified computerized adaptive testing with content blocking},
	author={Yi, Q and Chang, H},
	booktitle={Annual Meeting of the Psychometric Society, King of Prussia, PA},
	year={2001}
}

@article{barrada2006maximum,
	title={Maximum information stratification method for controlling item exposure in computerized adaptive testing},
	author={Barrada, Juan Ram{\'o}n and Mazuela, Paloma and Olea, Julio},
	journal={Psicothema},
	volume={18},
	number={1},
	pages={156--159},
	year={2006},
	publisher={Colegio Oficial de Psic{\'o}logos del Principado de Asturias}
}

@article{barrada2014optimal,
	title={Optimal number of strata for the stratified methods in computerized adaptive testing},
	author={Barrada, Juan Ram{\'o}n and Abad, Francisco Jos{\'e} and Olea, Julio},
	journal={The Spanish Journal of Psychology},
	volume={17},
	pages={E48},
	year={2014},
	publisher={Cambridge University Press}
}

@article{chang2001stratified,
	title={a-Stratified multistage computerized adaptive testing with b blocking},
	author={Chang, Hua-Hua and Qian, Jiahe and Ying, Zhiliang},
	journal={Applied Psychological Measurement},
	volume={25},
	number={4},
	pages={333--341},
	year={2001},
	publisher={Sage Publications Sage CA: Thousand Oaks, CA}
}

@article{magno2009demonstrating,
	title={Demonstrating the difference between classical test theory and item response theory using derived test data},
	author={Magno, Carlo},
	journal={The international Journal of Educational and Psychological assessment},
	volume={1},
	number={1},
	pages={1--11},
	year={2009}
}

@inproceedings{liu2022learning,
	title={Learning Evidential Cognitive Diagnosis Networks Robust to Response Bias},
	author={Liu, Jiawei and Hou, Jingyi and Zhang, Na and Liu, Zhijie and He, Wei},
	booktitle={CAAI International Conference on Artificial Intelligence},
	pages={171--181},
	year={2022},
	organization={Springer}
}

@article{green1984technical,
	title={Technical guidelines for assessing computerized adaptive tests},
	author={Green, Bert F and Bock, R Darrell and Humphreys, Lloyd G and Linn, Robert L and Reckase, Mark D},
	journal={Journal of Educational measurement},
	volume={21},
	number={4},
	pages={347--360},
	year={1984},
	publisher={Wiley Online Library}
}

@article{brigman1976multiple,
	title={Multiple Test Equating Using the Rasch Model.},
	author={Brigman, S Leellen and Bashaw, WL},
	year={1976},
	publisher={ERIC}
}

@article{van2000test,
	title={A test-theoretic approach to observed-score equating},
	author={van der Linden, Wim J},
	journal={Psychometrika},
	volume={65},
	number={4},
	pages={437--456},
	year={2000},
	publisher={Springer}
}

@article{sawaki2001comparability,
	title={Comparability of conventional and computerized tests of reading in a second language},
	author={Sawaki, Yasuyo},
	year={2001},
	publisher={University of Hawaii National Foreign Language Resource Center}
}

@article{jansubpopulation,
	title={Subpopulation Differences In Equating Computerized Adaptive And Paper-and-pencil Versions of The ASVAB},
	author={JAN, ELO},
	publisher={Citeseer}
}

@article{li2021improved,
	title={Improved regularization and robustness for fine-tuning in neural networks},
	author={Li, Dongyue and Zhang, Hongyang},
	journal={Advances in Neural Information Processing Systems},
	volume={34},
	pages={27249--27262},
	year={2021}
}

@article{wang2021regularizing,
	title={Regularizing deep networks with semantic data augmentation},
	author={Wang, Yulin and Huang, Gao and Song, Shiji and Pan, Xuran and Xia, Yitong and Wu, Cheng},
	journal={IEEE Transactions on Pattern Analysis and Machine Intelligence},
	volume={44},
	number={7},
	pages={3733--3748},
	year={2021},
	publisher={IEEE}
}

@inproceedings{dong2020benchmarking,
	title={Benchmarking adversarial robustness on image classification},
	author={Dong, Yinpeng and Fu, Qi-An and Yang, Xiao and Pang, Tianyu and Su, Hang and Xiao, Zihao and Zhu, Jun},
	booktitle={proceedings of the IEEE/CVF conference on computer vision and pattern recognition},
	pages={321--331},
	year={2020}
}

@article{kariyappa2019improving,
	title={Improving adversarial robustness of ensembles with diversity training},
	author={Kariyappa, Sanjay and Qureshi, Moinuddin K},
	journal={arXiv preprint arXiv:1901.09981},
	year={2019}
}

@book{baker2004item,
	title={Item response theory: Parameter estimation techniques},
	author={Baker, Frank B and Kim, Seock-Ho},
	year={2004},
	publisher={CRC press}
}

@article{yurtcu2021bibliometric,
	title={Bibliometric analysis of articles on computerized adaptive testing},
	author={Yurtcu, Meltem and G{\"U}ZELLER, Cem},
	journal={Participatory Educational Research},
	volume={8},
	number={4},
	pages={426--438},
	year={2021},
	publisher={{\"O}zgen KORKMAZ}
}

@article{gibbons2016computerized,
	title={Computerized adaptive diagnosis and testing of mental health disorders},
	author={Gibbons, Robert D and Weiss, David J and Frank, Ellen and Kupfer, David},
	journal={Annual review of clinical psychology},
	volume={12},
	number={1},
	pages={83--104},
	year={2016},
	publisher={Annual Reviews}
}

@article{gibbons2013computerized,
	title={The computerized adaptive diagnostic test for major depressive disorder (CAD-MDD): a screening tool for depression},
	author={Gibbons, Robert D and Hooker, Giles and Finkelman, Matthew D and Weiss, David J and Pilkonis, Paul A and Frank, Ellen and Moore, Tara and Kupfer, David J},
	journal={The Journal of clinical psychiatry},
	volume={74},
	number={7},
	pages={3579},
	year={2013},
	publisher={Physicians Postgraduate Press, Inc.}
}

@article{gibbons2017development,
	title={Development of a computerized adaptive test suicide scale—The CAT-SS},
	author={Gibbons, Robert D and Kupfer, David and Frank, Ellen and Moore, Tara and Beiser, David G and Boudreaux, Edwin D},
	journal={The Journal of clinical psychiatry},
	volume={78},
	number={9},
	pages={3581},
	year={2017},
	publisher={Physicians Postgraduate Press, Inc.}
}

@article{ando2018validity,
	title={Validity and reliability of computerized adaptive test of soccer tactical skill},
	author={Ando, Kozue and Mishio, Shota and Nishijima, Takahiko},
	journal={Football Science},
	volume={15},
	pages={38--51},
	year={2018},
	publisher={Japanese Society of Science and Football}
}

@article{montgomery2013computerized,
	title={Computerized adaptive testing for public opinion surveys},
	author={Montgomery, Jacob M and Cutler, Josh},
	journal={Political Analysis},
	volume={21},
	number={2},
	pages={172--192},
	year={2013},
	publisher={Cambridge University Press}
}

@book{crocker1986introduction,
	title={Introduction to classical and modern test theory.},
	author={Crocker, Linda and Algina, James},
	year={1986},
	publisher={ERIC}
}

@book{van2010elements,
	title={Elements of adaptive testing},
	author={Van der Linden, Wim J and Glas, Cees AW},
	volume={10},
	year={2010},
	publisher={Springer}
}

@article{feng2009addressing,
	title={Addressing the assessment challenge with an online system that tutors as it assesses},
	author={Feng, Mingyu and Heffernan, Neil and Koedinger, Kenneth},
	journal={User modeling and user-adapted interaction},
	volume={19},
	pages={243--266},
	year={2009},
	publisher={Springer}
}

@inproceedings{chang2015modeling,
	title={Modeling Exercise Relationships in E-Learning: A Unified Approach.},
	author={Chang, Haw-Shiuan and Hsu, Hwai-Jung and Chen, Kuan-Ta},
	booktitle={EDM},
	pages={532--535},
	year={2015}
}

@inproceedings{choi2020ednet,
	title={Ednet: A large-scale hierarchical dataset in education},
	author={Choi, Youngduck and Lee, Youngnam and Shin, Dongmin and Cho, Junghyun and Park, Seoyon and Lee, Seewoo and Baek, Jineon and Bae, Chan and Kim, Byungsoo and Heo, Jaewe},
	booktitle={Artificial Intelligence in Education: 21st International Conference, AIED 2020, Ifrane, Morocco, July 6--10, 2020, Proceedings, Part II 21},
	pages={69--73},
	year={2020},
	organization={Springer}
}

@article{veldkamp2019robust,
	title={Robust computerized adaptive testing},
	author={Veldkamp, Bernard P and Verschoor, Angela J},
	journal={Theoretical and practical advances in computer-based educational measurement},
	pages={291--305},
	year={2019},
	publisher={Springer International Publishing}
}

@inproceedings{dong2023towards,
	title={Towards Next-Generation Intelligent Assistants Leveraging LLM Techniques},
	author={Dong, Xin Luna and Moon, Seungwhan and Xu, Yifan Ethan and Malik, Kshitiz and Yu, Zhou},
	booktitle={Proceedings of the 29th ACM SIGKDD Conference on Knowledge Discovery and Data Mining},
	pages={5792--5793},
	year={2023}
}

@article{chang2023survey,
	title={A survey on evaluation of large language models},
	author={Chang, Yupeng and Wang, Xu and Wang, Jindong and Wu, Yuan and Zhu, Kaijie and Chen, Hao and Yang, Linyi and Yi, Xiaoyuan and Wang, Cunxiang and Wang, Yidong and others},
	journal={arXiv preprint arXiv:2307.03109},
	year={2023}
}

@inproceedings{yu2023sacat,
	title={SACAT: Student-Adaptive Computerized Adaptive Testing},
	author={Yu, Jingwei and Zhenyu, Mu and Lei, Jiayi and Yin, Li'Ang and Xia, Wei and Yu, Yong and Long, Ting},
	booktitle={The Fifth International Conference on Distributed Artificial Intelligence},
	pages={1--7},
	year={2023}
}

@article{an2014item,
	title={Item response theory: What it is and how you can use the IRT procedure to apply it},
	author={An, Xinming and Yung, Yiu-Fai},
	journal={SAS Institute Inc. SAS364-2014},
	volume={10},
	number={4},
	pages={1--14},
	year={2014},
	publisher={Citeseer}
}

@book{miller2002subset,
	title={Subset selection in regression},
	author={Miller, Alan},
	year={2002},
	publisher={CRC Press}
}

@inproceedings{bhandari23evaluating,
	title={Evaluating ChatGPT-generated Textbook Questions using IRT},
	author={Bhandari, Shreya and Liu, Yunting and Pardos, Zachary A},
	booktitle={Generative AI for Education Workshop (GAIED) at the Thirty-seventh Conference on Neural Information Processing Systems},
	year={2023}
}

@inproceedings{drousiotis2021capturing,
	title={Capturing fairness and uncertainty in student dropout prediction--a comparison study},
	author={Drousiotis, Efthyvoulos and Pentaliotis, Panagiotis and Shi, Lei and Cristea, Alexandra I},
	booktitle={International Conference on Artificial Intelligence in Education},
	pages={139--144},
	year={2021},
	organization={Springer}
}

@inproceedings{shi2014multifaceted,
	title={Multifaceted open social learner modelling},
	author={Shi, Lei and Cristea, Alexandra I and Hadzidedic, Suncica},
	booktitle={Advances in Web-Based Learning--ICWL 2014: 13th International Conference, Tallinn, Estonia, August 14-17, 2014. Proceedings 13},
	pages={32--42},
	year={2014},
	organization={Springer}
}

@inproceedings{xiao2022fine,
	title={Fine-grained main ideas extraction and clustering of online course reviews},
	author={Xiao, Chenghao and Shi, Lei and Cristea, Alexandra and Li, Zhaoxing and Pan, Ziqi},
	booktitle={International Conference on Artificial Intelligence in Education},
	pages={294--306},
	year={2022},
	organization={Springer}
}

@article{mehrabi2021survey,
	title={A survey on bias and fairness in machine learning},
	author={Mehrabi, Ninareh and Morstatter, Fred and Saxena, Nripsuta and Lerman, Kristina and Galstyan, Aram},
	journal={ACM computing surveys (CSUR)},
	volume={54},
	number={6},
	pages={1--35},
	year={2021},
	publisher={ACM New York, NY, USA}
}

@article{kipnis2024texttt,
	title={metabench--A Sparse Benchmark to Measure General Ability in Large Language Models},
	author={Kipnis, Alex and Voudouris, Konstantinos and Buschoff, Luca M Schulze and Schulz, Eric},
	journal={arXiv preprint arXiv:2407.12844},
	year={2024}
}

@inproceedings{otani2016irt,
	title={IRT-based aggregation model of crowdsourced pairwise comparison for evaluating machine translations},
	author={Otani, Naoki and Nakazawa, Toshiaki and Kawahara, Daisuke and Kurohashi, Sadao},
	booktitle={Proceedings of the 2016 Conference on Empirical Methods in Natural Language Processing},
	pages={511--520},
	year={2016}
}

@article{liu2024leveraging,
	title={Leveraging LLM-Respondents for Item Evaluation: a Psychometric Analysis},
	author={Liu, Yunting and Bhandari, Shreya and Pardos, Zachary A},
	journal={arXiv preprint arXiv:2407.10899},
	year={2024}
}

@inproceedings{lalor2016building,
	title={Building an evaluation scale using item response theory},
	author={Lalor, John P and Wu, Hao and Yu, Hong},
	booktitle={Proceedings of the Conference on Empirical Methods in Natural Language Processing. Conference on Empirical Methods in Natural Language Processing},
	volume={2016},
	pages={648},
	year={2016},
	organization={NIH Public Access}
}

@inproceedings{sedoc2020item,
	title={Item response theory for efficient human evaluation of chatbots},
	author={Sedoc, Jo{\~a}o and Ungar, Lyle},
	booktitle={Proceedings of the First Workshop on Evaluation and Comparison of NLP Systems},
	pages={21--33},
	year={2020}
}

@misc{wang2023evaluating,
	title={Evaluating General-Purpose AI with Psychometrics}, 
	author={Xiting Wang and Liming Jiang and Jose Hernandez-Orallo and David Stillwell and Luning Sun and Fang Luo and Xing Xie},
	year={2023},
	eprint={2310.16379},
	archivePrefix={arXiv},
	primaryClass={cs.AI}
}

@article{eignor1993case,
	title={CASE STUDIES IN COMPUTER ADAPTIVE TEST DESIGN THROUGH SIMULATION 1, 2},
	author={Eignor, Daniel R and Stocking, Martha L and Way, Walter D and Steffen, Manfred},
	journal={ETS Research Report Series},
	volume={1993},
	number={2},
	pages={i--41},
	year={1993},
	publisher={Wiley Online Library}
}

@inproceedings{polo2024tinybenchmarks,
	title={tinyBenchmarks: evaluating LLMs with fewer examples},
	author={Polo, Felipe Maia and Weber, Lucas and Choshen, Leshem and Sun, Yuekai and Xu, Gongjun and Yurochkin, Mikhail},
	booktitle={Forty-first International Conference on Machine Learning},
	year={2024}
}

@inproceedings{zhuang2025position,
	title={Position: {AI} Evaluation Should Learn from How We Test Humans},
	author={Yan Zhuang and Qi Liu and Zachary Pardos and Patrick C. Kyllonen and Jiyun Zu and Zhenya Huang and Shijin Wang and Enhong Chen},
	booktitle={Forty-second International Conference on Machine Learning Position Paper Track},
	year={2025}
}

@inproceedings{hendrycks2021measuring,
	title={Measuring Massive Multitask Language Understanding},
	author={Dan Hendrycks and Collin Burns and Steven Basart and Andy Zou and Mantas Mazeika and Dawn Song and Jacob Steinhardt},
	booktitle={International Conference on Learning Representations},
	year={2021}
}

@inproceedings{guinetautomated,
	title={Automated Evaluation of Retrieval-Augmented Language Models with Task-Specific Exam Generation},
	author={Guinet, Gauthier and Omidvar-Tehrani, Behrooz and Deoras, Anoop and Callot, Laurent},
	booktitle={Forty-first International Conference on Machine Learning}
}

@inproceedings{yu2024a,
	title={A Unified Adaptive Testing System Enabled by Hierarchical Structure Search},
	author={Junhao Yu and Yan Zhuang and Zhenya Huang and Qi Liu and Xin Li and Rui LI and Enhong Chen},
	booktitle={Forty-first International Conference on Machine Learning},
	year={2024}
}

@incollection{martinez2016making,
	title={Making sense of item response theory in machine learning},
	author={Mart{\'\i}nez-Plumed, Fernando and Prud{\^e}ncio, Ricardo BC and Mart{\'\i}nez-Us{\'o}, Adolfo and Hern{\'a}ndez-Orallo, Jos{\'e}},
	booktitle={ECAI 2016},
	pages={1140--1148},
	year={2016},
	publisher={IOS Press}
}

@article{MARTINEZPLUMED201918,
	title = {Item response theory in AI: Analysing machine learning classifiers at the instance level},
	journal = {Artificial Intelligence},
	volume = {271},
	pages = {18-42},
	year = {2019},
	issn = {0004-3702},
	author = {Fernando Martínez-Plumed and Ricardo B.C. Prudêncio and Adolfo Martínez-Usó and José Hernández-Orallo}
}
\vspace{-36pt}
\begin{IEEEbiography}[{\includegraphics[width=1in,height=1.25in,clip,keepaspectratio]{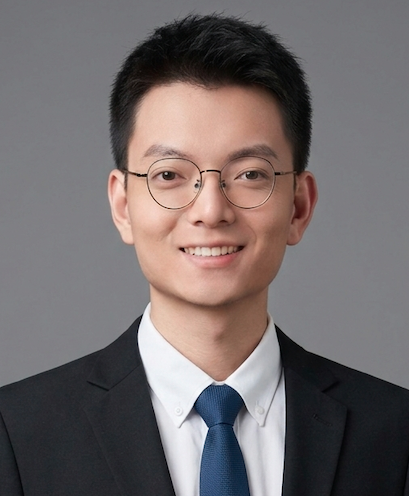}}]{Yan Zhuang} received the Ph.D. degree from the University of Science and Technology of China (USTC), in 2025. He is currently an Associate Professor with Nanjing University of Aeronautics and Astronautics. His main research interests include data mining and intelligent education systems. He has published more than 20 papers in top conferences and journals such as NeurIPS, ICML, ICLR, AAAI, and IEEE TPAMI. He received the Best Paper Runner-Up Award at CIKM 2023.
\end{IEEEbiography}

\vspace{-36pt}
\begin{IEEEbiography}[{\includegraphics[width=1in,height=1.25in,clip,keepaspectratio]{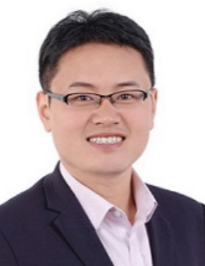}}]{Qi Liu}(Member, IEEE)
	received the Ph.D. degree from the University of Science and Technology of China (USTC), in 2013. He is currently a Professor with USTC. His general research areas include data mining and knowledge discovery, and artificial intelligence. His research is supported by the National Science Fund for Excellent Young Scholars and the Youth Innovation Promotion Association of Chinese Academy of Sciences. He has published more than 100 papers in refereed journals and conference proceedings, such as TKDE, TOIS, TNNLS, NeurIPS, ICML, ICLR, and KDD. Dr. Liu  is the recipient of the KDD 2018 Best Student Paper Award (Research) and the ICDM 2011 Best Research Paper Award.
\end{IEEEbiography}
\vspace{-36pt}
\begin{IEEEbiography}[{\includegraphics[width=1in,height=1.25in,clip,keepaspectratio]{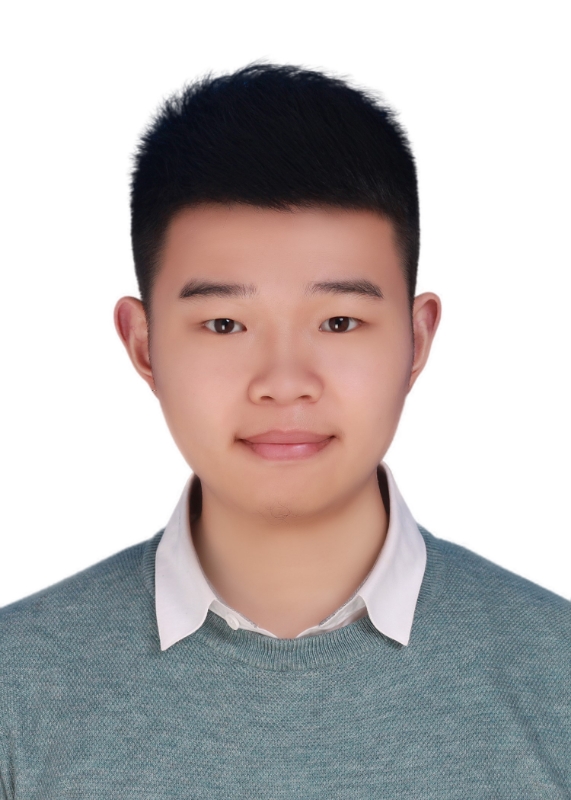}}]{Haoyang Bi}
	received the B.E. degree in computer science and technology from University of
	Science and Technology of China (USTC), Hefei, China, in 2019. He is currently a Ph.D. student in the School of Computer Science and Technology at University of Science and Technology of China (USTC), China. His research interests include active learning, Bayesian learning and meta-learning.
\end{IEEEbiography}
\vspace{-36pt}
\begin{IEEEbiography}[{\includegraphics[width=1in,height=1.25in,clip,keepaspectratio]{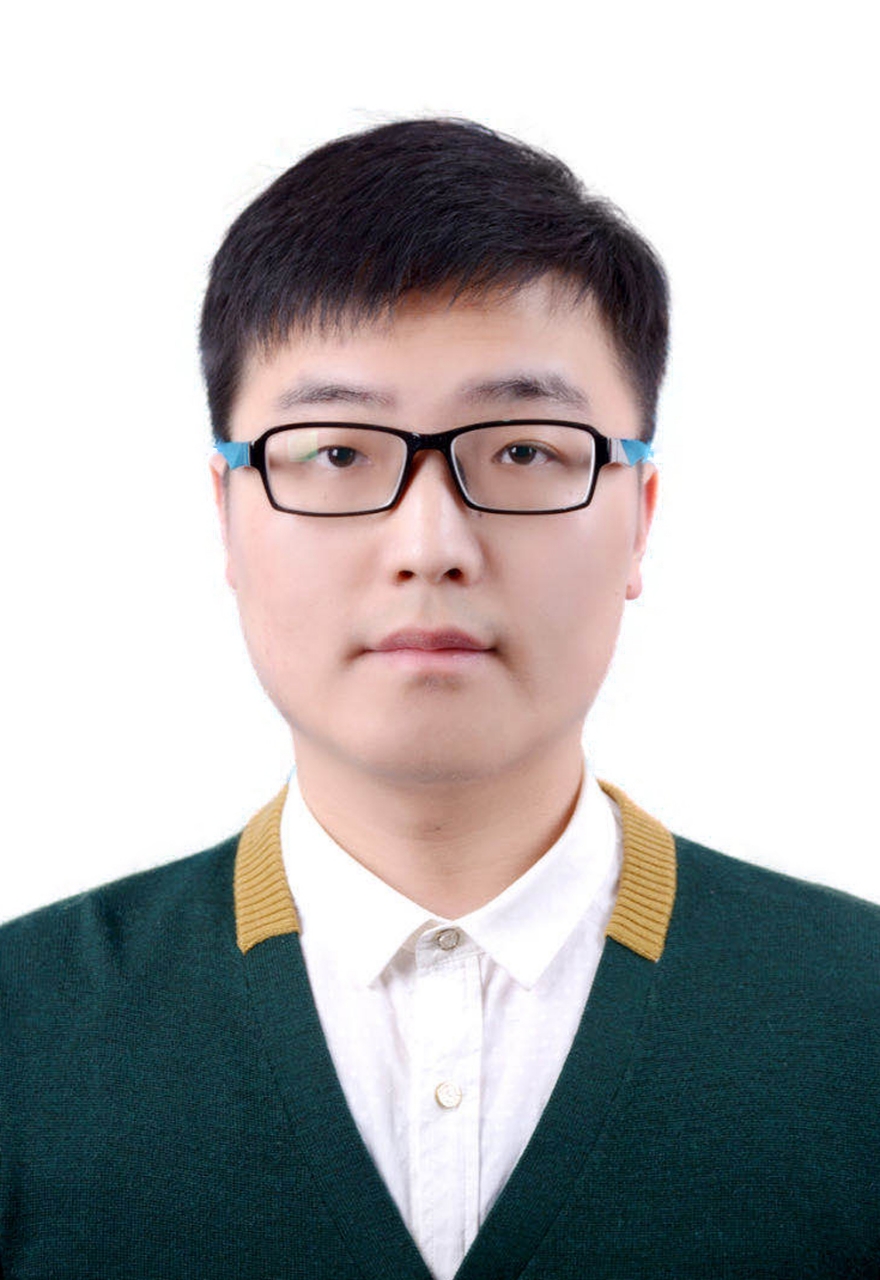}}]{Zhenya Huang}(Member, IEEE)
	received the Ph.D. degree from the University of Science and Technology of China (USTC), in 2020. He is currently an Associate Professor with USTC. His main research interests include artificial intelligence, knowledge reasoning, and intelligent education. He has published more than 50 papers in refereed journals and conference proceedings, including TKDE, TOIS, TNNLS, AAAI, KDD, SIGIR, and ICDM. Dr. Huang has served regularly on the program committee of numerous conferences and is a reviewer for the leading academic journals.
\end{IEEEbiography}
\vspace{-36pt}
\begin{IEEEbiography}[{\includegraphics[width=1in,height=1.25in,clip,keepaspectratio]{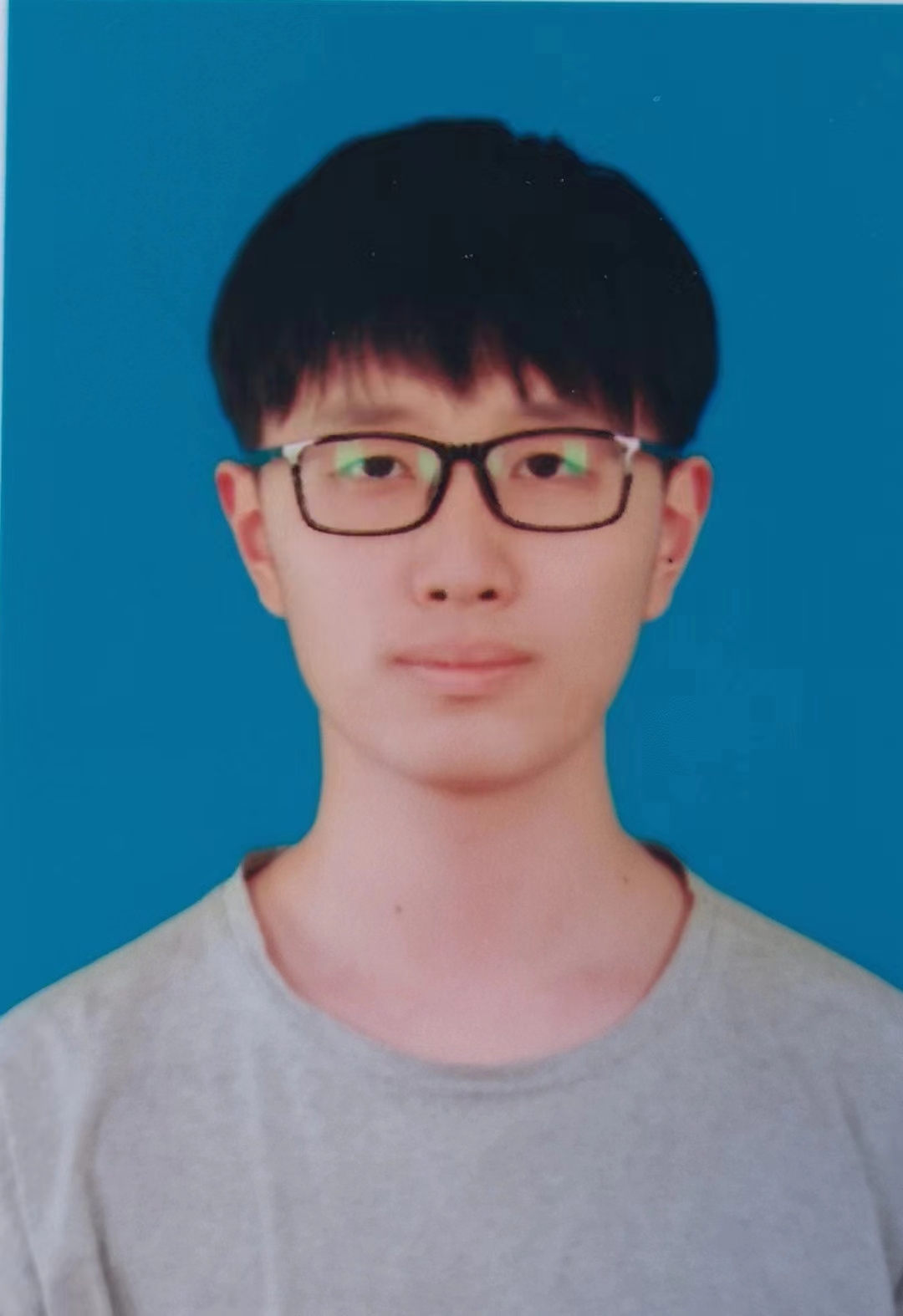}}]{Weizhe Huang}
	received his Bachelor's degree in computer science from University of Science and Technology of China (USTC) in 2022. He is currently pursuing a Master's degree at USTC. His research interests include sequence modeling, computerized adaptive testing, and educational data mining.
\end{IEEEbiography}
\vspace{-36pt}
\begin{IEEEbiography}[{\includegraphics[width=1in,height=1.25in,clip,keepaspectratio]{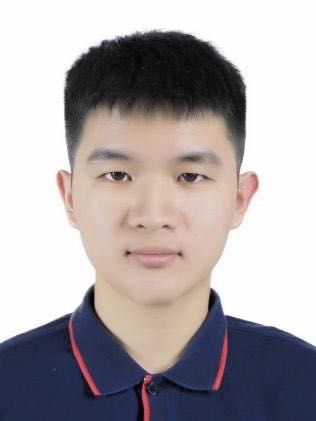}}]{Jiatong Li}
	received his BS degree from University of Science and Technology of China (USTC). He is currently working toward the master degree in School of Artificial Intelligence and Data Science, USTC. His research interests include educational data mining, trustworthy AI and model evaluation. His works in educational data mining have been published in major conference in related fields such as KDD, WWW, etc.
\end{IEEEbiography}
\vspace{-36pt}
\begin{IEEEbiography}[{\includegraphics[width=1in,height=1.25in,clip,keepaspectratio]{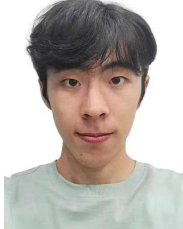}}]{Junhao Yu}
	He is currently working toward the master degree at the University of Science and Technology of China. His main research interests include artificial intelligence, large language models, data mining, and adaptive testing.
\end{IEEEbiography}
\vspace{-36pt}
\begin{IEEEbiography}[{\includegraphics[width=1in,height=1.25in,clip,keepaspectratio]{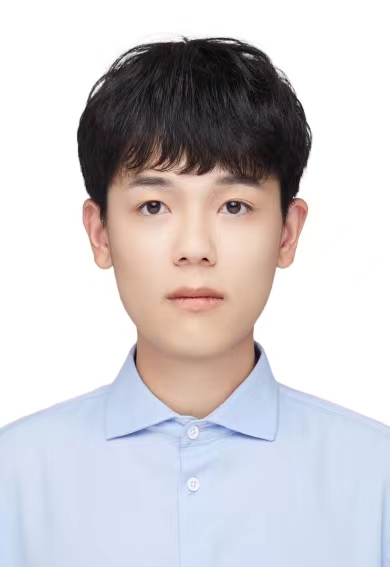}}]{Zirui Liu}
is master student in the University of Science and Technology of China (USTC). His main research interests include data mining and intelligent education. 
\end{IEEEbiography}

\vspace{-36pt}
\begin{IEEEbiography}[{\includegraphics[width=1in,height=1.25in,clip,keepaspectratio]{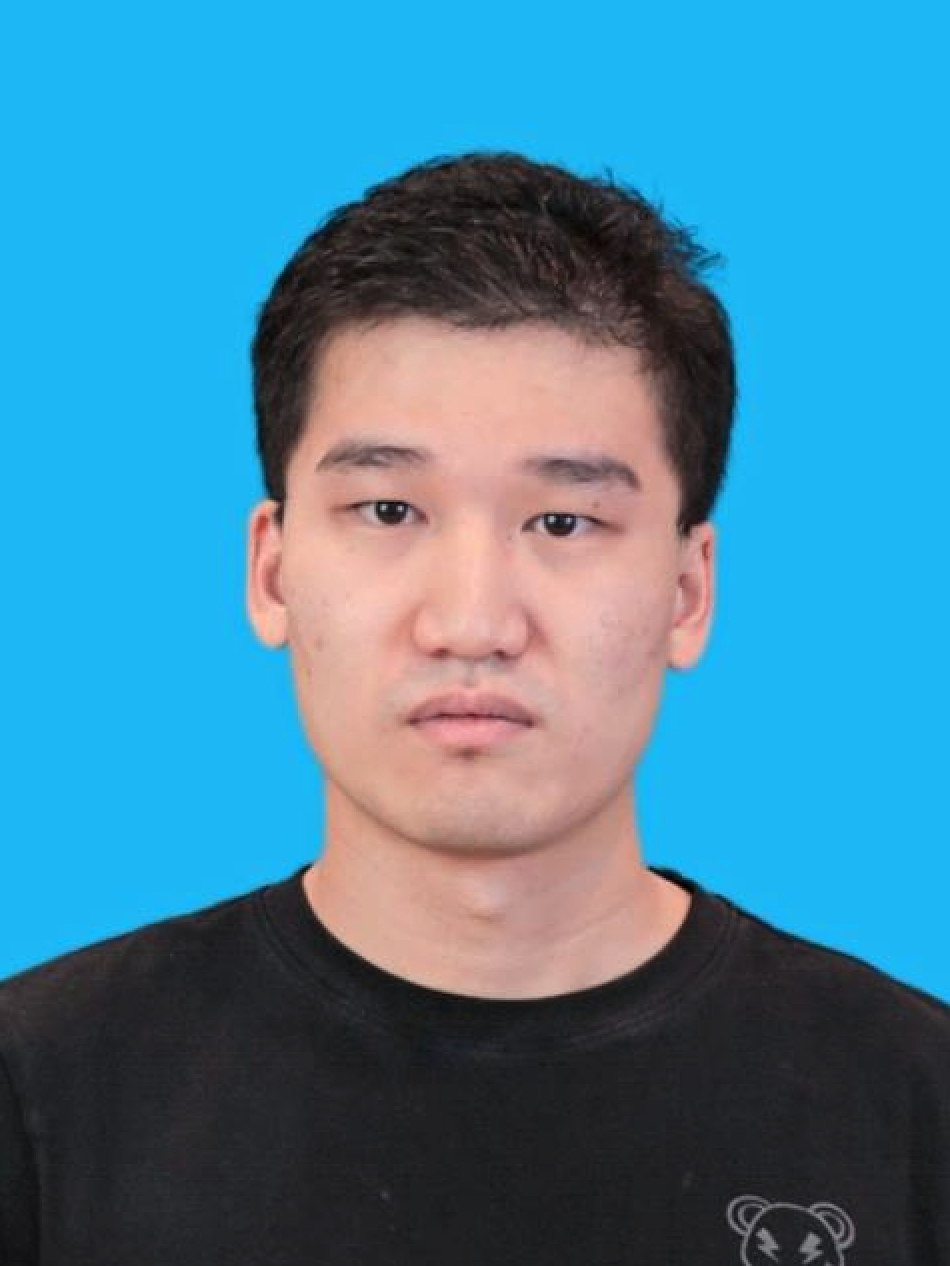}}]{Zirui Hu}
	received his master's degree from the University of Science and Technology of China (USTC). His research interests include fairness in recommender systems, causal inference, and intelligent education. He has published his work on fair learning in major conferences in these fields, such as DASFAA and KSEM, etc.
\end{IEEEbiography}
\vspace{-36pt}
\begin{IEEEbiography}[{\includegraphics[width=1in,height=1.25in,clip,keepaspectratio]{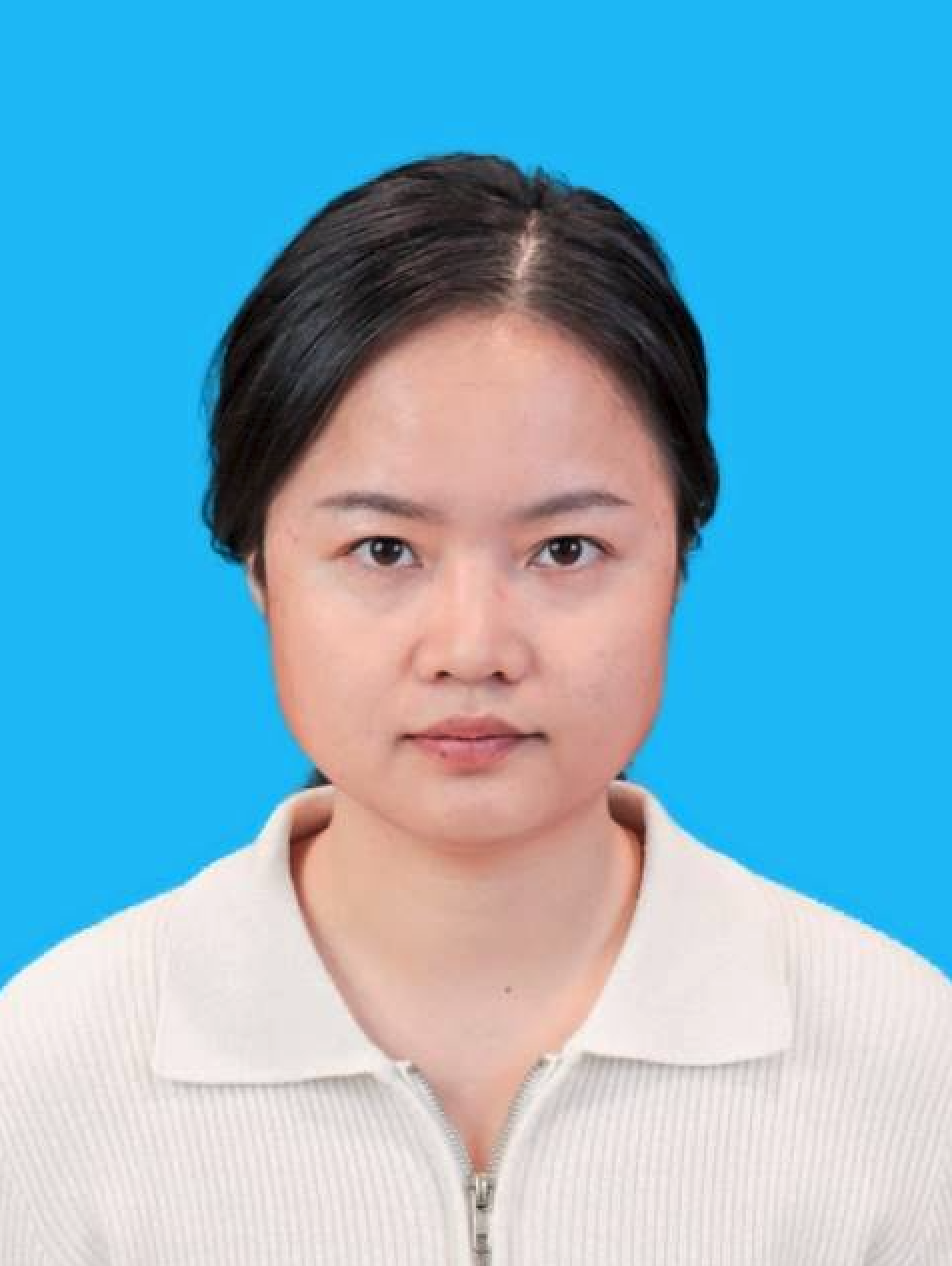}}]{Yuting Hong}
received the Masters' degree from the University of Science and Technology of China (USTC), in 2024. Her work in Computerized Adaptive Testing has been published in CIKM and received the Best Paper Runner-Up on CIKM 2023.
\end{IEEEbiography}
\vspace{-36pt}
\begin{IEEEbiography}[{\includegraphics[width=1in,height=1.25in,clip,keepaspectratio]{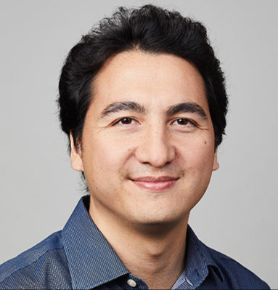}}]{Zachary A. Pardos}
	 earned his PhD in Computer Science at Worcester Polytechnic Institute. He is an Associate Professor of Education at UC Berkeley studying adaptive learning and AI. His early scholarship focused on formative assessment using Knowledge Tracing, the predominant model used for estimating skill mastery in computer tutoring system contexts. His recent work designing Human-AI collaborations to pave pathways to and within higher education systems has been published in venues such as SIGCHI, AAAI, The Internet and Higher Education, and Science.
\end{IEEEbiography}
\vspace{-36pt}
\begin{IEEEbiography}[{\includegraphics[width=1in,height=1.25in,clip,keepaspectratio]{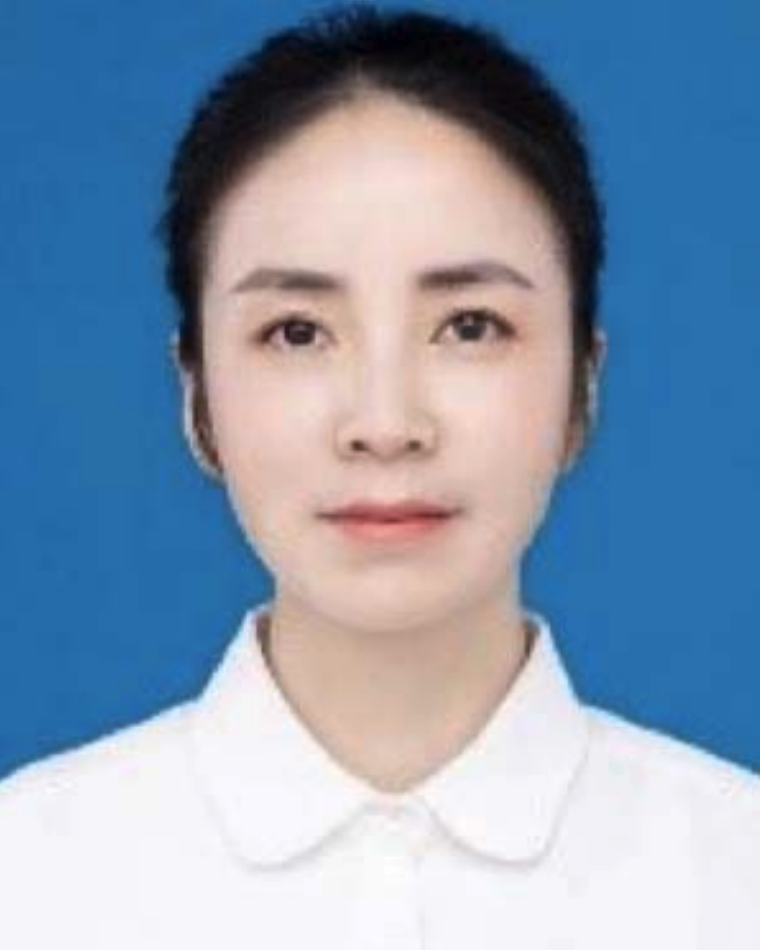}}]{Haiping Ma}
	received the BE degree from Anhui University, Hefei, China, in 2008, and the PhD degree
	from the University of Science and Technology of China, Hefei, China, in 2013. She is currently an associate professor with the Institutes of Physical Science and Information Technology, Anhui University,Hefei, China. Her current research interests include data mining and multi-objective optimization methods and their applications.
\end{IEEEbiography}
\vspace{-36pt}
\begin{IEEEbiography}[{\includegraphics[width=1in,height=1.25in,clip,keepaspectratio]{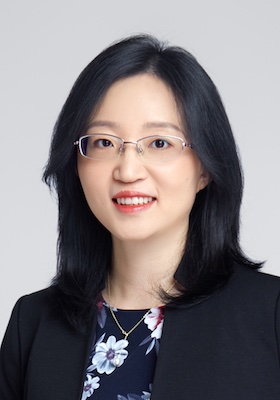}}]{Mengxiao Zhu} (Member, IEEE) received the Ph.D. degree in industrial engineering and management sciences from Northwestern University in 2012. She has been a Distinguished Research Professor at the University of Science and Technology of China (USTC) since 2020. Before joining USTC, she worked as a Research Scientist in the Research and Development division at Educational Testing Service (ETS) for over seven years. She has been leading and involved in multiple NSFC, NSF, and NIH-funded projects in the past 20 years.
\end{IEEEbiography}
\vspace{-26pt}
\begin{IEEEbiography}[{\includegraphics[width=1in,height=1.25in,clip,keepaspectratio]{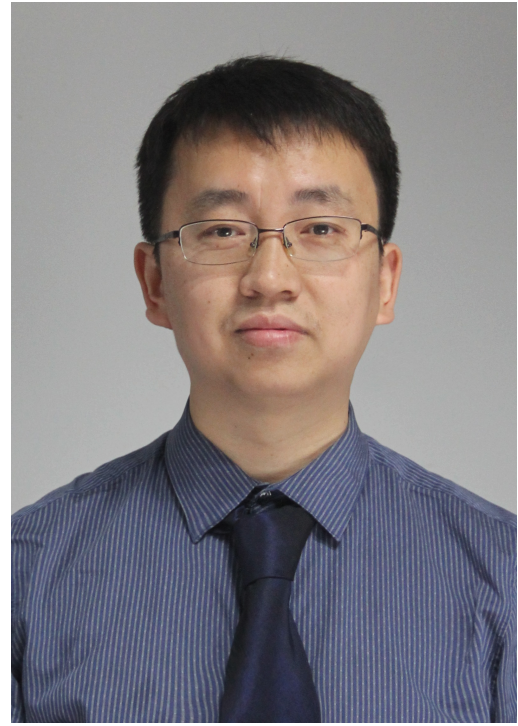}}]{Shijin Wang}
	received the Ph.D. degree from the Institute of Automation, Chinese Academy of Science. He is currently the vice president of IFLYTEK Co., Ltd. and the president of IFLYTEK AI Research (Central China). His research interests include speech and natural language processing. He has published more than 60 papers in refereed conferences such as ACL, KDD, and AAAI. He led the team that won more than ten championships in international technical evaluation such as Blizzard Challenge and CHiME.
\end{IEEEbiography}
\vspace{-26pt}
\begin{IEEEbiography}[{\includegraphics[width=1in,height=1.25in,clip,keepaspectratio]{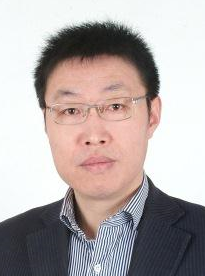}}]{Enhong Chen} (Fellow, IEEE)
	received the Ph.D. degree from the University of Science and Technology of China (USTC), in 1996. He is currently a Professor and the Vice Director of State Key Laboratory of Cognitive Intelligence. His research areas include data mining and machine learning, artificial intelligence. His research is supported by the National Science Foundation for Distinguished Young Scholars of China. He has published more than 200 papers in refereed conferences and journals, including TPAMI, TKDE, TNNLS, TOIS, ICML, NeurIPS, KDD, ICLR and AAAI. He is an associate editor of the IEEE TKDE, IEEE TSMCS, ACM TIST, WWWJ. Dr. Chen received the Best Application Paper Award on KDD 2008, the Best Research Paper Award on ICDM 2011, the Best Student Paper Award on KDD 2018 (Research), and the Best Student Paper Award on KDD 2024 (Research).	

\end{IEEEbiography}

\newpage

\clearpage

\appendix

\section*{Comparison of Fisher Information and KL Information} \figurename\ \ref{fisher_kl} illustrates the KL and Fisher information functions for two distinct questions. For $\theta$ near $\theta_0$, KL Information and Fisher information are always close. If we envision KL Information as a curve, Fisher information corresponds to its curvature (second derivative) at $\theta=\theta_0$. This suggests that Fisher information can be derived from KL Information, but the converse is not true.

	\begin{figure}[h]
		
		\centering
		
		\includegraphics[width=\linewidth]{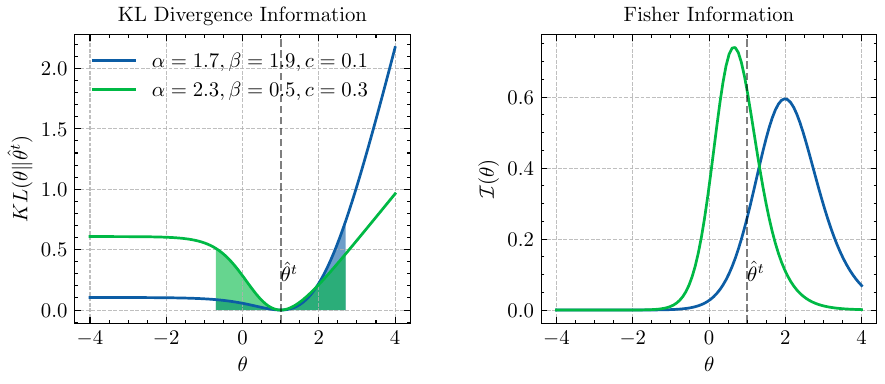}
		\vspace{-12pt}
		\caption{Illustration of KL and Fisher information functions for two questions (Question 1: $\alpha=1.7, \beta=1.9, c=0.1$; Question 2: $\alpha=2.3, \beta=0.5, c=0.3$). Assuming the current proficiency estimate $\hat{\theta}^t=1$. The KL information (left) for the given question represents an integral centered around $\hat{\theta}^t$, while the Fisher information (right) corresponds to the value at the specific point $\hat{\theta}^t$.}
		\label{fisher_kl}
		\vspace{-12pt}
	\end{figure}

\section*{Analysis of Various Key Factors in Testing}
 Table \ref{table:CAT_factors} showcases the underlying causes and advantages of different factors in CAT test control.
		\begin{table*}[h]
	\renewcommand{\arraystretch}{1.2}
	\centering
	\caption{Test Control: Key Factors in CAT Implementation}
	\label{table:CAT_factors}
	\resizebox{1\textwidth}{!}{%
		\begin{tabular}{|m{2.3cm}|m{3cm}|m{3.5cm}|m{4.4cm}|m{1.8cm}|}
			\hline
			\textbf{Factors} & \textbf{Category} & \textbf{Causes} & \textbf{Advantages} & \textbf{Pubs} \\ \hline\hline
			\multirow{1}{*}{\textbf{Exposure Control}} & --& \multirow{1}{*}{Unbalanced question usage} & Mitigates overexposure; \newline  Test security; \newline Comprehensive assessment& \cite{sympson1985controlling,chang1999stratified}\newline\cite{barrada2009multiple, barrada2014optimal} \newline
			\cite{bi2020quality,wang2023gmocat}
			 \\ \hline
			\multirow{7}{*}{\textbf{Fairness}} & Bias in Measurement \newline Models & Skewed training data;\newline Underrepresentation of certain groups & Promotes equitable outcomes;\newline Improves accuracy of proficiency estimation & \multirow{1}{*}{\cite{kizilcec2022algorithmic,Thompson2022Is}}\newline \multirow{1}{*}{\cite{liu2022learning}} \\ \cline{2-5} 
			& \multirow{1}{*}{Bias in question Bank} & \multirow{1}{*}{Unequal applicability;} \newline \multirow{1}{*}{Cultural or regional biases} & Ensures content relevance; \newline Reduces disadvantage for certain groups &  \multirow{1}{*}{\cite{camilli1994methods,hambleton1991fundamentals}}\newline  \multirow{1}{*}{\cite{Roberts2017Stan,chu2013detecting}} \\ \cline{2-5} 
			& Bias in Selection \newline Algorithms & \multirow{1}{*}{Algorithmic preferences} & Reduces disadvantage for certain groups &  \multirow{1}{*}{\cite{lord2012applications}} \\ \cline{2-5} 
			& \multirow{1}{*}{Equating} & Different selected questions across examinees &  Score comparability; \newline Fairness across different tests & \cite{green1984technical,van2000test}\newline\cite{jansubpopulation,sawaki2001comparability} \\ \hline
			\multirow{3}{*}{\textbf{Robustness}} & \multirow{1}{*}{Noise Resistance} & Random variability;\newline Guessing and slipping factors  & Stabilizes estimation; \newline Improves reliability & \multirow{1}{*}{\cite{baker2004item,zhuang2022robust}} \\ \cline{2-5} 
			& Modeling Uncertainty & \multirow{1}{*}{Uncertainty in response} & Improves accuracy of proficiency estimation & \cite{veldkamp2019robust,zhuang2022robust} \\ \hline
			\multirow{1}{*}{\textbf{Search Efficiency}} & \multirow{1}{*}{--} & Large question banks; \newline  Brute-force search& \multirow{1}{*}{Reduces search complexity} & \multirow{1}{*}{\cite{huang2009adaptive,hong2023search}} \\ \hline
		\end{tabular}%
	}
	
\end{table*}

\section*{Comparison of Different Selection Algorithms}
	Table \ref{auc_result} displays some representative methods of each category of selection algorithms and their AUC results on two different datasets. The comparison in this survey focuses on the results at the early testing stage (step=5) and the final testing stage (step=20). It is important to note that the results cannot be directly compared if experimental settings are not standardized. Despite this, the table as a whole reveals that data-driven methods (e.g., reinforcement learning, meta-learning methods)  generally outperform statistical methods. This is because these methods can train and optimize selection algorithms from examinee large-scale response data, while statistical methods simply adhere to fixed functions for selecting questions. The latest subset selection methods do not require training but are remarkably effective. This is primarily because they attempt to explicitly approximate the objectives of CAT and provide theoretical guarantees on estimation errors. Furthermore, it is observed that considering factors within test control, such as robustness, can enhance accuracy.

	\begin{table*}[h]
		\centering
		
		\renewcommand\arraystretch{1.3}
		\caption{AUC results reported by different CAT methods}	
		\resizebox{1\textwidth}{!}{%
				\begin{tabular}{|ll|cccc|cccc|}
					\hline
					\multicolumn{2}{|l|}{}                                                                                                                                                            & \multicolumn{4}{c|}{\textbf{ASSISTments}}                                                                                                                                                                                                                                                                                                                                     & \multicolumn{4}{c|}{\textbf{Eedi2020}}                                                                                                                                                                                                                                                                                                                          \\ \hline
					\multicolumn{2}{|l|}{\multirow{2}{*}{
							\diagbox{\textbf{Selection Algorithm}}{\textbf{Measurement Model}} }}                                                                                                     & \multicolumn{2}{c|}{\textbf{IRT} \cite{embretson2013item}
					}                                                                                                                                                                    & \multicolumn{2}{c|}{\textbf{NeuralCD} \cite{wang2020neural}
					}                                                                                                                                          & \multicolumn{2}{c|}{\textbf{IRT} \cite{embretson2013item}
					}                                                                                                                                                      & \multicolumn{2}{c|}{\textbf{NeuralCD} \cite{wang2020neural}
					}                                                                                                                                          \\ \cline{3-10} 
					\multicolumn{2}{|l|}{}                                                                                                                                                            & \multicolumn{1}{c|}{AUC@5}                                                                          & \multicolumn{1}{c|}{AUC@20}                                                           & \multicolumn{1}{c|}{AUC@5}                                                                          & AUC@20                                                           & \multicolumn{1}{c|}{AUC@5}                                                            & \multicolumn{1}{c|}{AUC@20}                                                           & \multicolumn{1}{c|}{AUC@5}                                                                          & AUC@20                                                           \\ \hline
					\multicolumn{2}{|l|}{Random}                                                                                                                                                      & \multicolumn{1}{c|}{\begin{tabular}[c]{@{}c@{}}67.68 \cite{wang2023gmocat}\\ 70.68 \cite{zhuang2023bounded}\\ 65.86 \cite{ma2023novel}\end{tabular}} & \multicolumn{1}{c|}{\begin{tabular}[c]{@{}c@{}}68.43 \cite{wang2023gmocat}\\ 72.61 \cite{zhuang2023bounded}\end{tabular}} & \multicolumn{1}{c|}{\begin{tabular}[c]{@{}c@{}}67.73 \cite{wang2023gmocat}\\ 71.19 \cite{zhuang2023bounded}\\ 70.52 \cite{yu2023sacat}\end{tabular}} & \begin{tabular}[c]{@{}c@{}}69.70 \cite{wang2023gmocat}\\ 72.83 \cite{zhuang2023bounded}\end{tabular} & \multicolumn{1}{c|}{\begin{tabular}[c]{@{}c@{}}68.38 \cite{wang2023gmocat}\\ 69.05 \cite{zhuang2023bounded}\end{tabular}} & \multicolumn{1}{c|}{\begin{tabular}[c]{@{}c@{}}71.98 \cite{wang2023gmocat}\\ 74.82 \cite{zhuang2023bounded}\end{tabular}} & \multicolumn{1}{c|}{\begin{tabular}[c]{@{}c@{}}68.45 \cite{wang2023gmocat}\\ 69.32 \cite{zhuang2023bounded}\\ 73.67 \cite{yu2023sacat}\end{tabular}} & \begin{tabular}[c]{@{}c@{}}72.98 \cite{wang2023gmocat}\\ 74.99 \cite{zhuang2023bounded}\end{tabular} \\ \hline
					\multicolumn{1}{|l|}{\multirow{7}{*}{Statistical Algorithms}}                                             & Fisher Information  \cite{lord2012applications}                                                  & \multicolumn{1}{c|}{\begin{tabular}[c]{@{}c@{}}67.95 \cite{wang2023gmocat}\\ 71.33 \cite{zhuang2023bounded}\\ 66.41 \cite{ma2023novel}\end{tabular}} & \multicolumn{1}{c|}{\begin{tabular}[c]{@{}c@{}}69.26 \cite{wang2023gmocat}\\ 73.54 \cite{zhuang2023bounded}\end{tabular}} & \multicolumn{1}{c|}{--}                                                                             & --                                                               & \multicolumn{1}{c|}{\begin{tabular}[c]{@{}c@{}}68.92 \cite{wang2023gmocat}\\ 70.60 \cite{zhuang2023bounded}\end{tabular}} & \multicolumn{1}{c|}{\begin{tabular}[c]{@{}c@{}}72.66 \cite{wang2023gmocat}\\ 76.24 \cite{zhuang2023bounded}\end{tabular}} & \multicolumn{1}{c|}{--}                                                                             & --                                                               \\ \cline{2-10} 
					\multicolumn{1}{|l|}{}                                                                                    & KL Information  \cite{chang1996global}                                                 & \multicolumn{1}{c|}{\begin{tabular}[c]{@{}c@{}}67.92 \cite{wang2023gmocat}\\ 71.38 \cite{zhuang2023bounded}\end{tabular}}               & \multicolumn{1}{c|}{\begin{tabular}[c]{@{}c@{}}69.23 \cite{wang2023gmocat}\\ 73.57 \cite{zhuang2023bounded}\end{tabular}} & \multicolumn{1}{c|}{--}                                                                             & --                                                               & \multicolumn{1}{c|}{\begin{tabular}[c]{@{}c@{}}68.69 \cite{wang2023gmocat}\\ 69.79 \cite{zhuang2023bounded}\end{tabular}} & \multicolumn{1}{c|}{\begin{tabular}[c]{@{}c@{}}72.60 \cite{wang2023gmocat}\\ 75.73 \cite{zhuang2023bounded}\end{tabular}} & \multicolumn{1}{c|}{--}                                                                             & --                                                               \\ \cline{2-10} 
					\multicolumn{1}{|l|}{}                                                                                    & \begin{tabular}[c]{@{}l@{}}Fisher Information\\ + Robust \cite{zhuang2022robust}\end{tabular} & \multicolumn{1}{c|}{--}                                                                             & \multicolumn{1}{c|}{--}                                                               & \multicolumn{1}{c|}{--}                                                                             & --                                                               & \multicolumn{1}{c|}{68.93 \cite{zhuang2022robust}}                                                        & \multicolumn{1}{c|}{75.99 \cite{zhuang2022robust}}                                                        & \multicolumn{1}{c|}{--}                                                                               &             \multicolumn{1}{c|}{--}                                                       \\ \cline{2-10} 
					\multicolumn{1}{|l|}{}                                                                                    & \begin{tabular}[c]{@{}l@{}}KL Information\\ + Robust \cite{zhuang2022robust}\end{tabular}     & \multicolumn{1}{c|}{--}                                                                             & \multicolumn{1}{c|}{--}                                                               & \multicolumn{1}{c|}{--}                                                                             & --                                                               & \multicolumn{1}{c|}{68.90 \cite{zhuang2022robust}}                                                        & \multicolumn{1}{c|}{76.03 \cite{zhuang2022robust}}                                                        & \multicolumn{1}{c|}{--}                                                                               &        \multicolumn{1}{c|}{--}                                                              \\ \hline
					\multicolumn{1}{|l|}{\multirow{2}{*}{Active Learning}}                                                    & MAAT  \cite{bi2020quality} & \multicolumn{1}{c|}{\begin{tabular}[c]{@{}c@{}}68.24 \cite{wang2023gmocat}\\ 71.54 \cite{zhuang2023bounded}\\ 66.24 \cite{ma2023novel}\end{tabular}} & \multicolumn{1}{c|}{\begin{tabular}[c]{@{}c@{}}69.7 \cite{wang2023gmocat}\\ 73.08 \cite{zhuang2023bounded}\end{tabular}}  & \multicolumn{1}{c|}{\begin{tabular}[c]{@{}c@{}}67.96 \cite{wang2023gmocat}\\ 70.98 \cite{zhuang2023bounded}\\ 70.85 \cite{yu2023sacat}\end{tabular}} & \begin{tabular}[c]{@{}c@{}}71.17 \cite{wang2023gmocat}\\ 72.27 \cite{zhuang2023bounded}\end{tabular} & \multicolumn{1}{c|}{\begin{tabular}[c]{@{}c@{}}69.09 \cite{wang2023gmocat}\\ 70.32 \cite{zhuang2023bounded}\end{tabular}} & \multicolumn{1}{c|}{\begin{tabular}[c]{@{}c@{}}73.19 \cite{wang2023gmocat}\\ 74.46 \cite{zhuang2023bounded}\end{tabular}} & \multicolumn{1}{c|}{\begin{tabular}[c]{@{}c@{}}69.03 \cite{wang2023gmocat}\\ 70.12 \cite{zhuang2023bounded}\\ 74.33 \cite{yu2023sacat}\end{tabular}} & \begin{tabular}[c]{@{}c@{}}73.75 \cite{wang2023gmocat}\\ 75.83 \cite{zhuang2023bounded}\end{tabular} \\ \cline{2-10} 
					\multicolumn{1}{|l|}{}                                                                                    & \begin{tabular}[c]{@{}l@{}}MAAT\\ + Robust \cite{zhuang2022robust}\end{tabular}               & \multicolumn{1}{c|}{--}                                                                             & \multicolumn{1}{c|}{--}                                                               & \multicolumn{1}{c|}{--}                                                                             & --                                                               & \multicolumn{1}{c|}{68.93 \cite{zhuang2022robust}}                                                        & \multicolumn{1}{c|}{76.09 \cite{zhuang2022robust}}                                                        & \multicolumn{1}{c|}{70.39 \cite{zhuang2022robust}}                                                                      & 76.63 \cite{zhuang2022robust}                                                        \\ \hline
					\multicolumn{1}{|l|}{\multirow{4}{*}{\begin{tabular}[c]{@{}l@{}}Reinforcement\\ Learning\end{tabular}}}   & GMOCAT  \cite{wang2023gmocat}   & \multicolumn{1}{c|}{69.13 \cite{wang2023gmocat}}                                                                     & \multicolumn{1}{c|}{71.91 \cite{wang2023gmocat}}                                                       & \multicolumn{1}{c|}{69.95 \cite{wang2023gmocat}}                                                                     & 72.95 \cite{wang2023gmocat}                                                       & \multicolumn{1}{c|}{69.81 \cite{wang2023gmocat}}                                                       & \multicolumn{1}{c|}{74.19 \cite{wang2023gmocat}}                                                       & \multicolumn{1}{c|}{71.25 \cite{wang2023gmocat}}                                                                     & 75.76 \cite{wang2023gmocat}                                                       \\ \cline{2-10} 
					\multicolumn{1}{|l|}{}                                                                                    & NCAT     \cite{zhuang2022fully} & \multicolumn{1}{c|}{\begin{tabular}[c]{@{}c@{}}68.67 \cite{wang2023gmocat}\\ 71.53 \cite{zhuang2023bounded}\end{tabular}}               & \multicolumn{1}{c|}{\begin{tabular}[c]{@{}c@{}}71.06 \cite{wang2023gmocat}\\ 73.50 \cite{zhuang2023bounded}\end{tabular}} & \multicolumn{1}{c|}{\begin{tabular}[c]{@{}c@{}}69.28 \cite{wang2023gmocat}\\ 71.59 \cite{zhuang2023bounded}\\ 72.53 \cite{yu2023sacat}\end{tabular}} & \begin{tabular}[c]{@{}c@{}}71.68 \cite{wang2023gmocat}\\ 73.59 \cite{zhuang2023bounded}\end{tabular} & \multicolumn{1}{c|}{\begin{tabular}[c]{@{}c@{}}69.04 \cite{wang2023gmocat}\\ 72.11 \cite{zhuang2023bounded}\end{tabular}} & \multicolumn{1}{c|}{\begin{tabular}[c]{@{}c@{}}73.32 \cite{wang2023gmocat}\\ 76.66 \cite{zhuang2023bounded}\end{tabular}} & \multicolumn{1}{c|}{\begin{tabular}[c]{@{}c@{}}69.09 \cite{wang2023gmocat}\\ 74.10 \cite{zhuang2023bounded}\\ 74.49 \cite{yu2023sacat}\end{tabular}} & \begin{tabular}[c]{@{}c@{}}74.55 \cite{wang2023gmocat}\\ 79.12 \cite{zhuang2023bounded}\end{tabular} \\ \hline
					\multicolumn{1}{|l|}{\multirow{3}{*}{\begin{tabular}[c]{@{}l@{}}Meta Learning\\ Algorithms\end{tabular}}} & BOBCAT \cite{ghosh2021bobcat} & \multicolumn{1}{c|}{\begin{tabular}[c]{@{}c@{}}68.65 \cite{wang2023gmocat}\\ 71.68 \cite{zhuang2023bounded}\\ 66.41 \cite{ma2023novel}\end{tabular}} & \multicolumn{1}{c|}{\begin{tabular}[c]{@{}c@{}}70.97 \cite{wang2023gmocat}\\ 73.39 \cite{zhuang2023bounded}\end{tabular}} & \multicolumn{1}{c|}{\begin{tabular}[c]{@{}c@{}}69.50 \cite{wang2023gmocat}\\ 71.45 \cite{zhuang2023bounded}\\ 71.98 \cite{yu2023sacat}\end{tabular}} & \begin{tabular}[c]{@{}c@{}}71.80 \cite{wang2023gmocat}\\ 72.84 \cite{zhuang2023bounded}\end{tabular} & \multicolumn{1}{c|}{\begin{tabular}[c]{@{}c@{}}68.94 \cite{wang2023gmocat}\\ 74.42 \cite{zhuang2023bounded}\end{tabular}} & \multicolumn{1}{c|}{\begin{tabular}[c]{@{}c@{}}73.24 \cite{wang2023gmocat}\\ 76.58 \cite{zhuang2023bounded}\end{tabular}} & \multicolumn{1}{c|}{\begin{tabular}[c]{@{}c@{}}69.17 \cite{wang2023gmocat}\\ 76.00 \cite{zhuang2023bounded}\\ 75.12 \cite{yu2023sacat}\end{tabular}} & \begin{tabular}[c]{@{}c@{}}74.51 \cite{wang2023gmocat}\\ 79.00 \cite{zhuang2023bounded}\end{tabular} \\ \cline{2-10} 
					\multicolumn{1}{|l|}{}                                                                                    & DL-CAT  \cite{ma2023novel}                                                               & \multicolumn{1}{c|}{66.68 \cite{ma2023novel}}                                                                    & \multicolumn{1}{c|}{--}                                                               & \multicolumn{1}{c|}{--}                                                                             & --                                                               & \multicolumn{1}{c|}{--}                                                               & \multicolumn{1}{c|}{--}                                                               & \multicolumn{1}{c|}{--}                                                                             & --                                                               \\ \cline{2-10} 
					\multicolumn{1}{|l|}{}                                                                                    & SACAT \cite{yu2023sacat}                                                            & \multicolumn{1}{c|}{--}                                                                             & \multicolumn{1}{c|}{--}                                                               & \multicolumn{1}{c|}{75.24 \cite{yu2023sacat}}                                                                    & --                                                               & \multicolumn{1}{c|}{--}                                                               & \multicolumn{1}{c|}{--}                                                               & \multicolumn{1}{c|}{75.48 \cite{yu2023sacat}}                                                                    & --                                                               \\ \hline
					\multicolumn{1}{|l|}{\begin{tabular}[c]{@{}l@{}}Subset Selection \\ Algorithms\end{tabular}}              & BECAT \cite{zhuang2023bounded}                                                                  & \multicolumn{1}{c|}{71.44 \cite{zhuang2023bounded}}                                                                    & \multicolumn{1}{c|}{73.61 \cite{zhuang2023bounded}}                                                      & \multicolumn{1}{c|}{71.60 \cite{zhuang2023bounded}}                                                                    & 73.70 \cite{zhuang2023bounded}                                                      & \multicolumn{1}{c|}{73.15 \cite{zhuang2023bounded}}                                                      & \multicolumn{1}{c|}{76.82 \cite{zhuang2023bounded}}                                                      & \multicolumn{1}{c|}{76.30 \cite{zhuang2023bounded}}                                                                    & 79.36 \cite{zhuang2023bounded}                                                      \\ \hline
				\end{tabular}%

		}
		\label{auc_result}
	\end{table*}

\section*{Introduction to Representative Datasets}

	The following are introductions to several commonly used datasets, and more datasets can be found at our EduData GitHub link: \url{https://github.com/bigdata-ustc/EduData}
	
	\begin{itemize}
		\item ASSISTments \cite{feng2009addressing}, established in 2004, is an online tutoring platform in the United States that offers examinees both assessments and instructional support. To date, the ASSISTments team has released four public datasets\footnote{\url{https://sites.google.com/site/assistmentsdata/datasets/}}: ASSISTments2009, ASSISTments2012, ASSISTments2015, and ASSISTments2017. These datasets are response data and mostly collected from mathematics in middle school. They also include valuable side information, such as attempt count (the number of tries an examinee has made), ms first response (the time it takes for an examinee's first response), problem type, and average confidence.
		
		\item Junyi Dataset \cite{chang2015modeling} includes logs and exercise data from Junyi Academy, a Chinese online learning platform launched in 2012 using Khan Academy's open-source code. It features a detailed question hierarchy and relationships, labeled by experts. 
		
		
		\item MOOCCube\footnote{\url{https://www.biendata.xyz/competition/chaindream_mooccube_task2/}}, Massive Open Online Courses (MOOCs) are among the most prevalent platforms for online learning. This dataset collects examinees' responses to questions related to various computer science knowledge concepts. Additionally, the dataset includes the text of the problems, which can be used to enhance the performance of question selection, proficiency estimation, question characteristics analysis, etc.
		
		\item EdNet Dataset \cite{choi2020ednet} is a large collection of examinee learning records from the AI tutoring system Santa\footnote{\url{https://github.com/riiid/ednet}}, which is used for English language learning in South Korea. It focuses on examinees preparing for the eTOEIC (Test of English for International Communication) Listening and Reading Test, with over 131 million learning records from approximately 784,000 examinees.
		
		\item Eedi2020 Dataset \cite{wang2020diagnostic}, released for the NeurIPS 2020 Education Challenge, contains over 17 million records of examinees' responses to mathematics multiple-choice questions on the Eedi platform\footnote{\url{https://eedi.com/projects/neurips-education-challenge}}. It includes detailed information on examinees' choices, demographics, and containment relationships of knowledge concepts, as well as associated quiz and curriculum metadata. This extensive dataset enables in-depth analysis of examinee behaviors and the development of personalized tools.

	\end{itemize}
	\section*{Systematic Literature Review Protocol}
	{\color{black}{ To improve the transparency and reproducibility of this survey, we followed a lightweight SLR-style protocol for collecting and screening the literature. We searched major scholarly databases and digital libraries (e.g., Google Scholar, IEEE Xplore, ACM Digital Library, and arXiv) using keyword combinations related to computerized adaptive testing and psychometrics (e.g., ``computerized adaptive testing'', ``CAT'', ``item response theory/IRT'', ``exposure control'', ``content balancing'', ``online calibration'', ``multidimensional IRT'') as well as recent extensions to AI/LLM evaluation (e.g., ``adaptive evaluation'', ``LLM benchmarking'', ``agent-based assessment''). We focused primarily on peer-reviewed papers and widely used technical reports within the period 2000--2025, while allowing earlier seminal works when necessary for completeness. We applied inclusion criteria requiring clear methodological relevance to CAT/IRT (or their use in AI model evaluation). The screening was conducted in two stages: an initial title/abstract filtering followed by full-text review for highly relevant candidates. The selected studies were then organized into the taxonomy and sections presented (e.g., Figure 2, Table 2 and 3).''}}
	
\end{document}